%% file: cam-ready.tex
\RequirePackage{fix-cm}
\documentclass[twocolumn]{svjour3}  %

\smartqed  %

\usepackage{graphicx}

\journalname{}

\usepackage{mathptmx} 
\usepackage{makecell}
\usepackage[final]{microtype}
\usepackage[USenglish]{babel}
\usepackage{etoolbox}
\usepackage[T1]{fontenc}
\usepackage{amsmath}
\usepackage{amssymb}
\usepackage{natbib}
\usepackage{makecell}
\usepackage{afterpage}
\usepackage{placeins}
\usepackage{pifont} 
\usepackage{threeparttable}
\usepackage{epigraph}

\usepackage{multirow}
\usepackage{tabularx}
\usepackage{dsfont}
\usepackage{url}
\usepackage[table]{xcolor}
\usepackage[ruled,vlined]{algorithm2e}
\usepackage[export]{adjustbox}
\usepackage{booktabs}
\usepackage{subcaption} 
\usepackage[font=footnotesize,labelfont=bf]{caption}
\usepackage{longtable}
\usepackage{tabu}
\usepackage{supertabular}
\usepackage{enumitem}

\newcommand{\tabincell}[2]{\begin{tabular}{@{}#1@{}}#2\end{tabular}}

\definecolor{dg}{rgb}{0,0.694,0.298}
\definecolor{purple}{rgb}{0.4,0.176,0.569}
\definecolor{royalblue}{RGB}{65,105,225}
\definecolor{googleblue}{RGB}{66,133,244}
\definecolor{googlered}{RGB}{234,67,53}
\definecolor{googleyellow}{RGB}{251,188,5}
\definecolor{googlegreen}{RGB}{52,168,83}
\newcommand{\cmark}{\textcolor{dg}{\ding{52}}}%
\newcommand{\xmark}{\textcolor{red}{\ding{56}}}%

\usepackage{xspace}
\makeatletter
\DeclareRobustCommand\onedot{\futurelet\@let@token\@onedot}
\def\@onedot{\ifx\@let@token.\else.\null\fi\xspace}
\def\eg{\emph{e.g}\onedot} 
\def\ie{\emph{i.e}\onedot} 
 
\def\etc{\emph{etc}\onedot} 
\def\wrt{w.r.t\onedot} 

\makeatother

\newcommand{\revised}[1]{{\textcolor{black}{#1}}}
\newcommand{\revisedd}[1]{{\textcolor{black}{#1}}}

\usepackage[breaklinks=true,colorlinks,citecolor=royalblue,bookmarks=false]{hyperref}
\begin{document}

\title{Countering Malicious DeepFakes: Survey, Battleground, and Horizon}

\author{Felix Juefei-Xu \and
        Run Wang \and
        Yihao Huang \and
        Qing Guo \and
        Lei Ma \and
        Yang Liu
}

\institute{Felix Juefei-Xu is with Alibaba Group, USA 
            \and
            Run Wang is with the Key Laboratory of Aerospace Information Security and Trust Computing, School of Cyber Science and Engineering, Wuhan University, China
            \and
            Yihao Huang is with East China Normal University, China
            \and
            Qing Guo is with College of Intelligence and Computing, Tianjin University, China, and Nanyang Technological University, Singapore
            \and
            Lei Ma is with University of Alberta and Alberta Machine Intelligence Institute (Amii), Canada
            \and
            Yang Liu is with Zhejiang Sci-Tech University, China and Nanyang Technological University, Singapore
            \and
            Corresponding authors: Run Wang (\href{mailto:wangrun@whu.edu.cn}{email: wangrun@whu.edu.cn}) and Qing Guo (\href{mailto:qing.guo@ntu.edu.sg}{email: tsingqguo@ieee.org})
            \and
            Project lead: Felix Juefei-Xu.
}

\date{Received: date / Accepted: date}

\maketitle

\begin{abstract}

The creation or manipulation of facial appearance through deep generative approaches, known as \textit{DeepFake}, have achieved significant progress and promoted a wide range of benign and malicious applications, \eg, visual effect assistance in movie and misinformation generation by faking famous persons. The evil side of this new technique poses another popular study, \ie, \textit{DeepFake detection} aiming to identify the fake faces from the real ones. With the rapid development of the DeepFake-related studies in the community, both sides (\ie, DeepFake generation and detection) have formed the relationship of battleground, pushing the improvements of each other and inspiring new directions, \eg, the evasion of DeepFake detection. 
Nevertheless, the overview of such battleground and the new direction is unclear and neglected by recent surveys due to the rapid increase of related publications, limiting the in-depth understanding of the tendency and future works.

To fill this gap, in this paper, we provide a comprehensive overview and detailed analysis of the research work on the topic of DeepFake generation, DeepFake detection as well as evasion of DeepFake detection, with more than $318$ research papers carefully surveyed. We present the taxonomy of various DeepFake generation methods and the categorization of various DeepFake detection methods, and more importantly, we showcase the battleground between the two parties with detailed interactions between the adversaries (DeepFake generation) and the defenders (DeepFake detection). The battleground allows fresh perspective into the latest landscape of the DeepFake research and can provide valuable analysis towards the research challenges and opportunities as well as research trends and future directions. We also elaborately design interactive diagrams (\url{http://www.xujuefei.com/dfsurvey}) to allow researchers to explore their own interests on popular DeepFake generators or detectors.

\end{abstract}


\input{doc}

\end{document}

%% file: doc.tex
\section{Introduction}\label{sec:intro}

\epigraph{If you know the enemy and know yourself, you need not fear the result of a hundred battles. If you know yourself but not the enemy, for every victory gained you will also suffer a defeat. If you know neither the enemy nor yourself, you will succumb in every battle.}{\textit{The Art of War \\ Sun Tzu}}


Ever since digital visual media came along, there has always been a need to manipulate them for various purposes. Usually, such digital media manipulation requires domain expertise and is quite time and effort consuming, such as using professional software like Adobe Photoshop \citep{adobe_photoshop} for editing a photograph, or Adobe Lightroom \citep{adobe_lightroom} for retouching it. In the sound and voice domain, similar professional software is available for carrying out various types of signal manipulation such as using Adobe Audition \citep{adobe_audition}, or Auto-Tune \citep{autotune}, \etc. In the domain of motion pictures, the manipulation of videos oftentimes require very sophisticated theatrical visual effects (VFX) in the post-processing, and when it comes to recreating animated faces with realistic facial muscle movements and expressions, motion capture techniques with the help of high-speed tracking of markers are usually adopted, such as in James Cameron's Avatar \citep{avatar} movie. 
\begin{figure*}
	\centering 
    \includegraphics[width=\linewidth]{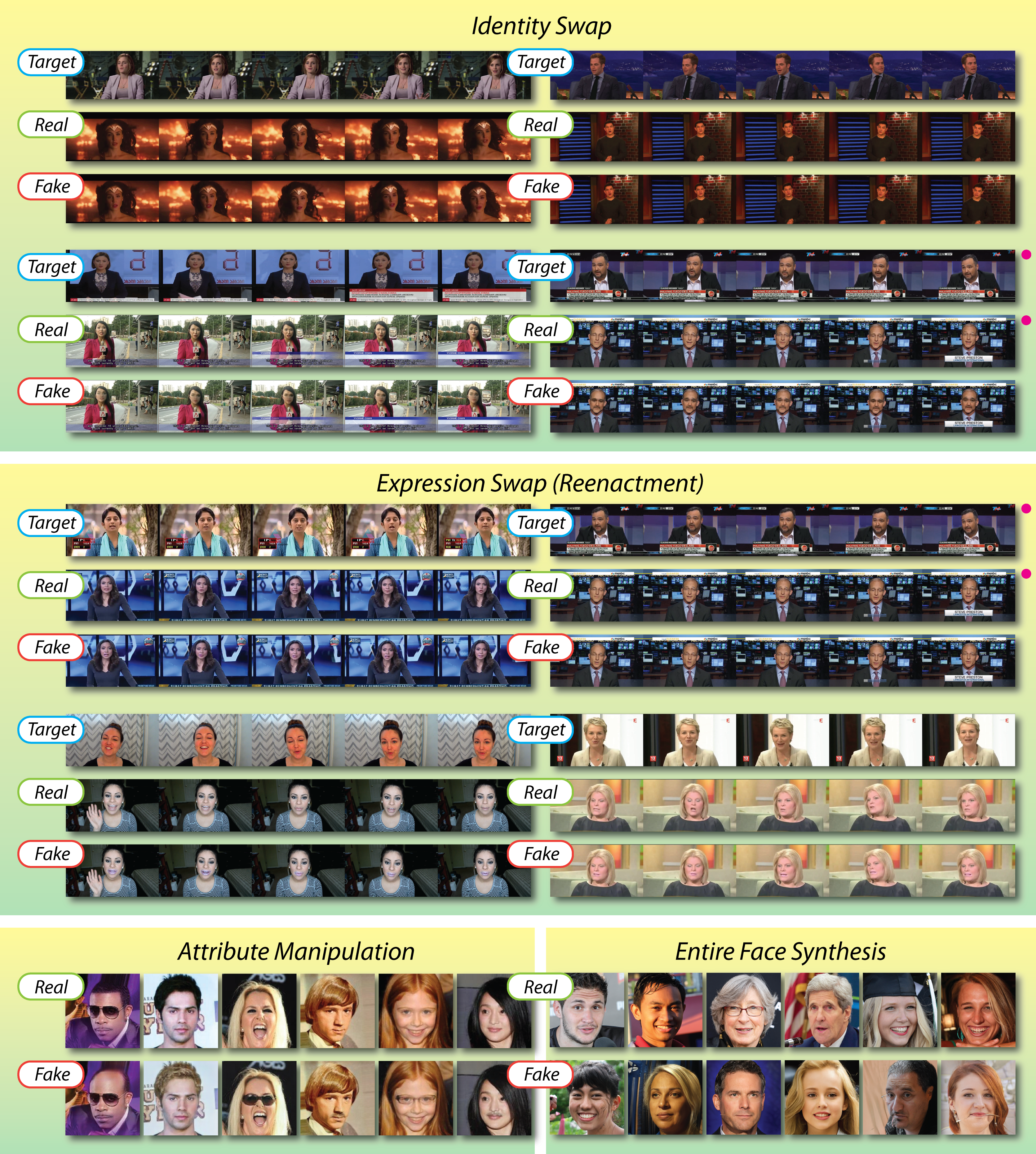}
	\caption{From top to bottom, the four panels illustrate the four categories of DeepFakes.  Four examples are shown for both `identity swap' and `expression swap', with each example associated with a target, real, and DeepFake sequences of 5 frames. Within each panel, the two examples in the top row show the DeepFake manipulation that is pretty subtle, which demonstrate the minuscule manipulations that some DeepFakes can present, and the two examples in the bottom row show more drastic DeepFake manipulations. Across `identity swap' and `expression swap', as a comparison, one example is shown in both scenarios and is highlighted by \textcolor{magenta}{\ding{108}}, to showcase the difference in the DeepFake frames for these two modalities coming from the same `target' and `real' sources. Readers are encouraged to zoom in on the image. Actual full-resolution videos are available on the project website (\url{http://www.xujuefei.com/dfsurvey}) to better illustrate the DeepFake phenomenon. For `attribute manipulation' and `entire face synthesis' on the bottom panel, both real and DeepFakes are shown. In terms of popularity as being attempted by DeepFake detectors according to the survey results, the ranking is as the following: identity swap $>$ entire face synthesis $>$ attribute manipulation $\gg$ expression swap.}
	\label{fig:fourcls}
\end{figure*}


With the advances of deep generative models such as autoregressive models \citep{van2016pixel,oord2016conditional}, variational autoencoders (VAE) \citep{kingma2013auto,kingma2019introduction}, normalizing flow models \citep{rezende2015variational}, and generative adversarial networks (GAN) \citep{goodfellow2014generative}, anyone can now produce a realistically looking face whose identity does not exist in the world, or perform facial manipulations, such as identity swap, in a video with a high level of realism. The AI- or deep learning-based face image and video manipulation is what the community refers to as the DeepFake. In contrast, the notion of CheapFake is recently coined to encompass non-AI (``cheap'') manipulations of multimedia content \citep{aneja2021mmsys}.
The low barriers to entry and wide accessibility of pre-trained high-performance DeepFake generator are what the problem is. DeepFake, when used maliciously, is a pressing and tangible threat to the integrity of media information available to us.



In this survey, we follow the widely adopted conventions and define DeepFake as \emph{the creation and the manipulation of facial appearance (attributes, identity, expression) through deep generative approaches}, and it can be classified into the following 4 categories: (1) entire face synthesis, (2) attribute manipulation, (3) identity swap, and (4) expression swap (\emph{aka} reenactment), as depicted in Figure~\ref{fig:fourcls}. 
\revisedd{Here, the facial attributes (such as hair color, facial hair, skin tone, eyewear, \etc) exclude identity and expression as attributes. DeepFake manipulations may exhibit different risk levels and the risk level highly depends on the type of specific applications and somewhat subjectively depending on the actual use case. Here, we provide some examples that may pose risks based on different DeepFake categories. The identity swap, altering the hard biometrics \citep{jain2007handbook} (pertains to identity information) of a subject, poses risks to a wide range of safety-critical scenarios since the identity information is tampered. The expression swap and attribute manipulation, although only tampers with the soft biometrics \citep{jain2007handbook} (pertains to facial attributes) of the subject, may also pose risks to certain applicable scenarios where people can easily verify the identity of the subject, but not what she/he says, such as in political elections, \eg, the DeepFaked Obama video \citep{elections}.
Generically speaking, the entire face synthesis may seem to pose a lower risk than the three categories mentioned above since it is not based on the manipulation of hard or soft biometrics of the subject. However, depending on the applications, even the entire face synthesis can become risky and troublesome, imagining the swarm of fake accounts on social media platforms.
In short, all four aforementioned DeepFake modalities can potentially pose high level of risks and need to be addressed properly.}
\revised{In terms of the popularity ranking measured by how often the categories are tested by DeepFake detectors according to the surveyed literature, the identity swap ranks the highest for the popularity, and the expression swap is the least attempted category.} For attribute manipulation and entire face synthesis, usually the DeepFake generation process does not require a target face, and one can tune the desired facial attributes through adjusting the latent vector during the deep generative modeling. For identity swap, the target can either be a video sequence or simply a single face image, with the former renders better swap results. For expression swap, the target is usually in the form of a video sequence. Although it is technically manageable to use just a single face image as the target, the result will look weird since the expression won't be changing throughout the entire DeepFake video.
\revisedd{From top to bottom, the four panels in Figure~\ref{fig:fourcls} illustrate the four categories of DeepFakes. 
Four examples are shown for both `identity swap' and `expression swap', with each example associated with a target, real, and DeepFake sequences of 5 frames.
Within each panel, the two examples in the top row show the DeepFake manipulation that is pretty subtle, which demonstrate the minuscule manipulations that some DeepFakes can present, and the two examples in the bottom row show more drastic DeepFake manipulations. 
Across `identity swap' and `expression swap', as a comparison, one example is shown in both scenarios and is highlighted by \textcolor{magenta}{\ding{108}}, to showcase the difference in the DeepFake frames for these two modalities coming from the same `target' and `real' sources. Readers are encouraged to zoom in on the image. Actual full-resolution videos are available on the project website (\url{http://www.xujuefei.com/dfsurvey}) to better illustrate the DeepFake phenomenon. 
For `attribute manipulation' and `entire face synthesis' on the bottom panel, both real and DeepFakes are shown.}


The DeepFake technology itself, in our opinion, is neutral and can be used both benignly and maliciously. We first discuss some of the positive and benign uses of DeepFake technology. For example, \cite{Synthesia} uses DeepFake technology to provide cost-effective synthetic training videos for companies during the Covid-19 pandemic when it is getting harder and more expensive to shoot corporate training videos with real actors. On the same line of thought, there has been an increase in a digital avatar or virtual assistance by means of DeepFake technology \citep{aiavatar} to be used for \eg, video conferencing scenarios \citep{wang2021one}. DeepFake can also be used for assisting the facial visual effects in movie and TV show production for re-creating a role appearance for some celebrities that may have passed away, or for paying tribute to the lost ones in a memorial concert. Sometimes, creative scenes can be made by using DeepFake to join together celebrities across geographic and generation boundaries. For example, \cite{Pinscreen} has used DeepFake technology to bring Prime Minister Mark Rutte, and Queen M\'{a}xima of The Netherlands, as well as other Dutch celebrities to a live TV broadcast. We have also seen a surge in popularity of DeepFake being used in consumer smartphone applications for everyday entertainment purposes, especially targeted for making viral videos on social media platforms, such as Zao \citep{Zao}, Reface \citep{Reface}, Facebrity \citep{Facebrity}, \etc. 

The aforementioned cases are benign DeepFakes. Powerful as the technology is, there is a thin line between good and evil depending on what the content is, as well as what the intent is, and it is easy to cross over. DeepFake can be maliciously capitalized by bad actors to cause real harm. For example, when it is applied to politicians and fueled with targeted misinformation or disinformation, it can really sway people’s opinions and can lead to detrimental outcomes such as manipulated and interfered election without people even knowing about it. To this date, many politicians and state leaders have been misrepresented by DeepFakes. Recently in September of 2020, DeepFake videos featuring Russian President Vladimir Putin and North Korean leader Kim Jong-un appeared on social media stating the message that ``America doesn't need any election interference from them; it will ruin its democracy by itself''. This was a campaign put forward by the nonpartisan advocacy group RepresentUs to protect voting rights during the then upcoming US presidential election, and the goal of the videos is to shock Americans into understanding the fragility of democracy \citep{Putin}. In April 2018, a DeepFake of Barack Obama was created by comedian Jordan Peele in collaboration with BuzzFeed which served as a public service announcement (PSA) to increase awareness of DeepFakes \citep{Obama}. During the most recent Christmas holidays, a DeepFake Queen Elizabeth II was shown dancing across TV screens as part of a British broadcaster's warning against the proliferation of misinformation \citep{Elizabeth}. Not just world leaders, celebrities or even average people can also fall victims of malicious DeepFakes. In fact, some of the earliest infamous use cases of malicious DeepFakes have been DeepFake pornography often featuring female celebrities. Financial fraud as a result of DeepFake or looming fake accounts on social media platforms created realistically by using DeepFakes are all examples of malicious use of DeepFake technology.


Lawmakers and governing bodies across the world are responding to the proliferation of malicious DeepFakes with new policies, regulations, and laws. Take the United States for example, in December 21, 2018, US Senator Ben Sasse introduced a bill ``S.3805 - Malicious Deep Fake Prohibition Act of 2018'' \citep{Sasse} that ``establishes a new criminal offense related to the creation or distribution of fake electronic media records that appear realistic''. In June 12, 2019, US Congresswoman Yvette Clarke introduced the ``H.R.3230 Defending Each and Every Person from False Appearances by Keeping Exploitation Subject to Accountability Act of 2019'', also known as the ``DEEP FAKES Accountability Act''. The act aims at ``combating the spread of disinformation through restrictions on deep-fake video alteration technology'' \citep{Clarke}. In July 9, 2019, US Senator Rob Portman introduced the bill ``S.2065 - Deepfake Report Act of 2019'' \citep{Portman} that requires ``the Science and Technology Directorate in the Department of Homeland Security to report at specified intervals on the state of digital content forgery technology. Digital content forgery is the use of emerging technologies, including artificial intelligence and machine learning techniques, to fabricate or manipulate audio, visual, or text content with the intent to mislead''. State-wide legislature entities have also responded by proposing counter-measures of DeepFakes such as the ``Nonconsensual pornography law'' in Virginia \citep{Virginia}, a law ``to criminalize publishing and distributing DeepFake videos intended to harm a candidate or influence results within 30 days of an election'' in Texas \citep{Texas}, and a similar law in California \citep{California}. Mirroring this new California law on political ads, the Chinese government ``makes it a criminal offense to publish deepfakes or fake news without disclosure'' \citep{Chinese}. Many more countries followed suit. 
  

Social media platforms are also actively taking measures to tackle synthetic and manipulated media on their platform. For example, Twitter signals viewers that a tweet contains manipulated media content such as DeepFakes by placing a tag on the tweet and providing a link to credible news articles debunking the hoax \citep{Twitter}. Facebook fosters the development of high-performance DeepFake detection tools by hosting the DeepFake Detection Challenge (DFDC) \citep{dolhansky2020deepfake} in December 2019 with 2,114 world-wide participants generating more than 35,000 models. 



\begin{figure}
	\centering 
    \includegraphics[width=0.49\linewidth]{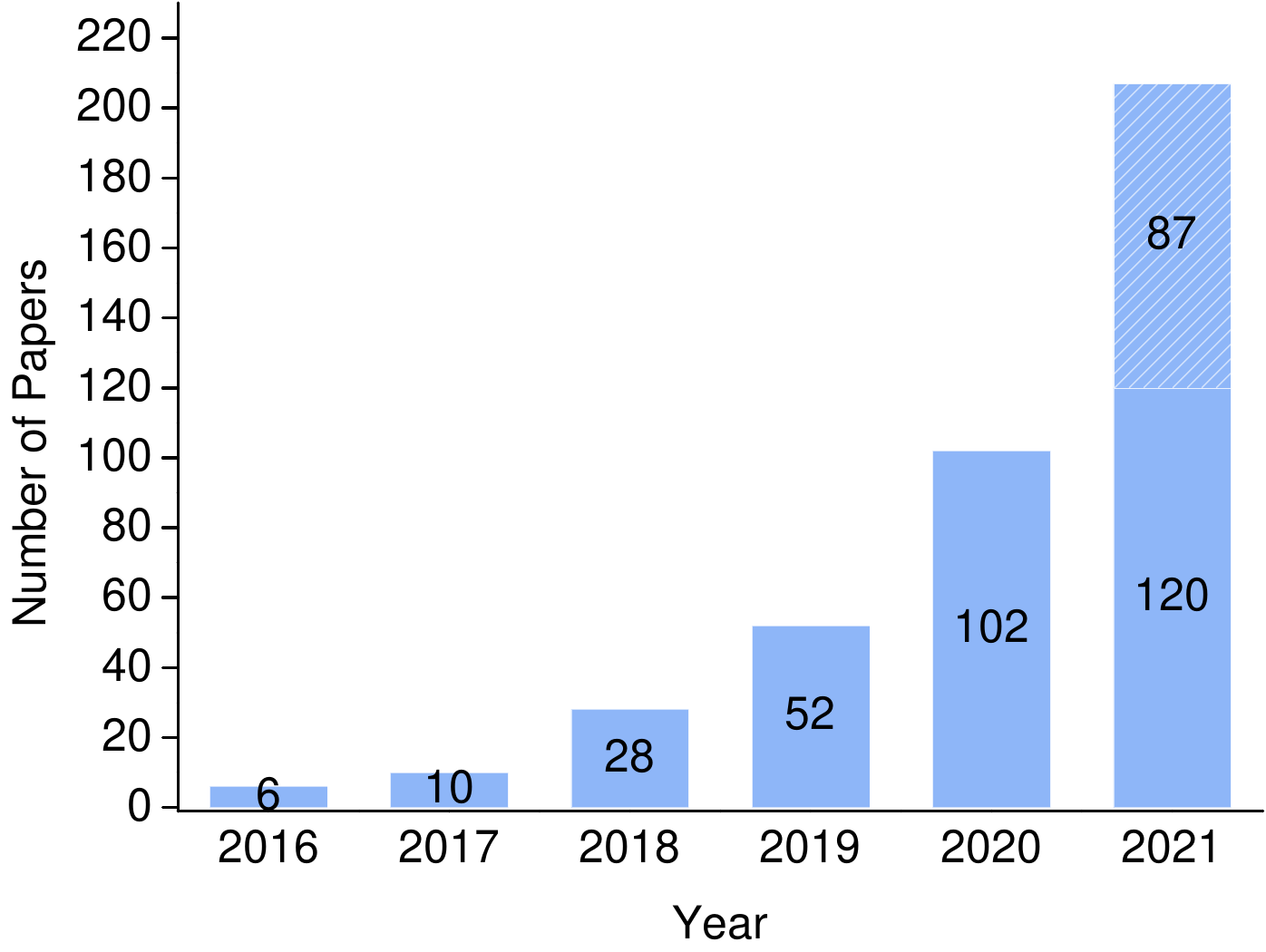}
    \includegraphics[width=0.49\linewidth]{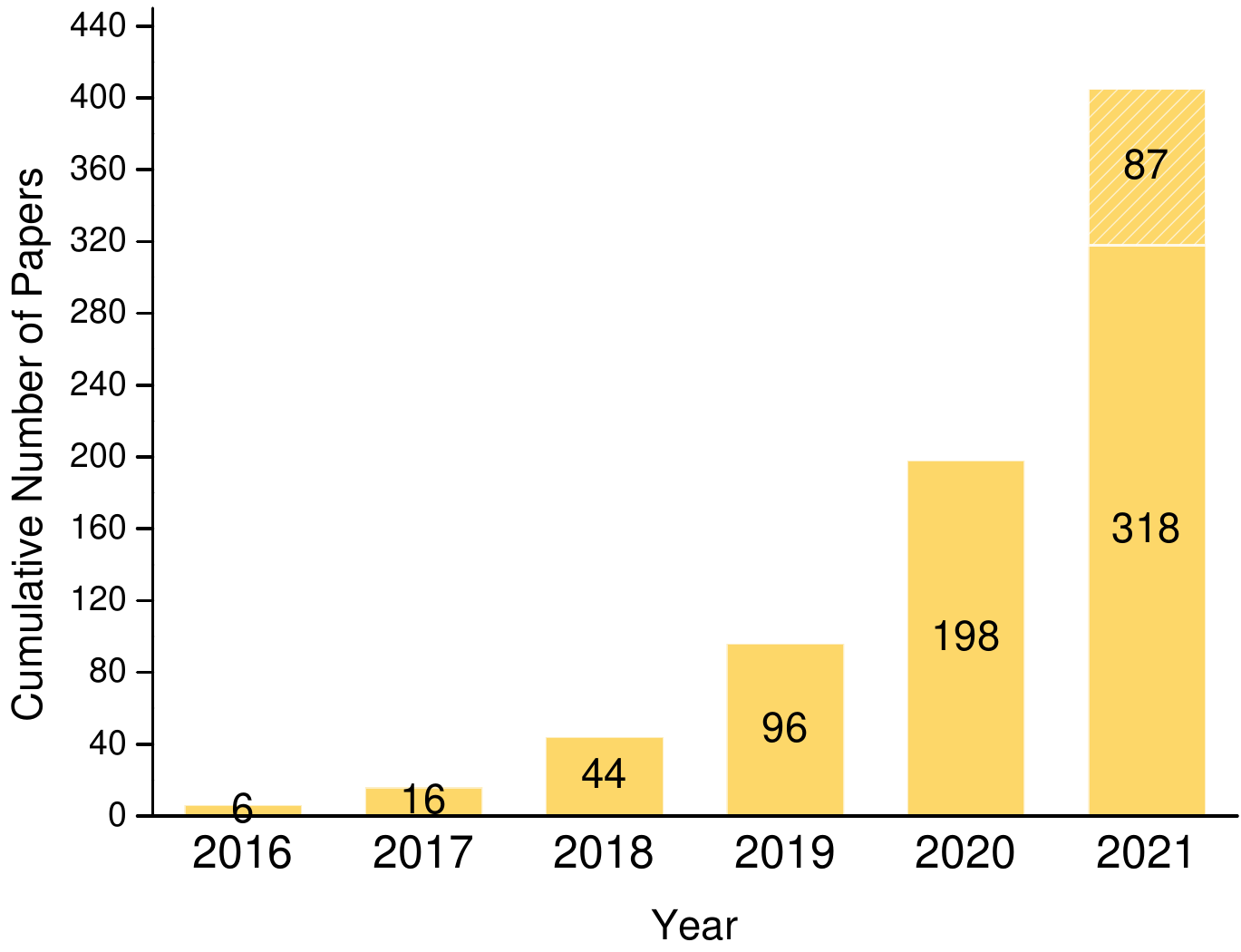}
	\caption{This figure shows a growing trend in the number of papers in the DeepFake field in recent years. The papers are collected according to the criteria introduced in Section~\ref{Survey Scope}, including arXiv, conference, and journal articles. They are categorized by the year of the last updated version. (L) The number of published DeepFake related papers year over year since its inception in 2016. (R) The cumulative number of published DeepFake related papers year over year since its inception in 2016. It shows that over 78\% papers were published in the last 2 years. The bars with shadow are the projected number of published papers.}
	\vspace{-15pt}
	\label{fig:paper_year}
\end{figure}

In the computer vision community, the study of DeepFake has certainly gained traction in recent years. Figure~\ref{fig:paper_year} shows the year-by-year number of papers on the topic of DeepFakes from its inception in 2016, and we will detail the paper collection schema in Section~\ref{sec:schema}. As shown in Figure~\ref{fig:paper_year}, around 78\% of the papers appeared in the last two years, indicating the trending research interests revolved around the topic of DeepFakes.

This paper seeks to further raise the awareness of the danger of the emerging DeepFake technology by surveying the state-of-the-art DeepFake literature. We organize the literature according to three aspects: the generation of DeepFakes in Section~\ref{sec:gen} (such as the aforementioned four main DeepFake categories, other generation methods not covered, and datasets), detection of DeepFakes in Section~\ref{sec:det} (such as methods based on spatial, frequency, and biological cues, as well as other detection methods not covered), and the evasion of DeepFake detection in Section~\ref{sec:evasion}. More importantly, we present the battleground between DeepFake generation methods and DeepFake detection methods in Section~\ref{sec:battle} where we analyze the tightly-knit interactions between the two parties and how the later DeepFake detectors outcompete the earlier ones, supplemented by statistics and analyses through large-scale visualizations. In Section~\ref{sec:horizon}, we recapitulate the findings after surveying each DeepFake-related topics and identify existing challenges and research opportunities while being forward-looking with discussions on the horizons and how the future generation technology surrounding DeepFakes may evolve into.


\subsection{Importance of the Survey of the Battleground}\label{sec:importance}

%


Unlike many other sub-fields of computer vision, the emerging domain of DeepFake generation and detection is, by nature, competitive. The rapid development of the defenders (\ie, DeepFake detectors) are further accelerated by the push from the adversaries (\ie, DeepFake generators), and vice versa. When a novel method, say from the DeepFake generator side, is being developed, the authors will naturally want their method to penetrate some latest DeepFake detectors, and the resulting generator, will later be attempted to be blocked by some newer detectors down the line. Just within last year or two, we have seen huge progresses made alternately on both sides of the battleground, both aiming to outcompete the other side. 

That is exactly why understanding each individual DeepFake generation and detection method, albeit important, may not be enough, and it does not tell the whole story. The collective understanding of the interplay and the interactivity of the methods both within each side and across two sides will bring fresh and clearer view of the landscape, and will bring new knowledge and insight for the development of the next-generation defenders and adversaries.



For these very reasons mentioned above, we have taken the effort to survey, extract, tabulate, and finally construct the battleground landscape between the DeepFake generation methods and DeepFake detection methods that have been proposed so far. By analyzing which of the DeepFake generation methods are attempted by each of the DeepFake detection method from the battleground, we are able to present to the readers the visualization of one aspect of the battleground using a Sankey diagram in Figure~\ref{fig:battle} with interactive diagrams available on the project website for better interactive probing. Readers can simply select one node on either side of the battleground, and all the highlighted paths will connect the corresponding opponents on the other side. Similarly, by analyzing which baseline DeepFake detection methods that each detector has benchmarked against, as observed from the battleground, we are able to showcase and trace back the technical evolution of each detector using a chord diagram in Figure~\ref{fig:cd} with interactive diagrams available on the project website. Readers can select one node on the interactive chord diagram, and the highlighted paths will connect all the corresponding baseline detection methods benchmarked by the selected detector.



Some collective knowledge are quite unique to the battleground and are not conspicuous if we only analyze individual methods. For example, from the battleground landscape in Figure~\ref{fig:battle} and Figure~\ref{fig:cd}, we can identify important trends on both sides, as well as algorithmic hot zones where seminal papers are indicated by busy nodes. We can also locate and discover where the major battles are fought, which nodes are the uprising feuds and which nodes are becoming obsolete algorithmically. With the color-coded paths indicating various types of detection methods (\ie, spatial-based, frequency-based, biological signal-based, \etc), we are able to provide an insightful understanding of what types of detection methods are being attempted on which particular generation methods. From the battleground, we are able to systemically extract the performance scores of each DeepFake detection method on every generation method it has evaluated on. 
Of course, the same generation method will be attempted by multiple detectors, each with a detection accuracy scores. By knitting the entire network and sorting the rankings, we are able to provide some strength measurement for each DeepFake generation method by means of Elo rating \citep{Elo}, as can be seen in Table~\ref{tab:paper_information}. Maybe more surprisingly, we may be able to identify the paths in the woods that are less traveled. From a practical point of view, if a practitioner is just entering the field, we are quite confident that the battleground presented in this survey paper will serve as an asset to both help identify a research direction more effectively, and to help understand the status quo more comprehensively. We provide more detailed analysis in Section~\ref{sec:battle}. 



As the battleground landscape serving as the cornerstone, we strive to continue building a DeepFake survey that is both content-rich and content-distinctive especially in terms of the following important traits such as the timeliness, the scale, the detailedness of the textual, tabular, and visual presentation, the thoroughness of the technical evolution, battleground analyses, and horizon analyses.





There has been previous work that surveyed or discussed aspects of the literature on topics related to DeepFake generation and detection. However, our survey that uniquely manifest the battleground landscape between the adversaries and the defenders still stands out.
\cite{mirsky2021creation} pivoted their survey to the DeepFake generation aspect with detailed model architecture charts for each individual DNN used for DeepFake generation methods the authors have surveyed, which is both informative and illustrative. However, less attention is paid to the DeepFake detection aspect, the technical evolution of both the generation and detection methods, and interplay and the battle between the two in their survey.
\cite{neekhara2021adversarial} provided a practical perspective that focuses on the adversarial threats to DeepFake detection. By studying the commonalities between various DeepFake detection methods by interpreting the model decisions using gradient-based saliency maps, the authors can create adversarial examples that are highly transferable across different DeepFake detection methods, revealing the vulnerabilities of the DeepFake detectors.
\cite{verdoliva2020media} discussed the interplay between multimedia forensics and DeepFakes. From the visual media integrity verification point of view, the authors have provided a detailed discussion on how the conventional and modern media forensics methods are conducted for general-purpose image and video manipulation. Later, they discuss how some of the methods can be applied to DeepFake detection. With the majority of the surveyed papers being conventional media forensics approaches, the overlapping is insignificant.
In addition, there are also a few relatable DeepFake surveys first published in late 2019 and early 2020 \citep{nguyen2019deep,tolosana2020deepfakes,lyu2020DeepFake}. As far as we know, although these earlier surveys shared similar themes, they were still deficient in terms of the following important aspects such as timeliness, scale, and detailedness of the survey, as well as thorough analyses on the technical evolution together with discussions and analyses of the horizons. Most importantly, the battleground landscape between the DeepFake generation and detection methods were not covered in these prior surveys. Here we list some of the comparisons between ours and prior surveys in Table~\ref{tab:surve_comp}. 
\begin{table}
\scriptsize
\centering
\caption{\revised{Comparisons with prior survey papers on DeepFake-related topics. Our survey papers is more informative and competitive in terms of the timeliness, scale, and detailedness, as well as the detailed manifestation of the technical evolution, battleground analyses, and horizon analyses.}}\label{tab:surve_comp}
\resizebox{1\linewidth}{!}{
\begin{threeparttable}
\begin{tabular}{l|c|c|c|c|c|c}
	\toprule 
    \makecell[c]{Prior\\Surveys} & Timeliness & Scale & Detailedness & \makecell[c]{Technical\\Evolution} & \makecell[c]{Battleground\\Analyses}& \makecell[c]{Horizon\\Analyses} \tabularnewline
	\midrule
	
	\cite{mirsky2021creation}    & Sep 13, 2020 & 193$^*$ & High   & Briefly & \xmark & Very Briefly \\
	\cite{verdoliva2020media}    & Jan 18, 2020 & 265$^*$ & Medium & Briefly & \xmark & Briefly \\
	\cite{nguyen2019deep}        & Apr 26, 2021 & 162$^*$ & Medium & Briefly & \xmark & Briefly \\
	\cite{tolosana2020deepfakes} & Jun 18, 2020 & 200$^*$ & Medium & Briefly & \xmark & Briefly \\
	\cite{lyu2020DeepFake}       & Mar 11, 2020 & 34$^*$  & Low    & Briefly & \xmark & Briefly \\
	
	\midrule 
	
	Ours & Aug 1, 2021 \cmark & $318$ $^\dagger$ \cmark   & High \cmark  & Thoroughly \cmark & \cmark & Thoroughly \cmark \\
	
	\bottomrule 
	
\end{tabular}
\begin{tablenotes}\footnotesize
\item[*] Total reference count, including auxiliary papers.
\item[$\dagger$] Excluding auxiliary papers.
\end{tablenotes}
\end{threeparttable}
}
\end{table}

\begin{figure}
	\centering 
	\includegraphics[width=\linewidth]{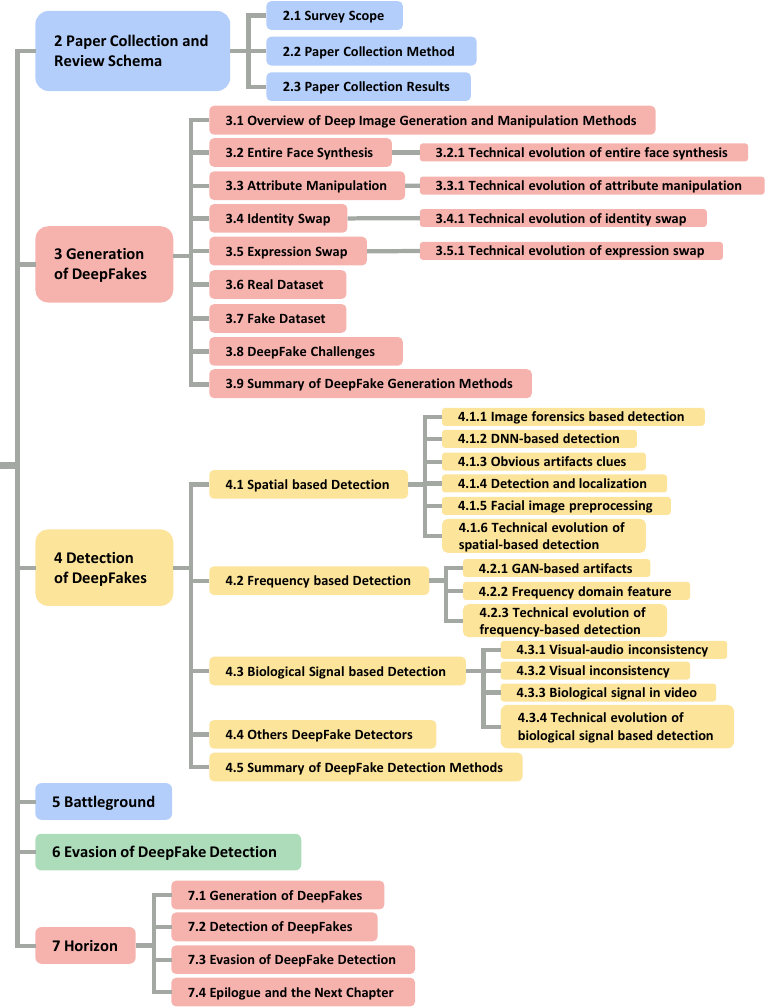}
	\caption{\revisedd{Tree diagram showing the paper structure.}}
	\label{fig:Treemap}
\end{figure}

In summary, this paper differentiates itself from the earlier survey papers with the following unique features and important contributions:
\begin{enumerate}[label={\arabic*)}]
    \item 
        \textbf{Timeliness.} The field of DeepFake generation and detection is fast growing. The paper collects and surveys the most up-to-date research work that shows the state-of-the-art performance in DeepFake generation and detection.
    \item
        \textbf{Scale.} The paper provides by far the largest scale and the most comprehensive survey of over $318$ research papers on the topics of DeepFake generation, DeepFake detection, and evasion of DeepFake detection, with detailed categorizations and analyses. 
    \item
        \textbf{Detailedness.} This survey utilizes many graphic visualizations and diagrams (Sankey diagrams, fishbone diagrams, chord diagrams, \etc.), as well as many very detailed long tables to best illustrate and highlight the properties, interactions, characteristics, evolution, and important traits of the technical methods surveyed and discussed. The diagrams and long tables may serve as a starting point for quick lookup and method comparisons, and the accompanying text provides in-depth discussion as a complement.  
    \item
        \textbf{Technical evolution analyses.} In addition to the detailed methodological introduction of each individual DeepFake generation and detection methods following the taxonomy, this paper uniquely manifests the technical evolution among the aforementioned methods, providing a more comprehensive and clearer picture of the evolutionarily technological landscape of the state-of-the-art DeepFake generation methods and detection methods.
    \item   
        \textbf{Battleground analyses.} There exists an adversarial and battling nature between DeepFake generation methods and DeepFake detection methods. Each party progresses by outcompeting the other side. The paper uniquely captures the tightly-connected interactivities between DeepFake generation and detection methods as well as among various detection methods themselves, revealing evidence for research hot zones and trends for future topics. 
    \item  
        \textbf{Horizon analyses.} By virtue of the detailed surveys and analyses of the battleground landscape, the paper exposes challenges, identifies open research problems, and hints promising future research directions on the topics of DeepFake generation and detection. 
    \end{enumerate}
Figure~\ref{fig:Treemap} depicts the tree diagram of the paper structure. \revised{For enhanced readability, here we provide suggestions for different types of readers and practitioners. For those who are new to the field and want to get up to speed quickly, it is advised to first go through Section~\ref{sec:sum_gen} and Section~\ref{sec:sum_det} for the summary of the DeepFake generation and detection methods, along with the technical evolution highlights in Sections~\ref{sec:tec_evo_entire_face_synthesis}, \ref{sec:tec_evo_attribute_manipulation}, \ref{sec:tec_evo_identity_swap}, \ref{sec:tec_evo_expression_swap} and Sections~\ref{spatial_summary}, \ref{frequency_summary}, \ref{biological_summary}, respectively. Then the readers can move on to Section~\ref{sec:battle} for the Battleground, Section~\ref{sec:evasion} for the Evasion methods, and Section~\ref{sec:horizon} for the Horizon. For those who are already in the field and are interested in the technical details, it is advised to first go through all the subsections in Section~\ref{sec:gen} and \ref{sec:det}, and then move on to Section~\ref{sec:battle} and \ref{sec:evasion}. For those who are already in the field and want to identify latest technical trend in the DeepFake generation and detection literature, or in a particular direction, it is advised to first go through Section~\ref{sec:battle} in detail, and then pay attention to Sections~\ref{sec:tec_evo_entire_face_synthesis}, \ref{sec:tec_evo_attribute_manipulation}, \ref{sec:tec_evo_identity_swap}, \ref{sec:tec_evo_expression_swap} and Sections~\ref{spatial_summary}, \ref{frequency_summary}, \ref{biological_summary} for the technical evolution, and finally move on to individual sections.}

\section{Paper Collection and Review Schema}\label{sec:schema}

This section covers the survey scope, survey methodology, and paper collection results.
\begin{figure*}
    \centering
    \begin{subfigure}[b]{0.225\linewidth}
        \centering
        \includegraphics[width=\linewidth]{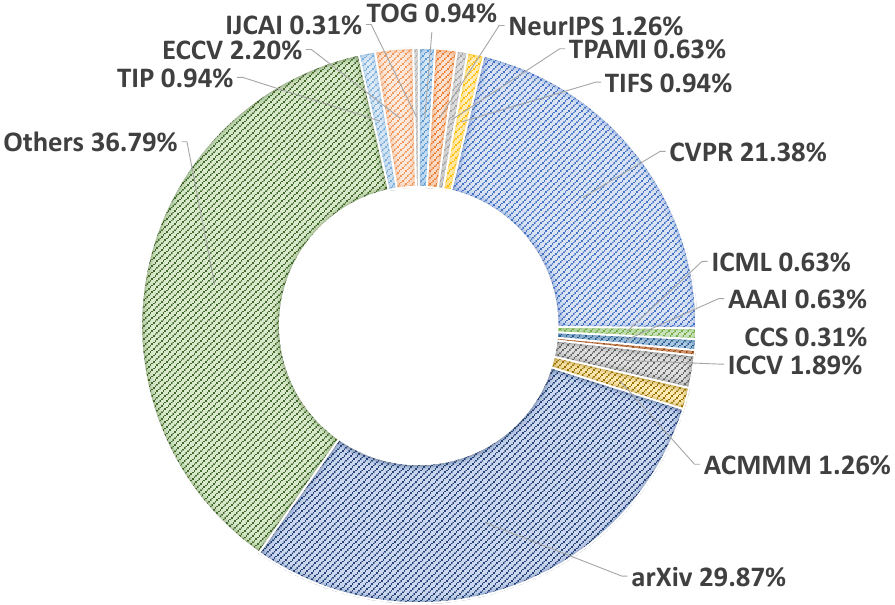}
        \caption{}
    \end{subfigure}
    \begin{subfigure}[b]{0.245\linewidth}
        \centering
        \includegraphics[width=\linewidth]{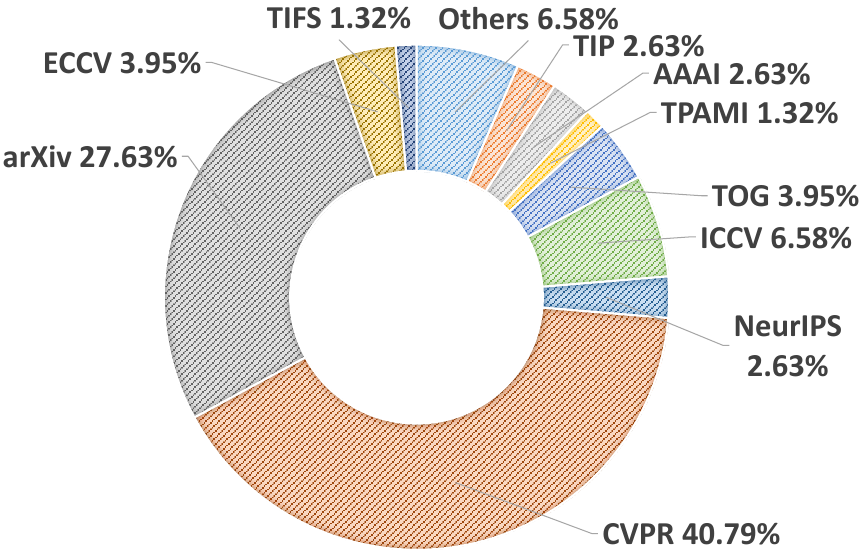}
        \caption{}
    \end{subfigure}
    \begin{subfigure}[b]{0.265\linewidth}
        \centering
        \includegraphics[width=\linewidth]{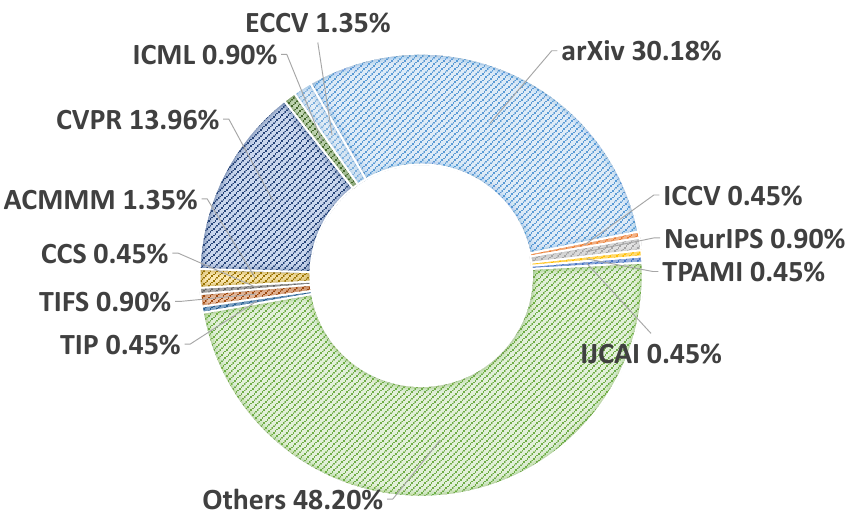}
        \caption{}
    \end{subfigure}
    \begin{subfigure}[b]{0.245\linewidth}
        \centering
        \includegraphics[width=\linewidth]{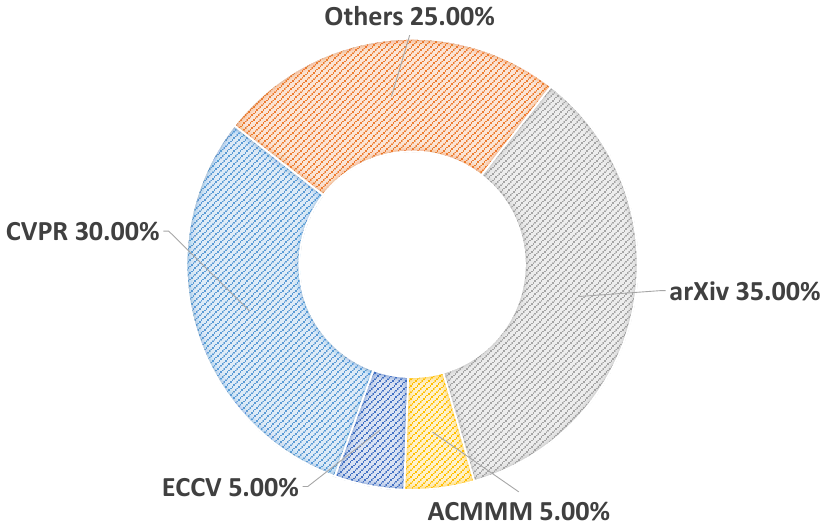}
        \caption{}
    \end{subfigure}
        \caption{(a) For the papers in DeepFake research area, we can find that a large amount of papers are from Others and arXiv. The papers published in top conferences and journals only account for a third of the total. Furthermore,  a lot of top papers are published in CVPR, which accounts for about half the published top papers. (b) For DeepFake generation methods, we can find that a large amount of the papers are from CVPR and arXiv. Two thirds of the generation papers are published in top conferences and journals. (c) For DeepFake detection methods, we can find that a large amount of the papers are from Others and CVPR. The published top papers only make up a small part of the total. This suggests that progress in DeepFake detection is not enough. (d) For DeepFake evasion methods, we can find that the volume of articles is not large and a large amount of the papers are from arXiv. One fourth of the papers are published in top conferences.}
        \label{fig:paper_source}
\end{figure*}

\subsection{Survey Scope}\label{Survey Scope}

The paper focuses on the technical aspect of DeepFakes via surveying related research papers on the topics of DeepFake generation, DeepFake detection, and evasion of DeepFake detection. The social and ethical aspects regarding DeepFakes, although briefly touched in this paper, are not the focus of this survey. 
A more loosely defined DeepFake could mean voice, gesture, body, and any type of media manipulation. In this survey paper, we focus solely on the topics of facial DeepFakes.

\subsection{Paper Collection Method}

To collect research papers on DeepFakes across different research areas as comprehensive as possible, we first collect the papers from a Github repository\footnote{\scriptsize{\url{https://github.com/clpeng/Awesome-Face-Forgery-Generation-and-Detection}}} which lists more than 100 papers about DeepFakes generation, DeepFake detection, and evasion of DeepFake detection. Then, we apply keywords matching to search DeepFake papers from two popular scientific databases (Google Scholar and DBLP) and arXiv where the newest papers are posted. The keywords are listed below.
\begin{itemize}
\item DeepFake / fake / editing / edit  + facial image / face / swapping / video  
\item synthesis / GAN-synthesized / AI-synthesized
\item manipulation / forgery / tampered face + detection 
\end{itemize}
To ensure a more comprehensive and accurate survey, we also manually browse the recent three years publications in top-tier conferences and the corresponding workshops to avoid the limitations of keywords matching. Additionally, for DeepFake generation papers, we mainly collect the methods which have been mentioned in the previous DeepFake detection papers. 

\subsection{Paper Collection Results}

Overall, we have collected $318$ papers from Google Scholar and arXiv. The papers mainly include DeepFake generation, DeepFake detection, and evasion of DeepFake detection topics. Figure \ref{fig:paper_source} shows the distribution of papers published in different research venues. Here we categorize the papers of top conferences and journals to specific classification (\ie, one of CVPR, ICCV, ECCV, TPAMI, \etc). Meanwhile, we bucket other published papers in non-top conferences and journals into the ``Others'' category. For unpublished papers with exposure on arXiv, we define them as the ``arXiv'' category. 

For DeepFake generation methods, we can find that a large amount of the papers are from CVPR and arXiv. Two-thirds of the generation papers are published in top conferences and journals. This shows that a large proportion of DeepFake generation methods have gone through strict peer-review process and are relatively trustworthy. For DeepFake detection methods, however, we can find that a mass of the papers are from arXiv and Others. The published top-tier papers only make up a small part of the total. This suggests that progress in DeepFake detection is slower compared to that of the DeepFake generators. For DeepFake evasion methods, we can find that the volume of articles is small and a large amount of the papers are from arXiv, ECCV and Others. Half of the papers are published in top conferences. This shows that there is considerable improvement in DeepFake evasion research area. 

In summary, for the papers in DeepFake research area, we can find that a large number of papers are from Others and arXiv. The papers published in top conferences and journals only account for a third of the total. Furthermore, a lot of top papers are published in CVPR, which accounts for about two-thirds the published top papers.





\section{Generation of DeepFakes}\label{sec:gen}

In the research area of DeepFake generation, there are two parts to focus on: generation methods and datasets. We first provide an overview of the general DeepFake techniques and introduce the methods which can be seen as DeepFake generation methods in a broader sense (\eg, style transfer, inpainting, super resolution, \etc{}) in Section \ref{Other generation methods}. This gives readers an understanding of the general DeepFake generation techniques.

Then we focus on face appearance-related DeepFake methods which are most anticipated and influential in the field. For DeepFake generation methods, according to the consensus in the DeepFake field \citep{mirsky2021creation,verdoliva2020media,tolosana2020deepfakes}, there are mainly four categories based on their function: entire face synthesis, attribute manipulation, identity swap, and expression swap as depicted in Figure~\ref{fig:fourcls}. We introduce these methods in Section \ref{Entire Face Synthesis}, \ref{Attribute Manipulation}, \ref{Identity Swap}, and \ref{Expression Swap}. 
The other important part is the dataset. We highlight the major real image/video datasets which are used in the generation methods above and the fake image/video datasets generated by them. The content is introduced in Section~\ref{Real Dataset} and \ref{Fake Dataset}, followed by DeepFake challenges in Section~\ref{DeepFake Challenges}. The various DeepFake generation methods are summarized in Section~\ref{sec:sum_gen}. The highlights of the technical evolution of DeepFake generation methods discussed in Sections~\ref{sec:tec_evo_entire_face_synthesis}, \ref{sec:tec_evo_attribute_manipulation}, \ref{sec:tec_evo_identity_swap}, \ref{sec:tec_evo_expression_swap}.

\subsection{Overview of Deep Image Generation and Manipulation Methods}\label{Other generation methods}

In this section, we aim to give readers an understanding of the general deep image generation techniques which can be seen as the DeepFake technique in a broader sense.

The methods such as style transfer \citep{chen2018gated}, \citep{yao2019attention}, image inpainting \citep{yu2018generative}, \citep{yu2018generative}, rendering \citep{chen2017photographic}, \citep{park2019semantic}, \citep{liu2021self}, super resolution \citep{dai2019second}, \citep{guo2020closed}, \citep{liu2020residual}, \citep{mei2020image}, fusion \citep{lin2019coco}, de-identification \citep{sun2018hybrid}, \citep{li2019anonymousnet}, \citep{maximov2020ciagan}, \etc{} share some of the technical similarities of DeepFake generation. However, these methods are not the focus of this survey. Instead, we mainly pay attention to the face appearance-related DeepFake methods.

\begin{figure*}
	\centering 
    \includegraphics[width=\linewidth]{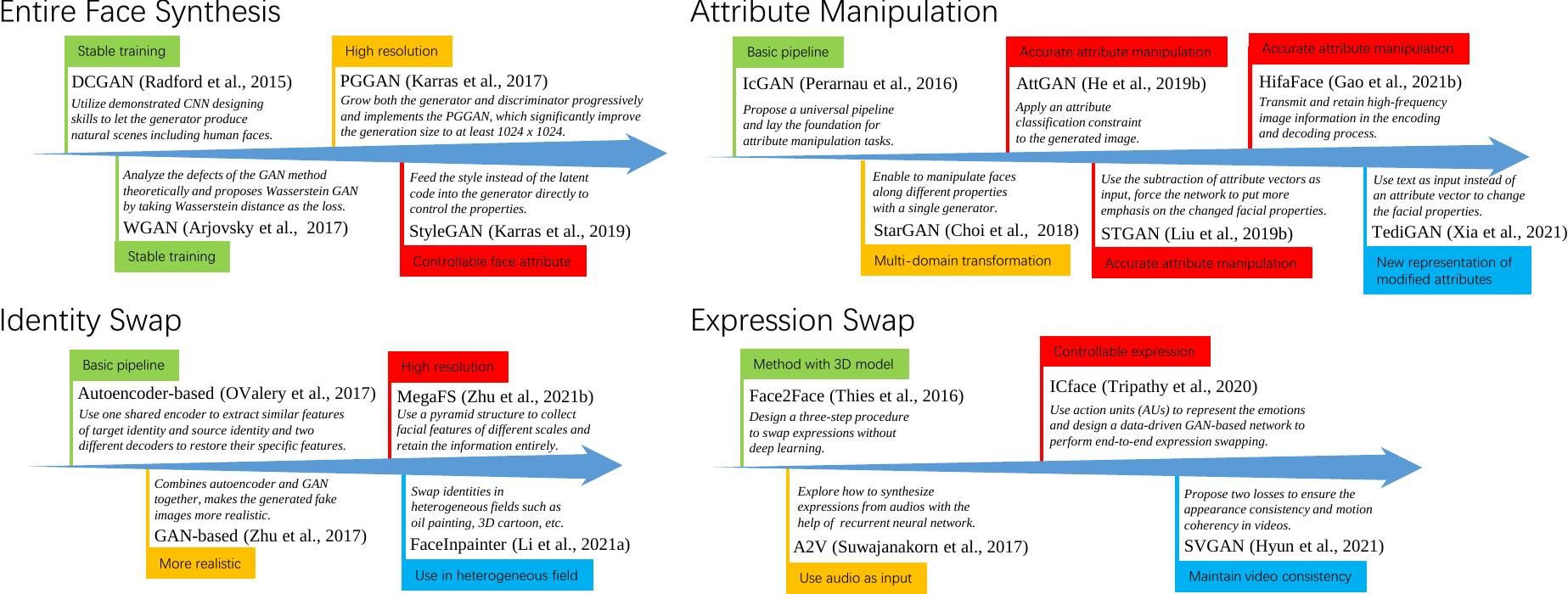}
	\caption{The evolution of DeepFake generation techniques with a fishbone diagram for each DeepFake generation type.}
	\label{fig:fish_bone_generation}
\end{figure*}

As introduced in Section~\ref{Entire Face Synthesis} to Section~\ref{Expression Swap}, existing DeepFake generation methods mainly consist of four types (\ie, entire face synthesis, attribute manipulation, identity swap, expression swap), depending on the tasks' requirements. 
To achieve a comprehensive survey, we detail the technological evolution of the four types, respectively. 
Note that, we focus on introducing their intuitive idea and categorizing their optimization methods. To make it clear,  we use Figure~\ref{fig:fish_bone_generation} to show the overall evolution of the four types of DeepFake generation methods, respectively.

\subsection{Entire Face Synthesis}\label{Entire Face Synthesis}
\begin{definition}
\label{def:definition_entire_face_synthesis}
Entire face synthesis aims to generate non-existent fake face image $\mathbf{x_{f}}$ from random vector $\mathbf{v}$ with neural network $\phi(\cdot)$. That is $\mathbf{x_{f}} = \phi(\mathbf{v}).$
\end{definition}

For entire face synthesis tasks, GANs and VAEs are both feasible neural networks $\phi(\cdot)$. However, according to the surveys \citep{verdoliva2020media,nguyen2019deep,tolosana2020deepfakes,lyu2020DeepFake}, GANs are the mainstream baseline technique. Many famous and popular entire face synthesis techniques such as PGGAN, StyleGAN, \etc{} are GAN-based and are able to generate high-quality DeepFake images. Compared with GANs, VAEs usually generate less realistic faces (\ie, being blurred). The reason why the images generated by VAEs tend to be blur is that the training principle makes VAEs assign a high probability to training data points, which cannot ensure that blurry data points are assigned to a low probability \citep{huang2018introvae}. Since the DeepFake images generated by VAE are not realistic enough, this section mainly introduces the GAN-related works.

Using GANs for entire face synthesis is actually a kind of distribution mapping. The GANs learn the mapping from random distribution to human face distribution. Existing state-of-the-art methods can stably generate high-resolution images. which is benefited from the continuous improvement of the GAN network and training procedure. However, the current methods still suffer from the training difficulty (\eg, mode collapse problem of GAN training procedure). Furthermore, the generated images are not realistic enough due to the lack of general knowledge of face distribution (\eg, facial symmetry).

As shown in the entire face synthesis part of Figure~\ref{fig:fourcls}, the fake images are very realistic and it is hard to distinguish real images from fake ones. Existing works mainly focus on improving the training stability, resolution, and controllable face attribute.

The classical examples are deep convolutional GAN (DCGAN) \citep{radford2015unsupervised}, Wasserstein GAN (WGAN) \citep{arjovsky2017wasserstein}, progressive growing GAN (PGGAN) \citep{karras2017progressive}, and style-based GAN (StyleGAN) \citep{karras2019style}. 

The very first work which combines convolutional neural network (CNN) and GAN is a deep convolutional generative adversarial network (DCGAN) \citep{radford2015unsupervised}. It focuses on unsupervised learning and has comparable performance in image classification tasks with the pre-trained discriminator. The generator of it can easily manipulate lots of the semantic properties (\ie, manipulate attribute of a human face) of generated images profile from its interesting vector arithmetic properties.

Two years later, there has been an explosion of in-depth research on GANs. Some GANs put emphasis on the stability of the GAN training. The groundbreaking work is Wasserstein-GAN (WGAN) \citep{arjovsky2017wasserstein}. In the first published GANs, the procedure requires researchers to carefully maintain a balance between generator and discriminator. The mode dropping phenomenon also occurs frequently. To solve these hot potatoes, WGAN has theoretically minimized a reasonable and efficient approximation of the expectation-maximization (EM) distance, which only needs a few optimization designs on the original GANs.

There are many types of research based on the WGAN. Gradient penalty WGAN (WGAN-GP) \citep{gulrajani2017improved} has indicated that WGAN sometimes still generates poor samples or fails to converge. The reason is that WGAN uses weight clipping to enforce a Lipschitz constraint. To improve the weight clipping operation, they have proposed to penalize the norm of the gradient of the discriminator with respect to its input fake image. The new designs train stably when generating high-quality home images.
Simply using Wasserstein probability can not simultaneously satisfy sum invariance, scale sensitivity, and unbiased sample gradients. To improve it, Cramer GAN (CramerGAN) \citep{bellemare2017cramer} has combined the best of the Wasserstein and Kullback-Leibler divergences to propose the Cram\'{e}r distance. The CramerGAN performs significantly better than the WGAN.
Boundary equilibrium GAN (BEGAN) \citep{berthelot2017began} is also an improved version of WGAN \citep{arjovsky2017wasserstein}. To further balance the power of the discriminator against the generator, they have suggested pairing an equilibrium enforcing method with a loss derived from the Wasserstein distance together. They also have proposed a new way to control the trade-off between image diversity and visual quality.

Some other works focus on how to generate high-resolution images. The resolution of the images generated by them is at least $1024\times1024$. Meanwhile, the images are detailed, and it is quite difficult to distinguish between the genuine and the fake, which is very amazing.
PGGAN \citep{karras2017progressive} is the very first and famous work that proposes an effective method to generate high-resolution images. The resolution of the generated images is $1024\times1024$. It has proposed to progressively grow both the image resolution of the generator and discriminator. The images are starting from a low resolution and being detailed step by step with the new layers added in the model. This method is very reasonable in that it can speed up the training as well as greatly stabilize the GAN. However, the training procedure is still not good enough that some of the generated images are far from real.
BigGAN \citep{brock2018large} has attempted to generate high-resolution diverse images from datasets such as ImageNet \citep{deng2009imagenet}. They have applied orthogonal regularization to enforce the generator to be satisfied with a simple ``truncation trick''. Thus, the user can control the trade-off between image fidelity and variety by reducing the variance of the generator’s input. 

To control the properties of generated images elaborately, StyleGAN \citep{karras2019style} has proposed a new design to automatically learn the unsupervised separation of high-level attributes such as pose and human identity. The architecture also leads to stochastic variation in the generated images (\eg, freckles, hair). Furthermore, it enables intuitive, scale-specific control of the synthesis. 
StyleGAN2 \citep{karras2020analyzing} has exposed several typical artifacts of StyleGAN and has proposed changes in both model architecture and training methods to address them. In particular, they have encouraged good conditioning in the mapping from latent codes to images by the new design of generator normalization, progressive growing, and generator regularization.

Different from the previous methods which use the GAN framework, generative flow (Glow) \citep{kingma2018glow} is a flow-based generative model that uses an invertible $1\times1$ convolution. The method is based on the theory that a generative model optimized towards the plain log-likelihood objective has the ability to generate efficient realistic-looking synthesis and manipulate large images.

\subsubsection{Technical Evolution of Entire Face Synthesis}\label{sec:tec_evo_entire_face_synthesis}

To summarize, a straightforward way for the entire face synthesis is to regard it as an image generation task.
\cite{goodfellow2014generative} propose the generative adversarial network (GAN) and the trained generator is able to produce meaningful examples, \eg, handwritten numbers and human faces. However, the generated examples are usually with low resolution and artifacts. Moreover, the networks are unstable to train \citep{goodfellow2014generative}.
Then, to address the issues of GANs, \cite{radford2015unsupervised} propose the deep convolutional generative adversarial network (DCGAN), which utilizes demonstrated CNN designing skills to let the generator produce natural scenes including human faces.
As a result, DCGAN is identified as the early work for DeepFake \citep{radford2015unsupervised}.

After that, some similar GAN-based methods are proposed. Nevertheless, they usually encounter difficulties (\eg, non-convergence, gradient vanishing, collapsed mode) in the training procedure. 
To solve these difficulties and generate images effectively and stably, \cite{arjovsky2017wasserstein} analyzes the defects of the GAN method theoretically and proposes Wasserstein GAN by taking Wasserstein distance as the loss.
Wasserstein distance cures the main training problems of GANs including DCGAN with relaxed requirements on the balance between discriminator and generator, the designs of network architecture, and reduced mode dropping phenomenon \citep{arjovsky2017wasserstein}. As a result, Wasserstein GAN usually generates more natural and higher quality fake faces than DCGAN.

Although achieving significant progress, the GAN-based methods could not synthesize fake faces with high resolution. In particular, the fake images generated by the previous methods are less than 256 $\times$ 256. The capability of generating high-resolution images is particularly important for real-world applications since the low-resolution fake faces can be easily identified.
To address this issue, \cite{karras2017progressive} propose to grow both the generator and discriminator progressively and implement the PGGAN, which significantly improves the generation size to at least 1024 $\times$ 1024. 
Specifically, the main challenges of generating high-resolution fake images stem from two aspects. First, it is easy to distinguish the fake images from the real ones with the discriminator when the resolution is high, which magnifies the vanishing gradient problem of the generator during GAN training. Second, due to the memory limitation, generating high-resolution images leads to smaller batch sizes, affecting the stability of training. To solve these problems, PGGAN proposes to progressively increase the image resolution of the generated images of the generator and the discriminator in the training procedure with the spatial resolution of the generator and discriminator being 4 $\times$ 4 pixels at the beginning. As the training advances, they incrementally add layers to the generator and discriminator and increase the spatial resolution of the generated images. As a result, the method is able to produce high-resolution fake faces.

The above methods are able to generate natural and realistic fake faces but cannot control the properties we want to be fake.
To complement the capability, based on PGGAN, \cite{karras2019style} propose StyleGAN that feeds the style instead of the latent code into the generator directly. Specifically, StyleGAN transforms the latent code to `style' code via a nonlinear mapping network. Then, the style code is used to conduct the adaptive instance normalization after each convolution. In addition, Gaussian noise images are embedded after each convolution layer for stochastic detail generation. The StyleGAN is able to generate high-resolution images with higher image quality and wider detailed variations.

Overall, the technical evolution of the entire face synthesis mainly follows the development of the GAN, which aims to solve the challenges that may arise in real-world applications, \eg, high-resolution and style-guided generations. For the entire face synthesis, we think the improvements of training stability, resolution, and controllable facial attributes are the major technical characteristics that have been evolved throughout the years, and we highlight seminal works DCGAN, WGAN, PGGAN, StyleGAN in the top left panel of Figure \ref{fig:fish_bone_generation}, showcasing the improvements in different angles.

\subsection{Attribute Manipulation}\label{Attribute Manipulation}
\begin{definition}
\label{def:definition_attribute_manipulation}
Attribute manipulation aims to modify facial properties $\mathbf{P}$ of a real face image $\mathbf{x_{r}}$ to generate a new fake image $\mathbf{x_{f}}$ with neural network $\phi(\cdot,\cdot)$. That is $\mathbf{x_{f}} = \phi(\mathbf{x_{r}},\mathbf{P}).$
\end{definition}

Using GANs for attribute manipulation is actually a kind of latent space editing. The key point is the quality of the GAN inversion technique. With a better attribute disentangle technique, the GANs for attribute manipulation can achieve more accurate attribute control. Existing state-of-the-art methods (\eg, HifaFace \citep{gao2021high}) can perform accurate face editing while maintaining rich details of non-editing areas. However, the current methods are still limited by the labels in the training dataset. That is, it is difficult to control the attributes that do not exist in the label of the training dataset.

As shown in the attribute manipulation part of Figure~\ref{fig:fourcls}, the real images are modified with facial attributes such as bald, blond hair, eyeglasses, \etc. Existing works mainly focus on improving attribute manipulation accuracy.

Attribute manipulation is also known as face editing, which can not only modify simple face attributes such as hair color, bald, smile, but also retouch complex attributes like gender, age, \etc. The classical examples are StarGAN \citep{choi2018stargan} and selective transfer GAN (STGAN) \citep{he2019attgan}.

Invertible conditional GAN (IcGAN) \citep{perarnau2016invertible} is the earliest attempt in GAN-based facial attribute manipulation. Based on an extension of the idea of conditional GAN (cGAN) \citep{mirza2014conditional}, they have evaluated encoders to map a real image into a latent space and a conditional representation, which allows the reconstruction and modification of arbitrary attributes of real human face images.
The expression generative adversarial network (ExprGAN) \citep{ding2018exprgan} has added an expression controller module that can learn an expressive and compact expression code to the encoder-decoder network. The expression controller module enables it to edit photo-realistic facial expressions with controllable expression intensity.

Previous studies can only perform image-to-image translation for two domains, which is cumbersome and time-consuming. To be more efficient, StarGAN \citep{choi2018stargan} has designed a single model to perform image-to-image translations for multiple domains. It allows simultaneous training of multiple different-domain datasets within a single network. 
As an improvement, StarGAN2 \citep{choi2020stargan} simultaneously satisfies two properties in image-to-image translation: diversity of generated images as well as scalability over multiple domains. To represent diverse styles of a specific domain, they have replaced StarGAN's domain label with their domain-specific style code. To adapt the style code, they have proposed two modules: a mapping network and a style encoder. The style code can be extracted from a given reference image with a style encoder while the mapping network can transform random Gaussian noise into a style code. Utilizing these style codes, the generator learns to successfully synthesize diverse images over multiple domains.

Although StarGAN is effective, due to the limitation of the content of the datasets, it can only generate a discrete number of expressions. To address this limitation, GAN animation (GANimation) \citep{pumarola2018ganimation} has introduced a novel GAN conditioning method based on action units (AU) annotations. It defines the human expression with a continuous manifold of the anatomical facial movements. The magnitude of activation of each AU can be controlled independently. Different AUs can also be combined with each other with this method.

Most of the previous work inevitably changes the attribute irrelevant regions. To solve this problem, spatial attention GAN (SaGAN) \citep{zhang2018generative} propose a module to only change the attribute-specific region and keep the other area unchanged. This work properly exploits the attention mechanism to ensure a better face editing effect, which shows the feasibility of the attention mechanism in face manipulation.

Previous methods have attempted to establish an attribute-independent latent representation for further attribute editing. However, since the facial attributes are relevant, requesting for the invariance of the latent representation to the attributes is excessive. Therefore, simply forcing the attribute-independent constraint on the latent representation not only restricts its representation ability but also may result in information loss, which is harmful to attribute editing. To solve this problem, facial Attribute editing (AttGAN) \citep{he2019attgan} has removed the strict attribute-independent constraint from the latent representation. It just applies the attribute classification constraint to the generated image to guarantee the correctness of attribute manipulation. Meanwhile, it groups attribute classification constraint, reconstruction learning, and adversarial learning together for high-quality facial attribute editing. The model supports direct attribute intensity control on multiple facial attribute editing within a single model.

Considering that the specific editing task is only related to the changed attributes instead of all target attributes, as an improvement of AttGAN, STGAN \citep{liu2019stgan} has selectively taken the difference between target and source attribute vectors as the input of the model. Furthermore, they have enhanced attribute editing by adding a selective transfer unit that can adaptively select and modify the encoder feature to the encoder-decoder.

Mask-guided portraiting editing (MaskPE) \citep{gu2019mask} proposes a unique way to manipulate face attributes. They use a face parsing mask to guide the generation of face attributes. The main idea is to separately embed five facial components (\ie, left eye, right eye, mouth, skin \& nose, and hair) into latent codes based on face parsing masks. Then they can modify any facial component independently.

Due to the lack of paired images during training, previous methods typically use cycle consistency to keep the non-editing attributes unchanged. However, even if the cycle consistency is satisfied, images may still be blurry and lose rich details from input images for that the generator tends to find a tricky way (\ie, encodes the rich details of the input image into the output image in the form of hidden signals) to satisfy the constraint of cycle consistency. To solve this problem, \cite{gao2021high} propose high-fidelity arbitrary face editing (HifaFace) to maintain rich details (\eg, wrinkles) of non-editing areas. Their work has two improvements. The first is that they directly feed the high-frequency information of the input image into the end of the generator with wavelet-based skip-connection, which relieves the pressure of the generator to synthesize rich details. The second is that they use another high-frequency discriminator as a complement to the image-level discriminator to encourage the image to have rich details.

Text-guided diverse image
generation and manipulation GAN (TediGAN) \citep{xia2021tedigan} is a special network for multi-modal image generation and manipulation with
textual descriptions. They map the image
and text into a common embedding space to learn text-image matching. The method allows the user to edit the appearance of different attributes interactively.

HistoGAN \citep{afifi2021histogan} chooses a special angle to manipulate images. They use color histograms to manipulate the color blending of the images only, which is a very targeted research content.

\subsubsection{Technical Evolution of Attribute Manipulation}\label{sec:tec_evo_attribute_manipulation}

In contrast to the entire face synthesis, the attribute manipulation of the face is usually regarded as the image manipulation task that changes facial properties (\eg, hair's color and style, facial hair, \etc) of input faces. The task can be tackled by incorporating encoder-decoder and GANs.

\cite{perarnau2016invertible} start the first work for attribute manipulation, which is denoted as IcGAN. This work proposes a universal pipeline and lays the foundation for attribute manipulation tasks. Specifically, IcGAN first encodes real images into the latent space and then changes the latent codes corresponding to different facial properties. After that, it decodes the changed latent codes to fake face images. Although effective, IcGAN can be time-consuming when it aims to perform the manipulation of multiple facial attributes since each attribute is addressed via an independent deep model.

To allow flexible GANs and avoid high time consumption, \cite{choi2018stargan} propose the StarGAN and design a network that enables to manipulate faces along with different properties with a single generator.
The intuitive idea is to encode the real image and its respective source domain label via the generator and produce the fake image.  
At the same time, the discriminator is designed to classify the real or fake faces and identify the domain. 
As a result, the learned GAN cannot only contain the semantic representation of facial properties and the respective domain information.

Although face attribute editing is available, the desired attribute variation is specified as the input for the encoder while the latent representation is not constrained, which may result in information loss and lead to over-smooth and distorted generation.
To alleviate these drawbacks, \cite{he2019attgan} propose the AttGAN that applies an attribute classification constraint to the generated image. As a result, the correct change of attributes can be guaranteed.

In addition to the AttGAN, \cite{liu2019stgan} implement a better generator (STGAN), forcing the network to emphasize the desired changed facial properties while preserving the other property areas.
To this end, they use the subtraction of attribute vectors (\ie, source vector - target vector) to replace the source vector as the inputs of the decoder.
Moreover, they propose a novel architecture named the selective transfer unit to improve attribute manipulation ability and image quality. As a result, this work can improve attribute manipulation accuracy as well as perception quality.

Although AttGAN and STGAN do not raise obvious variations on the undesired face attributes, they may lead to the variations of details (\eg, wrinkles) in those undesired areas. This is caused by the cycle consistency during training. Specifically, due to the lack of paired images during training, AttGAN and STGAN use the cycle consistency to avoid the changing of the undesired attributes. 
However, these generators map the details of the input image to a new one via hidden signals to achieve the cycle consistency that cannot be guaranteed.
To solve this problem, \cite{gao2021high} propose high-fidelity arbitrary face editing (HifaFace) to maintain rich details of undesired attribute areas.
The main idea is to transmit and retain high-frequency image information in the encoding and decoding process. There are two main improvements. First, they directly feed the high-frequency information of the input image into the end of the generator with wavelet-based skip-connection, which relieves the pressure of the generator to synthesize rich details. Second, they add a high-frequency discriminator as a complement to encourage the image to have rich details, which prevents the generator from finding a trivial solution for cycle consistency. As a result, the generated images have rich details with higher fidelity.

Recently, \cite{xia2021tedigan} propose a new design to use text as input instead of an attribute vector to change the facial properties of real images. The main idea of them is to map text and images to the same semantic latent space. Thus they can use text to replace attribute vectors. They extend the availability and diversity of attribute manipulation tasks.

In summary, recent works for attribute manipulation mainly focus on how to change the desired face attributions effectively while preserving other areas via a single generator. For the attribute manipulation, we think the basic pipeline and the improvements of multi-domain transformation, accurate attribute manipulation, and new representation of modified attributes are the major technical characteristics that have been evolved throughout the years, and we highlight seminal works IcGAN, StarGAN, AttGAN, STGAN, HifaFace, TediGAN in the top right panel of Figure \ref{fig:fish_bone_generation}, showcasing the improvements in different angles.

\subsection{Identity Swap}\label{Identity Swap}

\begin{definition}
\label{def:definition_identity_swap}
Identity swap aims to replace the identity of source image $\mathbf{x_{s}}$ by the identity $\mathbf{t_{i}}$ of target image $\mathbf{x_{t}}$ with neural network $\phi(\cdot,\cdot)$ and generate a new fake image $\mathbf{x_{f}}$. That is $\mathbf{x_{f}} = \phi(\mathbf{x_{s}},\mathbf{t_{i}}).$
\end{definition}
As shown in the identity swap part of Figure~\ref{fig:fourcls}, the images in the fake videos have uneven qualities. Existing works mainly focus on improving the realism and resolution of the image. 

In general, the architectures used for these functions mainly fall into two categories: autoencoder-based and GAN-based. The classical works are cycle-consistent GAN (CycleGAN) \citep{zhu2017unpaired} and \citep{zhu2021one}.

The methods which make the concept of DeepFake, especially identity swapping, become widely known are methods based on autoencoder. The autoencoder-based methods \citep{swap-face} have no specific name or architecture. However, as they are all based on autoencoder, their pipeline is similar. The methods use one shared encoder and two independent decoders. The encoder and one of the decoders are trained by source identity while the encoder and the other decoder are trained by target identity. When the model is well trained, the encoder has the ability to extract the common features of source and target identities while the decoder records the specific features. At inference time, the image of the source identity goes through the encoder and the opposite decoder, producing a realistic swap.

Nowadays, GAN-based methods are the mainstream in identity swap. The first work of the GAN-based method was CycleGAN \citep{zhu2017unpaired} proposed in 2017. In previous works, the absence of paired examples is always the limitation in image transformation tasks. CycleGAN has artfully solved this problem. Define a source domain $X$ and a target domain $Y$, it builds a mapping $G$ : $X\rightarrow Y$ which is highly under-constrained and similarly constructs an inverse mapping $F$ : $Y\rightarrow X$. Then the cycle consistency loss which enforces $F(G(X)) \approx X$ (and vice versa) is the optimization target of the model. Through this circulation, there is no need for paired samples. 
Meanwhile, although not mentioned in the paper, the framework of CycleGAN can be used for identity swap easily. Faceswap-GAN \citep{Faceswap-GAN} is the implementation of CycleGAN which provides an identity swap functionality. It simply adds the adversarial loss and perceptual loss to encoder architecture.

Face swapping GAN (FSGAN) \citep{nirkin2019fsgan} is a subject agnostic method that doesn't rely on the training of pairs of faces. It is also the first to simultaneously adjust the pose, expression, and identity variations for both a single image and a video sequence.

The research in identity swap has been stagnated for a long time until the appearance of FaceShifter \citep{li2020advancing}. It proposes a two-stage procedure for high fidelity and occlusion-aware face-swapping. Unlike many existing face-swapping works that leverage only limited information from the target image, FaceShifter generates the swapped face by thoroughly and adaptively exploiting the information of the target image. 

Appearance optimal transport (AOT) \citep{zhu2020aot} has formulated appearance mapping as an optimal transport problem. They have proposed an AOT model to formulate it in both latent and spatial space. In particular, a relighting module is designed to simulate the optimal transport plan. The optimization target is minimizing the Wasserstein distance of the learned features in the latent space, which enables better performance and less computation than conventional optimization.

Information disentangling and
swapping network (InfoSwap) \citep{gao2021information} aims to extract the most expressive information for identity representation. The main idea is to formulate the learning of disentangled representations as optimizing an information bottleneck trade-off. The information bottleneck principle provides a guarantee that in the latent space, areas scored as identity-irrelevant indeed contribute little information to predict identity.

Megapixel level face swapping (MegaFS) \citep{zhu2021one} has proposed the first one-shot ultra-high-resolution face swapping method. To overcome the information loss in the encoder, they use a hierarchical representation face encoder (HieRFE) to find the complete face representation. Then they use a face transfer module (FTM) to control multiple attributes synchronously without explicit feature disentanglement. The contributions are ground-breaking.

FaceInpainter \citep{li2021faceinpainter} proposes a controllable face inpainting network under heterogeneous domains (\ie, oil painting, 3D cartoons, pencil drawing, exaggerated drawing, \etc). The framework has two stages. In the first stage, they use a styled face inpainting network (SFI-Net) to map the identity and attribute properties to the swapped face. The second stage contains a joint refinement network (JR-Net) that refines the attributes and identity details, generating occlusion-aware and high-resolution swapped faces with visually natural fused boundaries.

\subsubsection{Technical Evolution of Identity Swap}\label{sec:tec_evo_identity_swap}
Identity swap is usually achieved by conducting replacement on the identity-related features and decoding these features to the image level. As a result, the identity of the input face image (\ie, the source identity) can be changed to the desired one (\ie, the target identity).
Specifically, the general pipeline is implemented via the autoencoder \citep{swap-face} that contains one shared encoder and two independent decoders. It first uses the encoder to extract features of the source and target identities, respectively, and get their respective latent codes. Then, the method uses the two independent decoders to reconstruct the source and target images, respectively. 
During the identity swap, the latent code of the source identity is fed to the decoder of the target identity. As a result, the decoded face is swapped. 
Under this pipeline, one of the key problems is how to extract or select faces' features as the latent code for replacement. Another problem is how to make the generated images more realistic.

To improve the fidelity of the generated face images, \cite{Faceswap-GAN} combines the autoencoder and GAN. Compared with the naive autoencoder-based methods, it adopts GAN's advantages while making the fake images more realistic due to the supervision of the discriminator.
The above methods do well on low-resolution images but cannot generate high-resolution swapped faces. This issue stems from the compressed representations whose information is insufficient for high-quality face generation during the swapping.
To alleviate this drawback, MegaFS \citep{zhu2021one} proposes the hierarchical representation face encoder (HieRFE), which uses a pyramid structure to collect facial features under different scales and retain the information entirely. 
As a result, the method can focus on processing the high-level semantic information while retaining the low-level details after identity swap at the megapixel level.

The previous methods have achieved great progress on photorealistic images. However, their capability of addressing source and target images with heterogeneous materials (\eg, oil painting, 3D cartoon, \etc) is less effective due to the different textures of the source and target images.
A recent work \citep{li2021faceinpainter} designs a two-stage framework to solve the issue. After explicitly disentangling the foreground (\ie, face and neck) from the background (\eg, hair, clothes, \etc) in the source identity, the first stage is to combine the attribute codes of the source identity and identity code of the target identity with the fixed background extracted from source identity. As a result, the swapped face contains the target identity with source background and attributes.
However, the segmented background and the generated foreground cannot be well integrated under some complex scenes. Then, the second stage is to refine the coarse result from the first stage to make the source and the target background consistent at the fusion boundary.

Overall, the technical evolution of identity swap mainly focuses on the better separation of identity and attribute features of the source and target images and how to fuse them. For the identity swap, we think that the basic pipeline and the improvements of realistic, resolution, and heterogeneous fields are the major technical characteristics that have been evolved throughout the years, and we highlight seminal works autoencoder-based method, GAN-based method, MegaFS, and FaceInpainter in the bottom left panel of Figure \ref{fig:fish_bone_generation}, showcasing the improvements in different angles.

\subsection{Expression Swap}\label{Expression Swap}

\begin{definition}
\label{def:definition_expression_swap}
Expression swap aims to replace the expression of source image $\mathbf{x_{s}}$ by the expression $\mathbf{t_{e}}$ of target image $\mathbf{x_{t}}$ with neural network $\phi(\cdot,\cdot)$ and generate a new fake image $\mathbf{x_{f}}$. That is $\mathbf{x_{f}} = \phi(\mathbf{x_{s}},\mathbf{t_{e}}).$
\end{definition}

As shown in the expression swap part of Figure~\ref{fig:fourcls}, usually the mouth of the real images are changed. Existing works mainly focus on improving the diversity of input source and video consistency.

Expression swap is also known as face reenactment. The classical examples are ICface \citep{tripathy2020icface} and SVGAN \citep{hyun2021self}.

Face2Face \citep{thies2016face2face} has proposed a three-step procedure. It first uses a global non-rigid model-based bundling approach to reconstruct the shape identity of the target human based on a prerecorded training sequence. Then it uses a transfer function to efficiently exploit deformation transfer in the low-dimensional semantic space. At last, the image-based mouth synthesis approach exploits the best matching mouth shapes offline sample sequence to generate a realistic mouth.

A2V \citep{suwajanakorn2017synthesizing} has used a recurrent neural network to train a model that can map from raw audio features of Obama's weekly address footage to mouth shapes. It is a cross-modal method that leverages the pronunciation features of the target person to synthesize the correct lip shapes for given audio content. It doesn't need an original video as expression-driven material. To match the input audio track, they have synthesized high-quality mouth texture and composited it with proper 3D pose matching to change what he appears to be saying.

Pose-controllable audio-visual system (PC-AVS) \citep{zhou2021pose} is another state-of-the-art cross-modal method. Previous audio-driven talking human face synthesis methods fail to model head pose, one of the key factors for talking faces to look natural. This is because pose information can rarely be inferred from audios. To solve this problem, PC-AVS introduces extra pose source video to compensate only for head motions and successfully disentangle the representations of talking human faces into the spaces of speech content, head pose, and identity respectively.

Previous cross-modal methods only put emphasis on the lip motions and ignore the implicit ones such as head poses and eye blinks that have a weak correlation with the input audio. To model these implicit relationships, face implicit attribute learning generative adversarial network (FACIAL-GAN) \citep{zhang2021facial} integrates the phonetics-aware, context-aware, and identity-aware information to synthesize the 3D face animation with realistic motions of lips, head poses, and eye blinks.

Previous works may lose detailed information of the target leading to a defective output. To solve this problem, MarioNETte \citep{ha2020marionette} has proposed a few-shot face reenactment framework that preserves the information of target identity even in situations where the facial characteristics of the source identity are far from the target. It has also introduced landmark transformation to cope with the varying facial characteristics of different people. 

Interpretable and controllable face reenactment network (ICface) \citep{tripathy2020icface} has proposed a two-stage neural network face animator which can control the pose and expressions of a given face image. The face animator is a data-driven and GAN-based system that is suitable for a large number of identities.

Self-supervised video GAN (SVGAN) \citep{hyun2021self} first puts emphasis on exploiting the discriminator of the GAN. They hypothesize two prominent constraints for realistic videos: consistency of appearance and coherency of motion. With these constraints, GANs are more likely to generate realistic videos. In other words, they have well defined what constraints should synthesized videos satisfy first.

\cite{wang2021one} propose a one-shot neural talking-head synthesis approach. The method uses unsupervised learning to decompose for key features of an image: appearance feature, canonical keypoint, head pose, expression deformation. With the appearance feature and canonical keypoint of the source image, and synthesized with the head pose and expression deformation, a new fake image can be created. This work clearly disassembles the face information and reasonably exploits them.

Most of the DeepFake detection methods did not take expression swap as the main detection objective. In our opinion, there are several reasons. As we can see from the previous description, expression swap has a similar technique to identity swap. Thus most of the detection methods are not specifically designed for them and only a few detection methods consider detecting expression swaps. On the other hand, it usually needs the coordination of audio to achieve a better display effect in the expression swap. Only the detection methods which simultaneously take images and audio into account are designed for this problem. 
The swapped expression strongly depends on the source video or image. The audio-video coordination opens the door for detection algorithm to tackle this problem from multiple angles, reducing the difficulty of this problem. Therefore, as we mainly investigate the DeepFake generation methods that are mentioned by the DeepFake detection methods, we take expression swap as an extension of identity swap in the survey.

\subsubsection{Technical Evolution of Expression Swap}\label{sec:tec_evo_expression_swap}
In contrast to identity swap, expression swap is to replace the features of the mouth in the source image and produce a new face with the same identity but a different expression.

The early work Face2Face \citep{thies2016face2face} designs a three-step procedure to swap expressions without deep learning. Specifically, Face2Face first strips the identity information from the source identity. Then, it transfers the source identity's expression to the target one. Finally, it synthesizes a realistic target mouth region. The whole pipeline is reasonable but requires complex 3D face models and considerable efforts to capture all the subtle movements in the face.

In addition, to swap the expression according to the given images, recent works also explore how to synthesize expressions from audios. For example, A2V \citep{suwajanakorn2017synthesizing} successfully synthesized fake videos of Obama (\ie, the 44th president of the United States) according to the given audio. To this end, \cite{suwajanakorn2017synthesizing} use recurrent neural networks to map audios to the sparse shape of the mouth (\ie, 18 lip fiducials). Then, they generate photo-realistic mouth texture based on the generated lip fiducials. As a result, the expression swap based on the audios is achieved.

To bypass the explicit 3D model fitting, a straightforward way is to learn a deep model implicitly via large-scale data. However, there are two problems. First, it is hard to collect expression and pose representation that is independent of the identity feature. Second, such an implicit model usually lacks interpretability and does not easily allow
Hence, it is difficult to synthesize diverse face attributes from the other faces.
To solve these problems, ICFA \citep{tripathy2020icface} proposes to use action units (AUs) \citep{friesen1978facial} to represent the emotions. The AUs represent the activations of 17 facial muscles and each combination of them can produce different facial expressions.
The advantages of AU-based expression representation are as follows. First, it is a relatively straightforward and flexible way to extract expressions from any facial image. Second, this representation is fairly independent of the identity-specific characteristics of the face. As a result, the first problem can be solved. 
To swap the expression to the target face, ICface eliminates the expression of the input face first, which is done by mapping the input image to a neutral state representing zero AU values.
Then, it uses a conditional GAN to take the neutral image and the previous facial attribute vector (\ie, AUs) as input to generate expression, which solves the second problem. 
The model is also interpretable in that the facial attribute vector can be manually defined.

Previous works focus on the expression swap of images. However, the expression swap on videos needs to maintain the consistency of the face across frames, which is much more difficult than swapping on images. To achieve a more realistic expression swap on video, \cite{hyun2021self} propose the SVGAN that clearly defines two constraints (\ie, appearance contrastive loss \& temporal structure loss) that should be satisfied in the video synthesis. 
The appearance contrastive loss makes the discriminator learn the representations of appearance which is invariant throughout time in videos. On the other hand, temporal structure loss forces the discriminator to figure out whether the video is coherent or not in temporal ordering.
These two losses ensure the appearance of consistency and motion coherency in videos. They achieve a good effect by simply adding these two constraints to the discriminator.

Overall, the development of expression swap techniques follows the requirements of real-world scenarios, such as the audios as guidance, controllable expression, and temporal consistency across video frames). For the expression swap, we think the improvements of the diversity of input sources, controllable expression, and video consistency are the major technical characteristics that have been evolved throughout the years, and we highlight seminal works A2V, ICface, SVGAN in the bottom right panel of Figure \ref{fig:fish_bone_generation}, showcasing the improvements in different angles.

\subsection{Real Dataset}\label{Real Dataset}
Real datasets are required for supervised training of DeepFake detectors. Here we introduce popular real datasets. Please note that the following datasets are real image datasets for that the independent real video dataset is infrequent and most of the real face video datasets used in generating the fake datasets are collected by them from YouTube or shot by them with the actors invited by them. We record the information in Table~\ref{tab:fake_database} as introduced in Section~\ref{Fake Dataset}. The datasets are introduced in ascending order by their release dates.

CASIA-WebFace \citep{yi2014learning} has proposed a semi-automatic way to collect a lot of face images from the Internet. The dataset contains 10,575 subjects and 494,414 images, which is both diverse and large.

CelebA \citep{liu2015deep} is constructed by labeling images selected from a famous face dataset: CelebFaces \citep{sun2013hybrid}. CelebA contains ten thousand identities, each of which has twenty images, a total of 200,000 images. Each image in CelebA is annotated with forty face attributes and five key points by a professional labeling company, which is extremely abundant and useful.

VGGFace \citep{parkhi2015deep} consists of 2,622 identities with 2.6 million images. The famous VGGNet \citep{parkhi2015deep} is trained by this dataset.

MegaFace \citep{kemelmacher2016megaface} includes more than 690K different individuals with one million photos. They have established MegaFace challenge which evaluates how face recognition algorithms perform under the perturbation of a very large number of ``distractors'' (\ie, individuals that are not in the probe set).

LSUN \citep{yu2015lsun} is the only non-face real dataset discussed by us, for that it is widely used by fake generation methods. It contains around one million labeled images for each of 10 scene categories and 20 object categories.

To develop face recognition technologies, Microsoft Celeb (MS-Celeb-1M) \citep{guo2016ms} has collected 10 million face images of nearly 100,000 individuals from the Internet.

VGGFace2 \citep{cao2018vggface2} contains 3.31 million images of 9,131 subjects. Images are harvested from the Internet and have large variations in pose, age, illumination, ethnicity, and profession.

\cite{karras2019style} has collected a new high-resolution dataset of human faces, Flickr-Faces-HQ (FFHQ). It contains 70,000 high-quality images at $1024\times1024$ resolution. The dataset includes vastly more variation than CelebA-HQ \citep{karras2017progressive} in terms of age, ethnicity, and image background. It also has much more accessories such as eyeglasses, sunglasses, hats, \etc.

All the above real face datasets can generate DeepFake dataset with three categories (\ie, entire face synthesis, identity swap, expression swap). The datasets which have the abundant label information, especially face attributes are superior ones for attribute manipulation. For example, CelebA and CelebA-HQ are the most usually used real face dataset to generate attribute manipulation images.

\subsection{Fake Dataset}\label{Fake Dataset}
Fake image/video datasets are an important benchmark for testing the performance of existing DeepFake generation methods. With the development of fake generation methods, the quality and fidelity of fake datasets are getting higher and higher. Here we introduce popular fake datasets. The datasets are introduced in ascending order by their release dates.

The UADFV dataset \citep{li2018ictu} consists of 98 videos, with 49 real videos from YouTube and 49 synthesized videos, which are made using the FakeAPP \citep{FaceApp}.
 
The DeepFake-TIMIT dataset \citep{korshunov2018deepfakes} consists of 620 DeepFake videos of 32 subjects. In DeepFake-TIMIT, each subject has 20 DeepFake videos. 10 videos are of size $64 \times 64$ while the other 10 videos are of size $128 \times 128$. The synthesized videos are generated using faceswap-GAN \citep{Faceswap-GAN}. 

DeepFakes Detection Challenge Preview (DFDC Preview) \citep{dolhansky2019deepfake} dataset consisting of 5K videos with two facial modification algorithms. The actors are of different gender, skin-tone, age, \etc. They record videos with arbitrary backgrounds thus bringing visual variability.

Google DFD \citep{dufour2019contributing} contains over 3,000 manipulated videos from 28 actors in various scenes. The videos are generated from hundreds of real videos by using publicly available DeepFake generation methods.

FaceForensics++ \citep{rossler2019FaceForensics++} is a famous fake video dataset consisting of 1,000 original video sequences that have been manipulated with four automated face manipulation methods: DeepFakes, Face2Face, FaceSwap, and NeuralTextures. The videos are generated from 977 trackable YouTube videos. The people in most of the videos are frontal faces. 

Celeb-DF \citep{li2020celeb} has presented a large-scale challenging DeepFake video dataset, which contains 5,639 high-quality DeepFake videos of celebrities generated using an improved synthesis process.

Diverse Fake Face Dataset (DFFD) \citep{dang2020detection} has collected a large-scale dataset that contains numerous types of facial forgeries. Among all images and video frames, 47.7\% are from male subjects, 52.3\% are from females, and the majority of samples are in the age of 21-50 years. They utilize FFHQ, CelebA, and source frames from FaceForensics++ as the real face samples. For facial identity and expression swap, they use all the video clips from FaceForensics++. They have adopted two methods FaceAPP \citep{FaceApp} and StarGAN \citep{choi2018stargan} to generate attribute manipulated images. Recent works such as PGGAN \citep{karras2017progressive} and StyleGAN \citep{karras2017progressive} are used for face image synthesis.

FakeCatcher \citep{ciftci2020fakecatcher} has collected over 140 online videos, up to 30GB. Unlike most of the fake datasets, it includes ``in the wild'' videos, independent of the generative model, resolution, compression, content, and context.

iFakeFaceDB \citep{neves2020ganprintr} is a fake image dataset for the study of synthetic face manipulation detection. It contains about 87,000 synthetic face images generated by the StyleGAN model \citep{karras2019style} and transformed with the GANprintR \citep{neves2020ganprintr} approach. All images are of size $224 \times 224$.

Facebook has constructed an extremely large face video dataset to enable the training of detection models. They organized a famous DeepFake Detection Challenge (DFDC) \citep{dolhansky2020deepfake} Kaggle competition. The DFDC is a publicly-available face swap video dataset, with 128,514 videos, over 100,000 total clips sourced from 3,426 actors. They use various face swap methods with two kinds of augmentations (distractor and augmenter). Distractor means overlaying various kinds of objects (including images, shapes, and text) onto a video while augmenter means applying geometric and color transforms, frame rate changes, \etc., onto a video.

\cite{dong2020identity} have built a large-scale DeepFake detection dataset ``Vox-DeepFake'', which has a total of 2 million real videos and fake videos. Compared to existing datasets, it has better quality and diversity in terms of identities and video content. Furthermore, they have supplied the explicit reference identity information for each real/fake video, which is more informative than previous datasets.

DeeperForensics-1.0 \citep{jiang2020deeperforensics} has represented the largest face forgery detection dataset by far, with 60,000 videos constituted by a total of 17.6 million frames. There are 50,000 original collected videos and 10,000 manipulated videos including 100 actors. In particular, 55 of them are males and 45 of them are females. The actors have four typical skin tones: white, black, yellow, brown. All faces are clean without glasses or decorations. Unlike previous data collection in the wild, they build a professional indoor environment for a more controllable data collection and add a mixture of perturbations to videos making the dataset better imitate real-world scenarios. The perturbation is added by systematically applying seven types of distortions (compression, blurry, noise, \etc) to the fake videos at five intensity levels. They also propose DeepFake variational auto-Encoder (DF-VAE) as a new end-to-end face swap method.

As the previous fake datasets were filmed with limited actors in limited scenes, and the fake videos are generated with a few popular DeepFake software, the diversity of the dataset is scarce. In contrast, wild DeepFake can have many persons in one scene, and the scenes vary significantly. Meanwhile, wild DeepFake may even be generated by combinations of DeepFake software. Furthermore, the fake videos in the fake dataset may not be well processed for that the face regions in them often have perceptible distortions. To provide a more realistic DeepFake dataset, \cite{zi2020wilddeepfake} collect their dataset WildDeepfake, which contains 7,314 face sequences extracted from 707 DeepFake videos collected completely from the internet. WildDeepfake is able to test the effectiveness of DeepFake detectors against real-world DeepFake.

ForgeryNet \citep{he2021forgerynet} has tried to build an extremely large face forgery dataset designed for four tasks: image forgery classification, spatial forgery localization, video forgery classification, temporal forgery localization. This fake dataset contains 2.9 million images and 221,247 videos, which is the largest one among the fake datasets. It also provides 15 manipulation approaches with more than 36 mix-perturbations on over 5,400 subjects. The dataset surpasses the other fake datasets both in scale and diversity.

\cite{pu2021deepfake} also pay attention to whether the existing detection methods can effectively adapt to the DeepFake videos in the wild. They build a fake dataset DF-W, which contains 1,869 fake videos collected from YouTube, \cite{Bilibili} and Reddit.

Most of the fake datasets put emphasis on collecting videos in which only exist one manipulated person. However, the existing detection methods fail to detect the multi-person videos effectively. Thus, \cite{zhou2021face} build a large dataset FFIW$_{10K}$, which comprises 10,000 high-quality fake videos and real videos, with an average of three human faces in each frame. This fake dataset is more challenging and points out the future research direction of the detection methods.

Similar to FFIW$_{10K}$, OpenForensics \citep{le2021openforensics} also take care of the capability of the DeepFake detection methods on multi-person images. It contains 45,473 real images and 70,325 fake images, a total of 115,325 images with 334,126 faces in the images. It is worth mentioning that OpenForensics not only introduce multi-face forgery detection task but also propose segmentation in-the-wild task. For these two tasks, they provide face-wise rich annotations such as forgery category, bounding box, segmentation mask, forgery boundary, and general facial landmarks. The abundant annotations make OpenForensics the first dataset that supports the DeepFake localization task, which is meaningful for multi-media forgery forensics. Furthermore, OpenForensics can also be used for general object detection and segmentation tasks, which shows its versatility.

As shown in Table \ref{tab:paper_information}, according to the citation metric, FaceForensics and FaceForensics++ are the datasets with the highest citations. According to the Elo rating score, DFDC is the dataset with the highest Elo score. As shown in Figure \ref{fig:battle}, FaceForensics++ is the most commonly detected fake dataset for facial appearance swapping detection task while PGGAN is the most commonly detected DeepFake technique for the entire face synthesis detection task. We suggest the researchers put more emphasis on these datasets.

\begin{table*}[!htbp]
\centering
\caption{Top-5 generation methods based on battleground (top) and Elo rating \citep{Elo} (bottom) separately. It is worth noting that the ``SAN'' method is special because it is inherently a super-resolution method. As can be observed, the latest DeepFake generation methods are able to produce highly realistic facial images that are immensely hard to tell apart from real ones using human perception.}
\label{tab:visualize}
\begin{adjustbox}{width=\linewidth,center}
    \begin{tabular}{|c|c|c|c|c|}
    \hline 
    \includegraphics[height=0.22\linewidth]{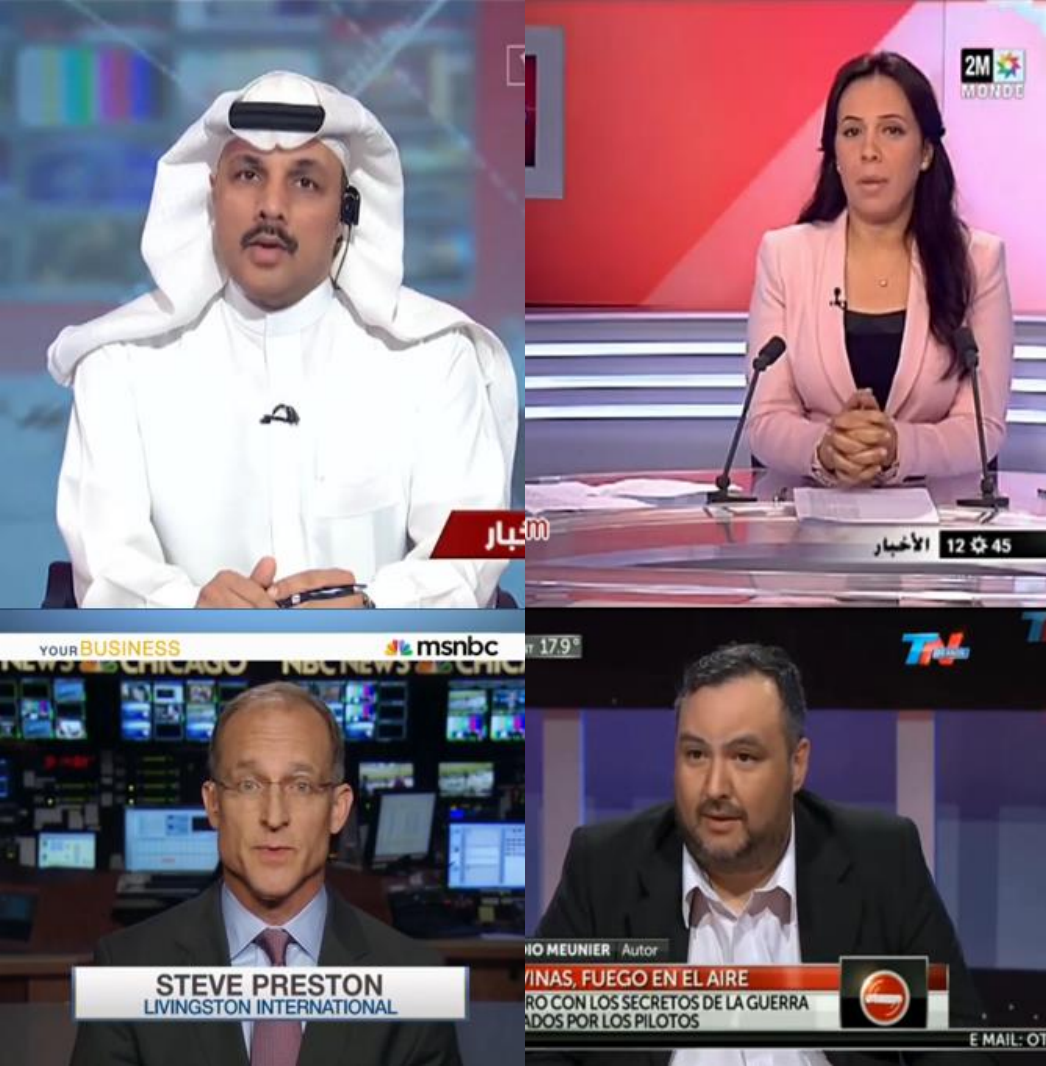} &  \includegraphics[height=0.22\linewidth]{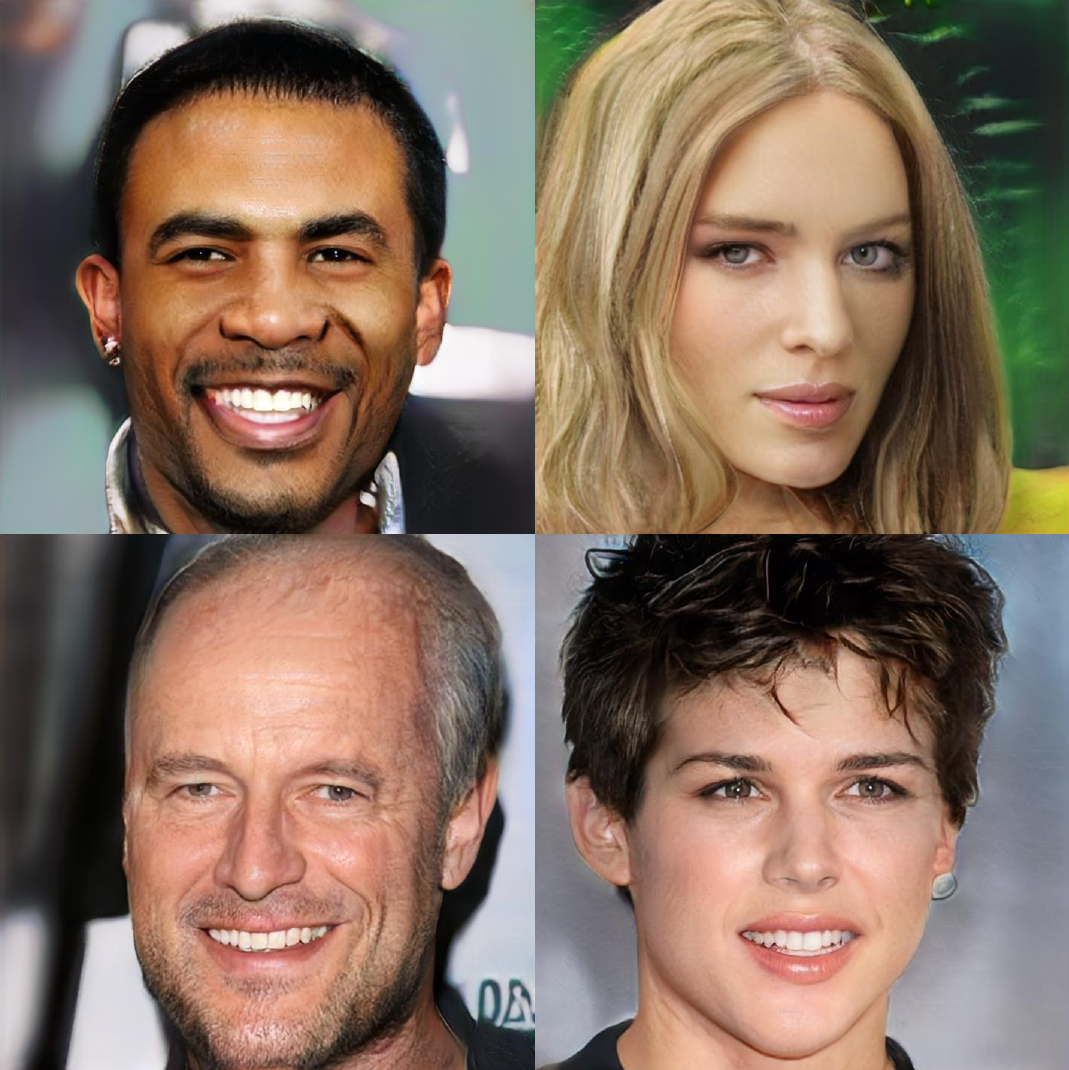} & \includegraphics[height=0.22\linewidth]{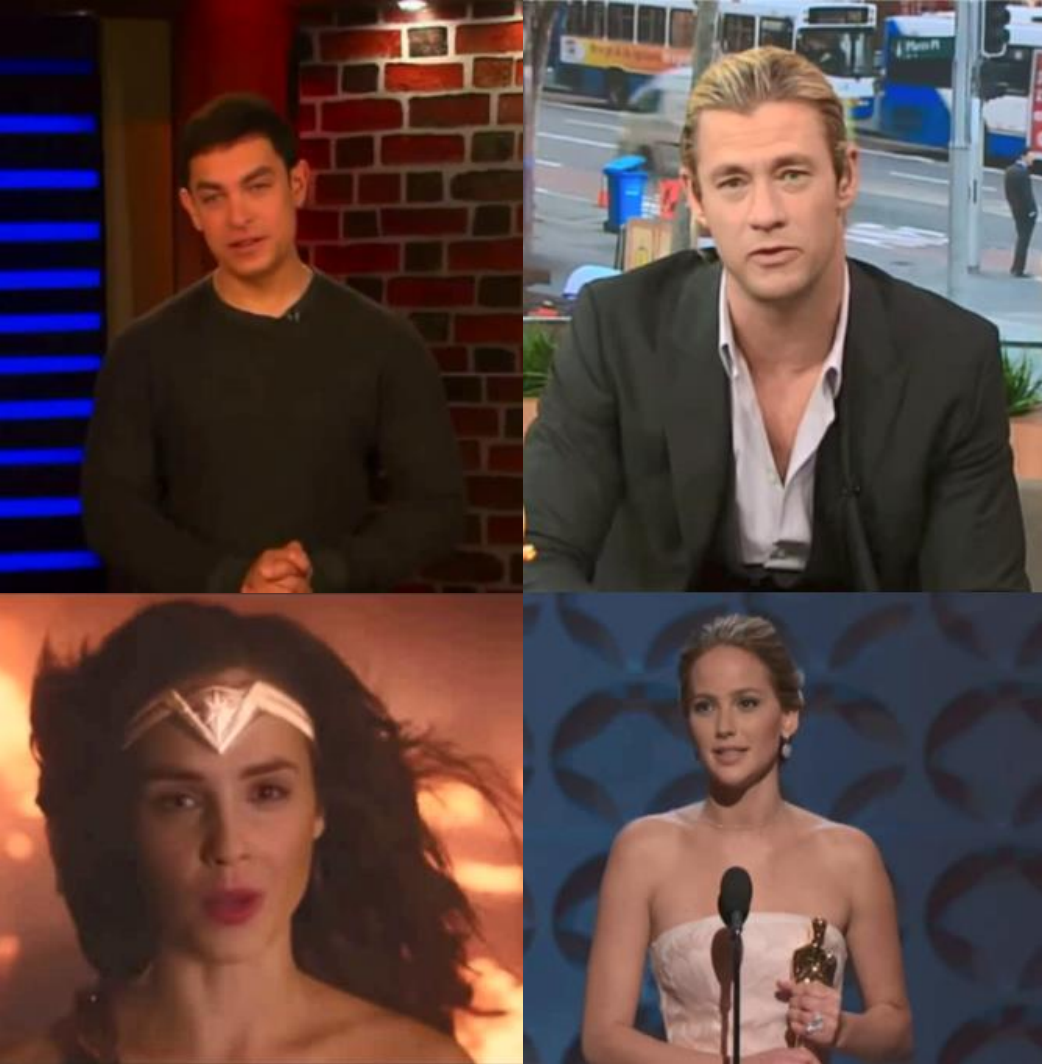} & \includegraphics[height=0.22\linewidth]{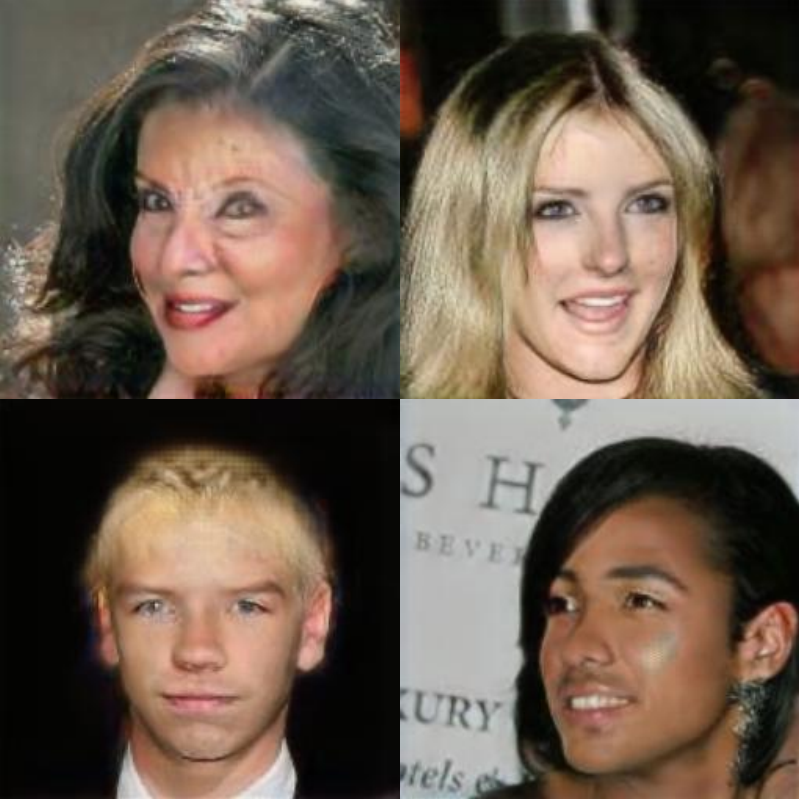} & \includegraphics[height=0.22\linewidth]{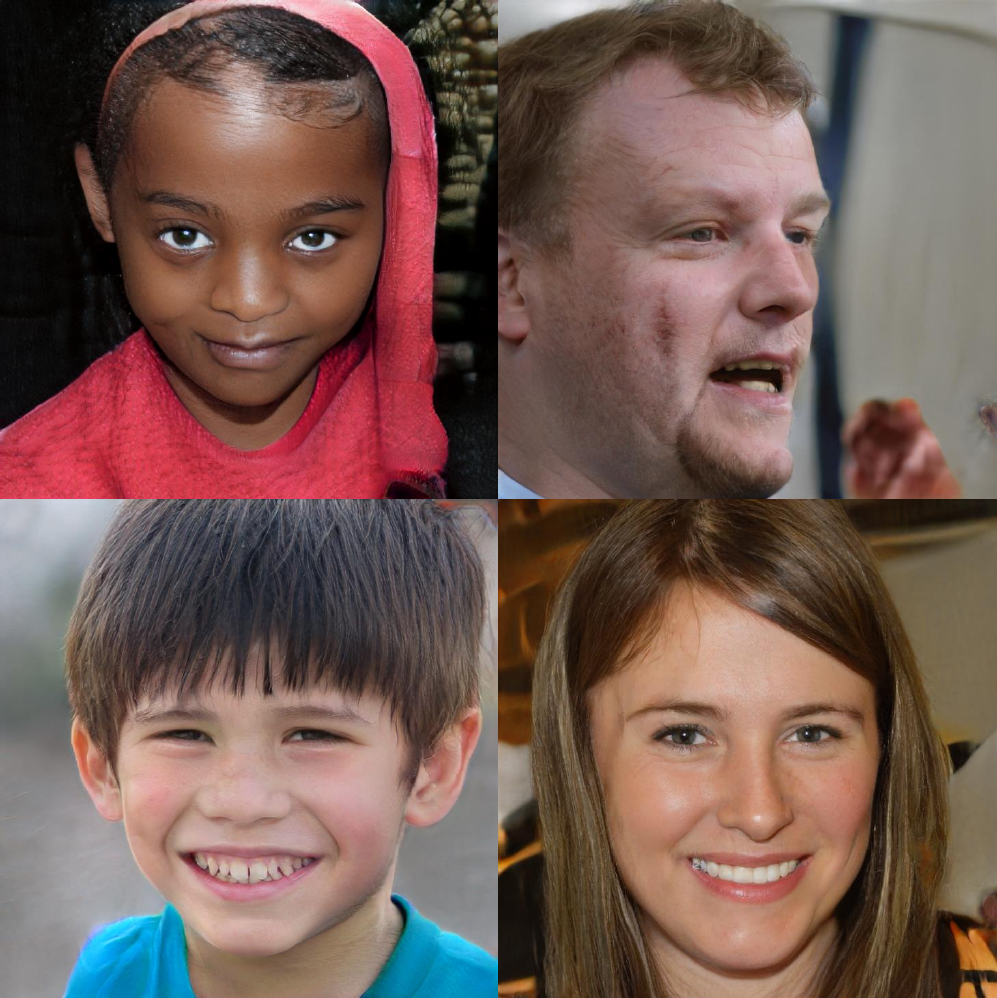} \tabularnewline
    \hline 
    \makecell[c]{FaceForensics++ \\ \citep{rossler2019FaceForensics++}}& \makecell[c]{PGGAN \\ \citep{karras2017progressive}} &  \makecell[c]{Celeb-DF \\ \citep{li2020celeb}} & \makecell[c]{StarGAN \\ \citep{miyato2018spectral}} & \makecell[c]{StyleGAN \\ \citep{karras2019style}} \tabularnewline
    \hline 
    \includegraphics[height=0.22\linewidth]{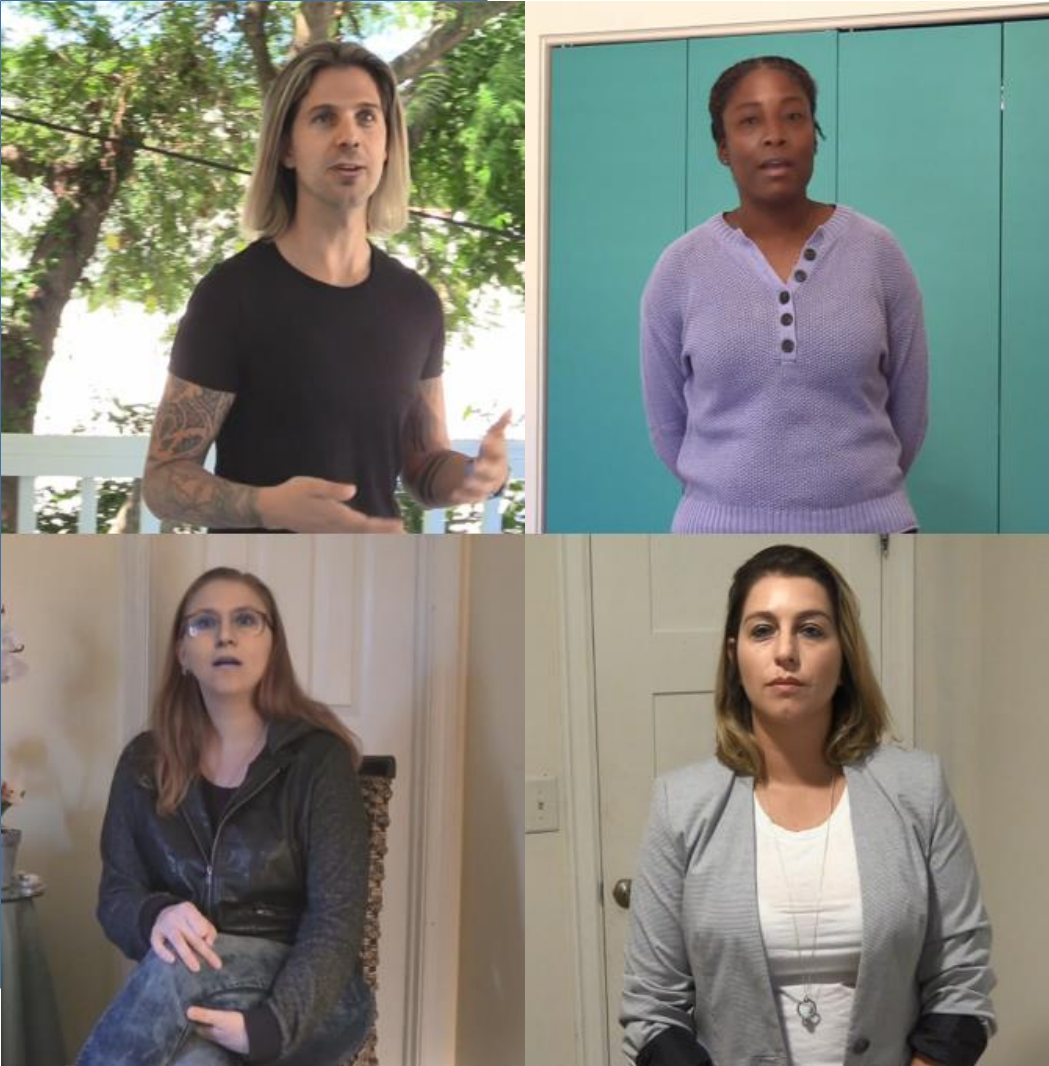} &  \includegraphics[height=0.22\linewidth]{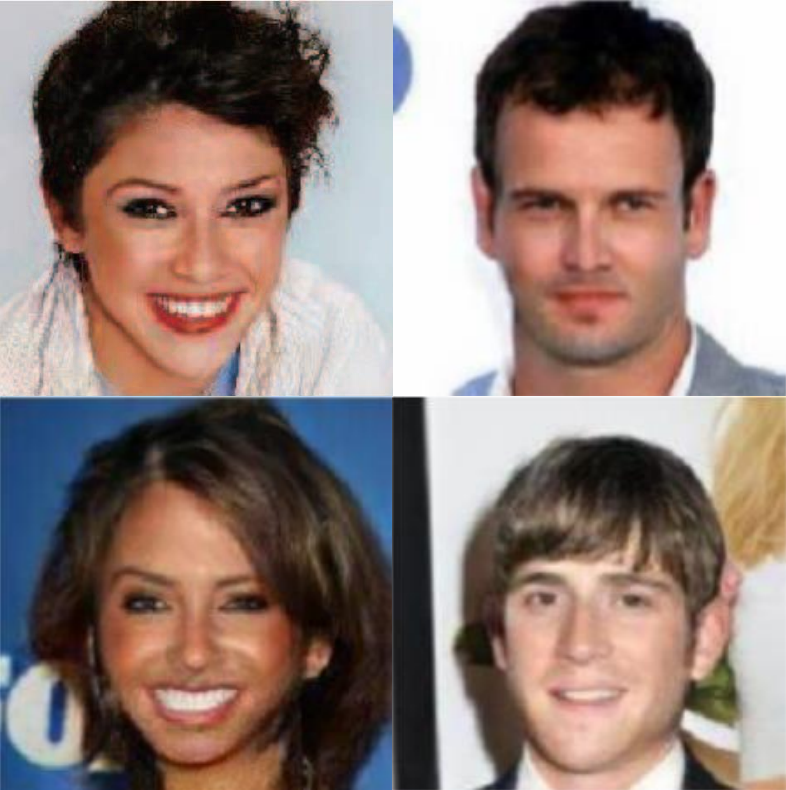} & \includegraphics[height=0.22\linewidth]{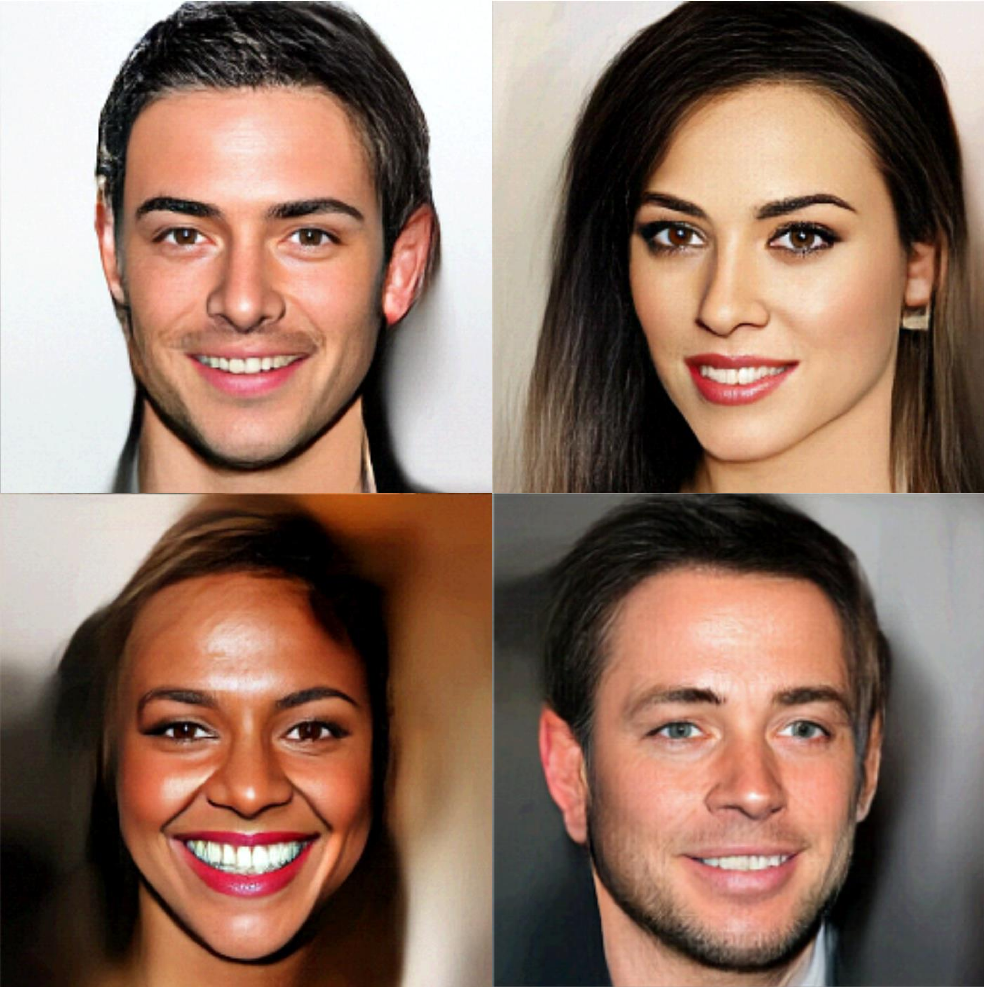} & \includegraphics[height=0.22\linewidth]{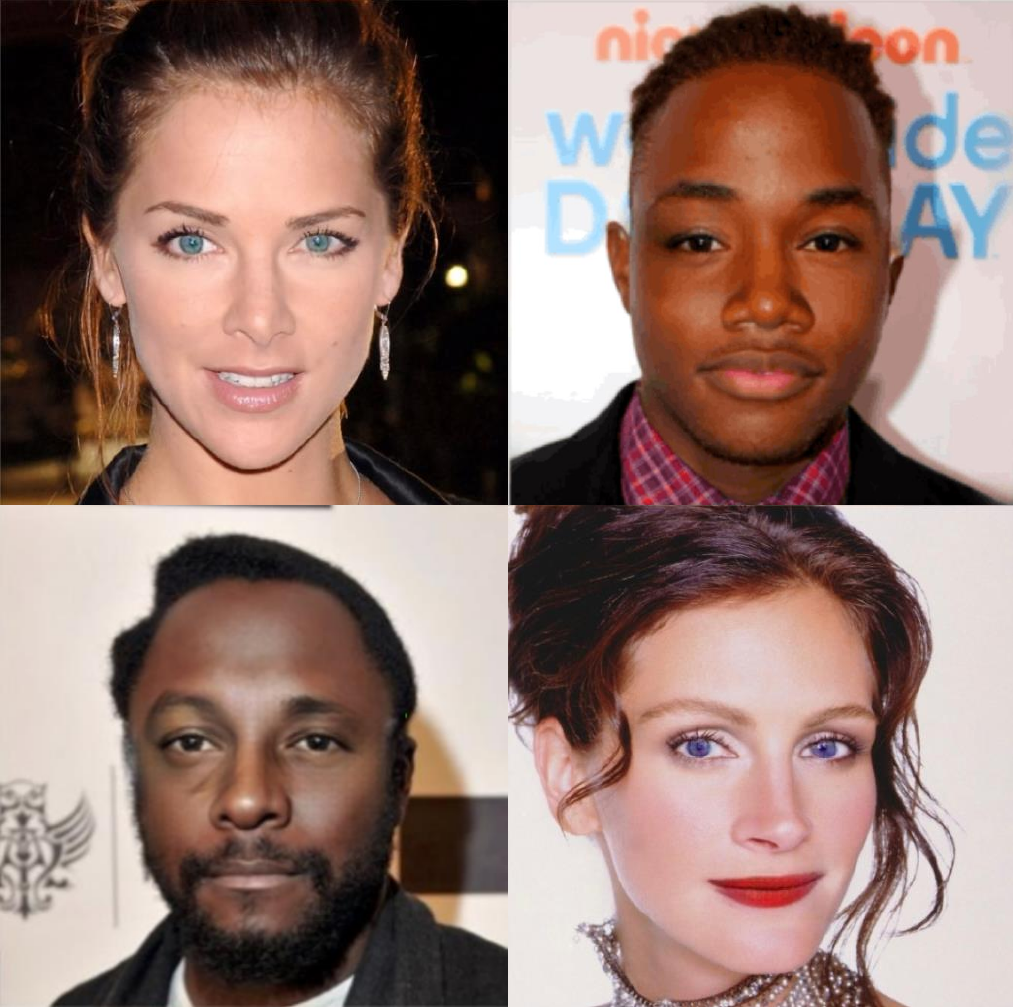} & \includegraphics[height=0.22\linewidth]{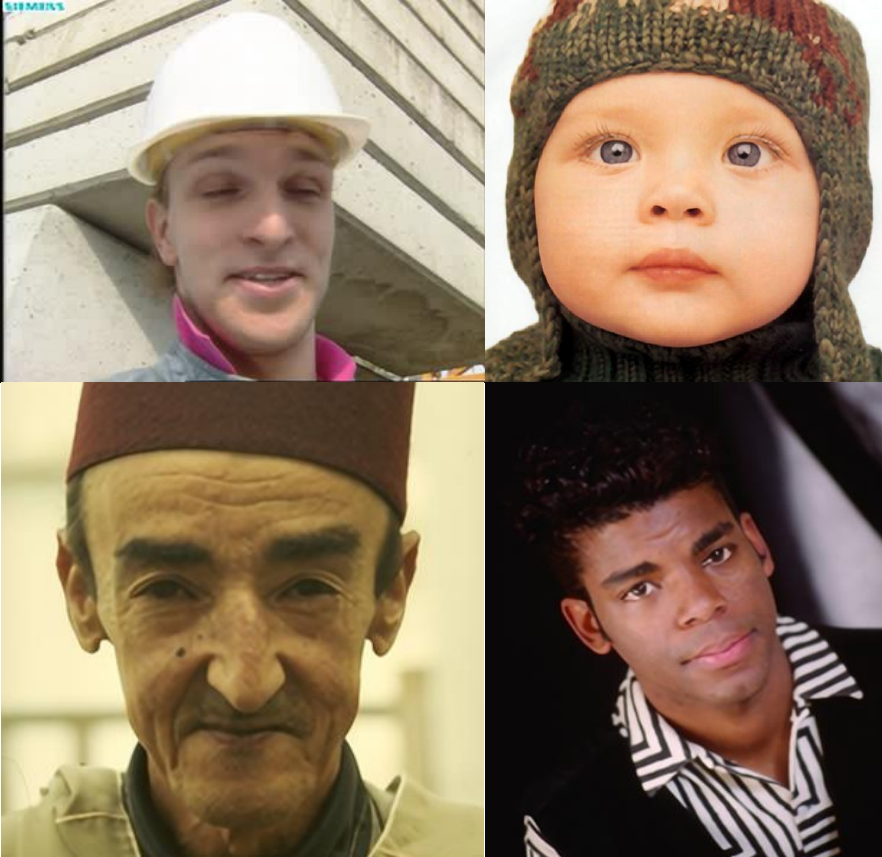} \tabularnewline
    \hline 
    \makecell[c]{DFDC \\ \citep{dolhansky2020deepfake}}& \makecell[c]{GDWCT \\ \citep{cho2019image}} &  \makecell[c]{Glow \\ \citep{kingma2018glow}}& \makecell[c]{SC-FEGAN \\ \citep{jo2019sc}}& \makecell[c]{SAN\\ \citep{dai2019second}} \tabularnewline
    \hline 
    \end{tabular}
\end{adjustbox}
\end{table*}

\begin{table*}[!htbp]
\centering
\caption{Some lower-quality image examples (that show more visible artifacts) generated by the DeepFake generation methods mentioned in Table \ref{tab:visualize}, with an exception of the SAN \citep{dai2019second} method, which is a super-resolution technique and super-resolution is not easy to produce low-quality images. Thus we show the same images as in Table \ref{tab:visualize}.}
\label{tab:visualize_bad}
\begin{adjustbox}{width=\linewidth,center}
    \begin{tabular}{|c|c|c|c|c|}
    \hline 
    \includegraphics[height=0.22\linewidth]{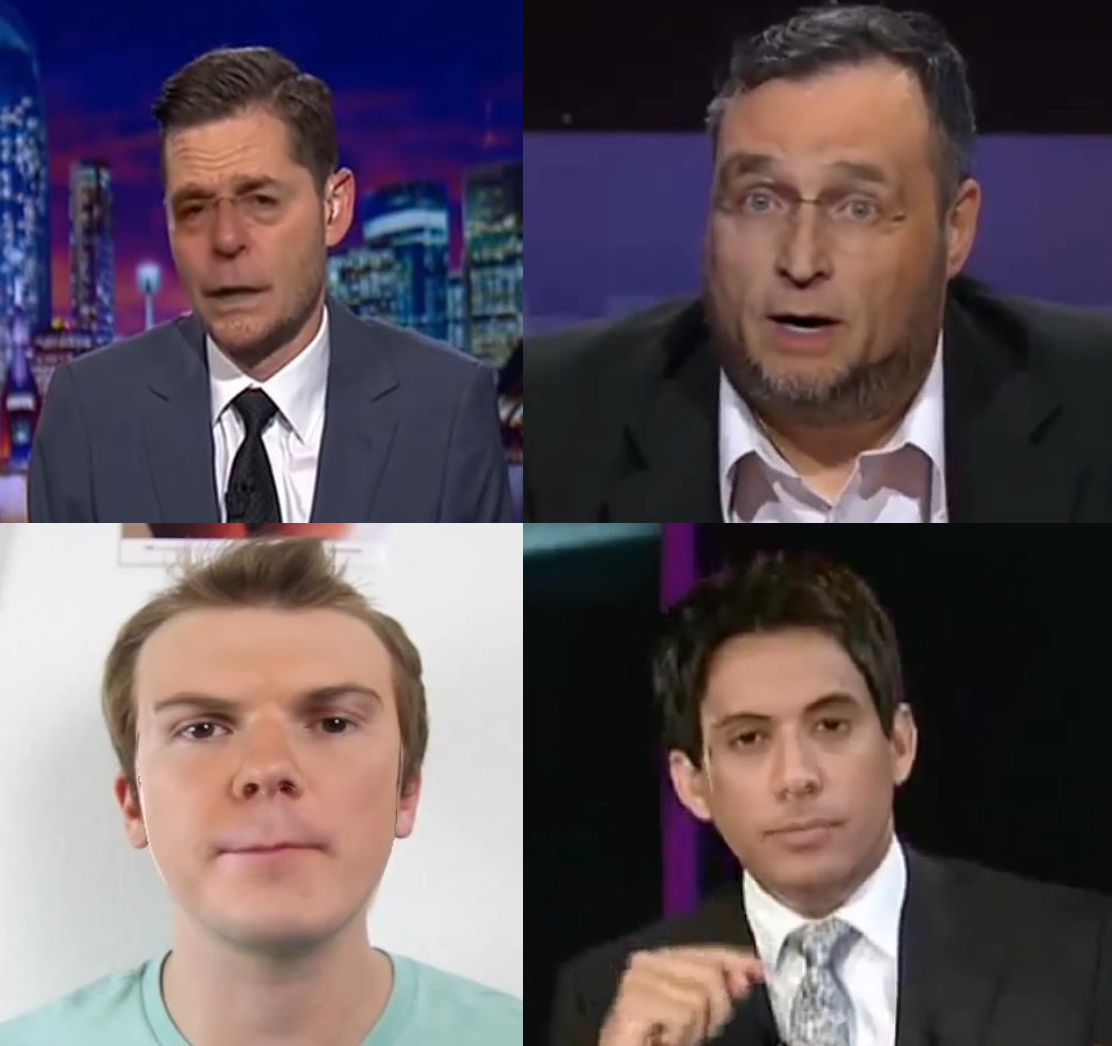} &  \includegraphics[height=0.22\linewidth]{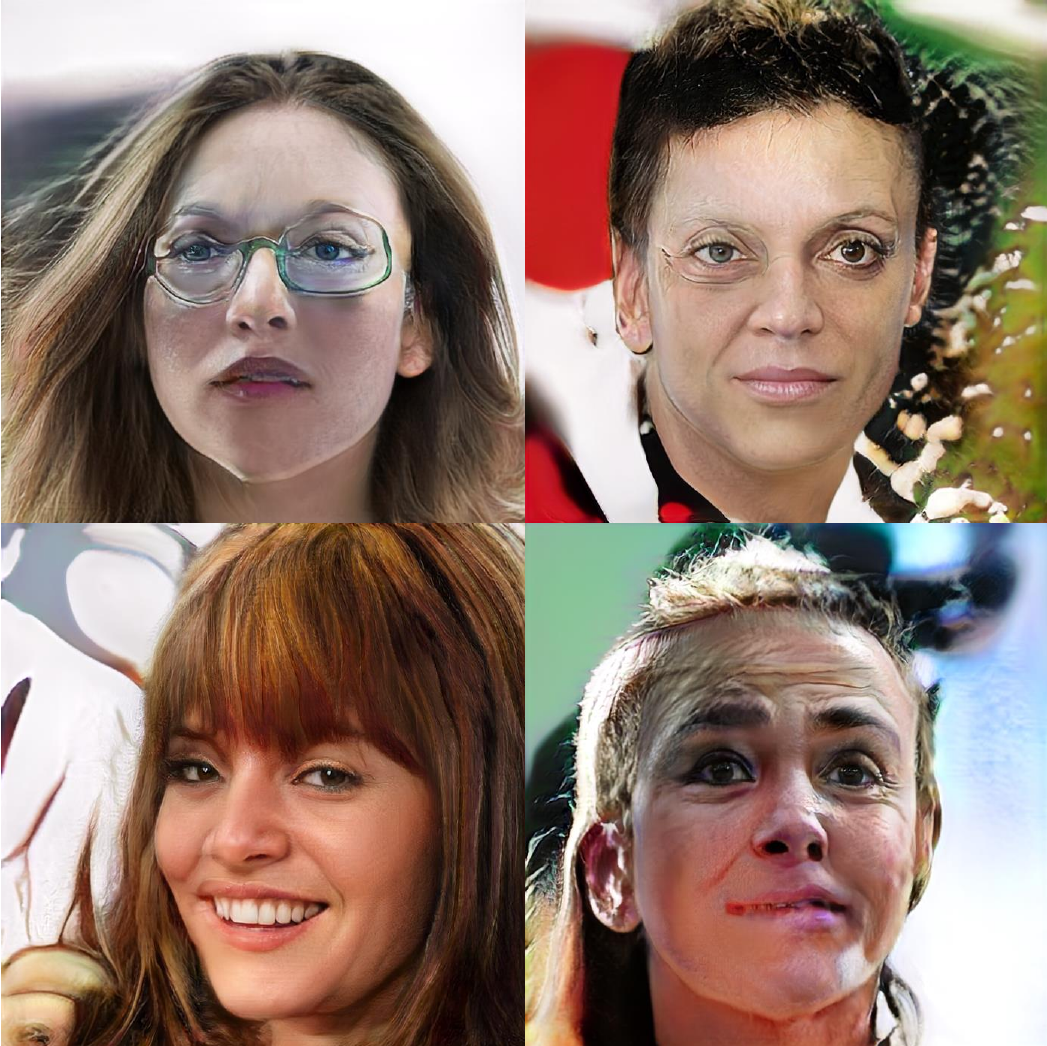} & \includegraphics[height=0.22\linewidth]{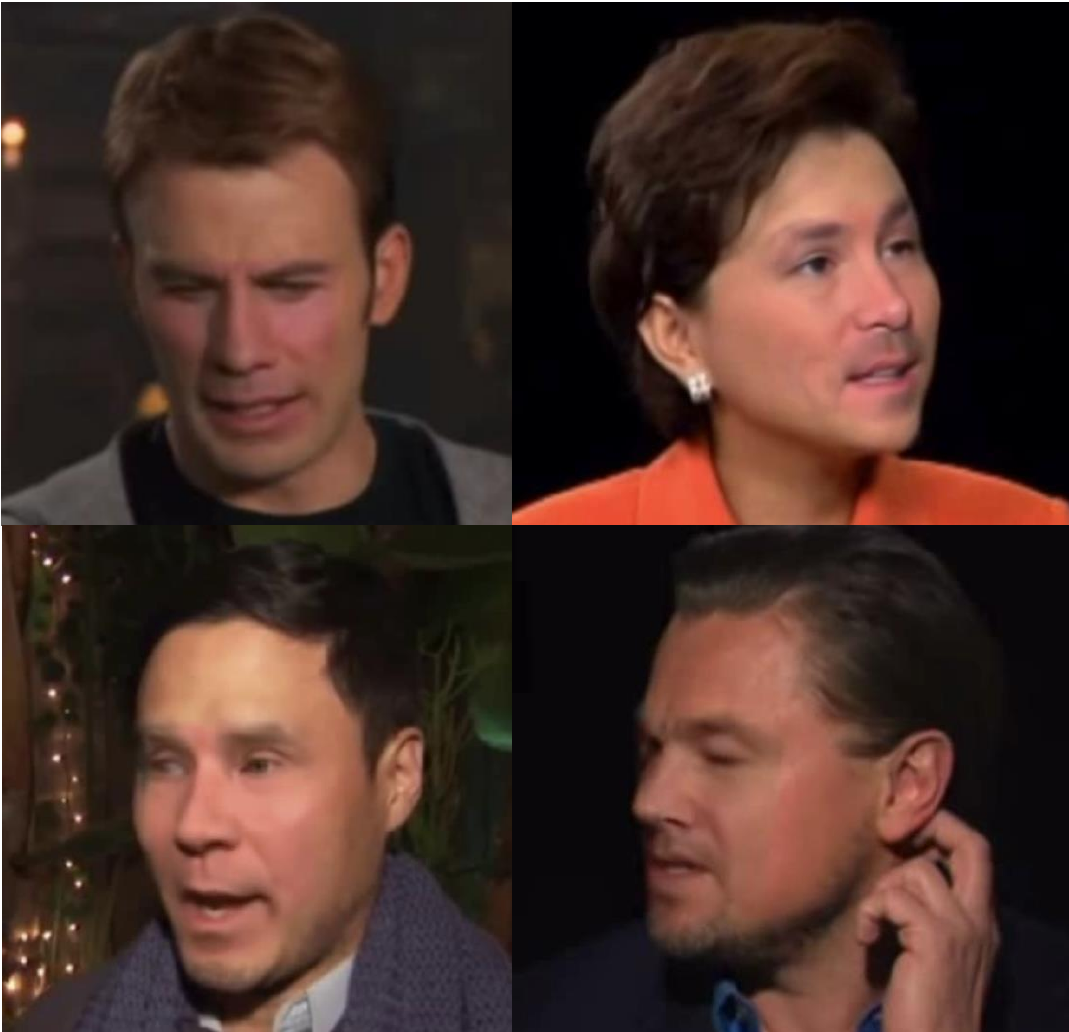} & \includegraphics[height=0.22\linewidth]{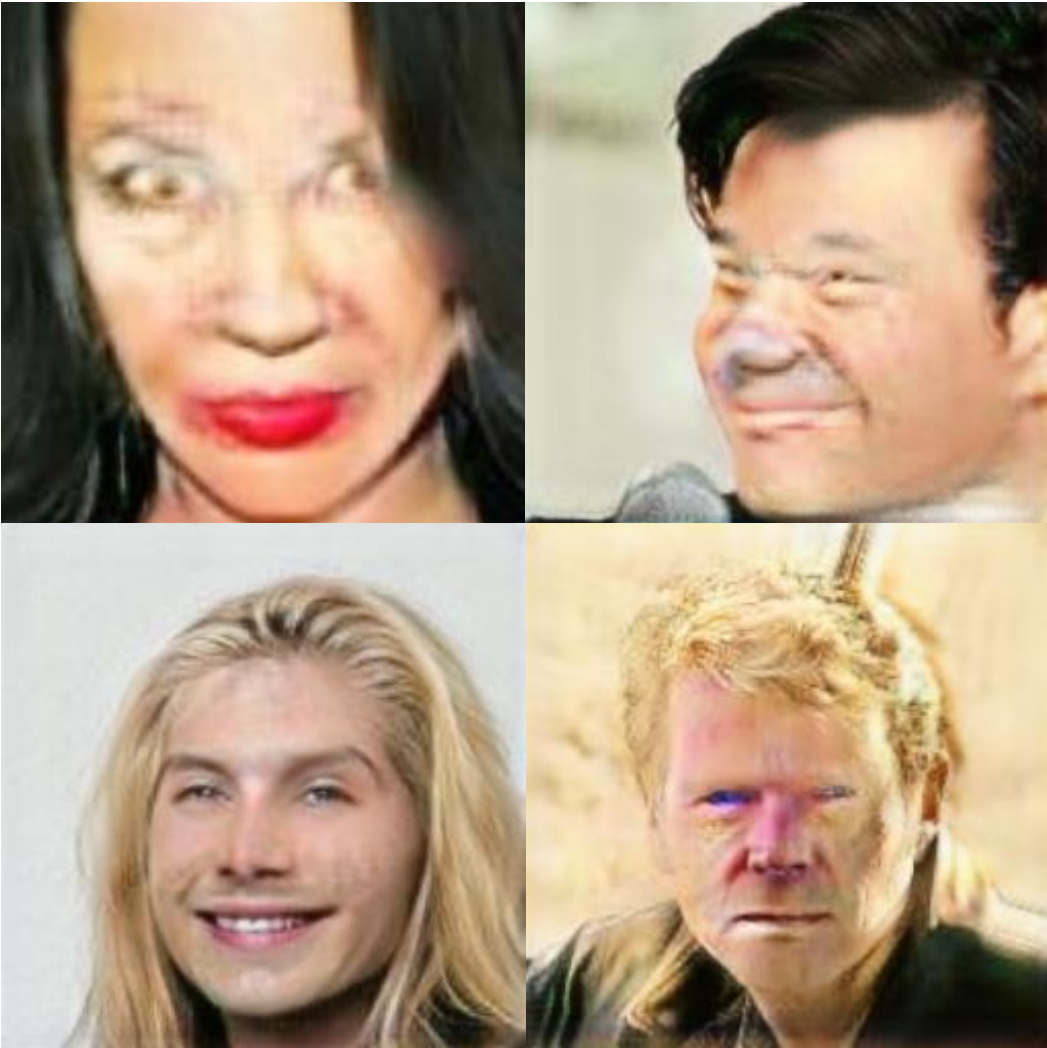} & \includegraphics[height=0.22\linewidth]{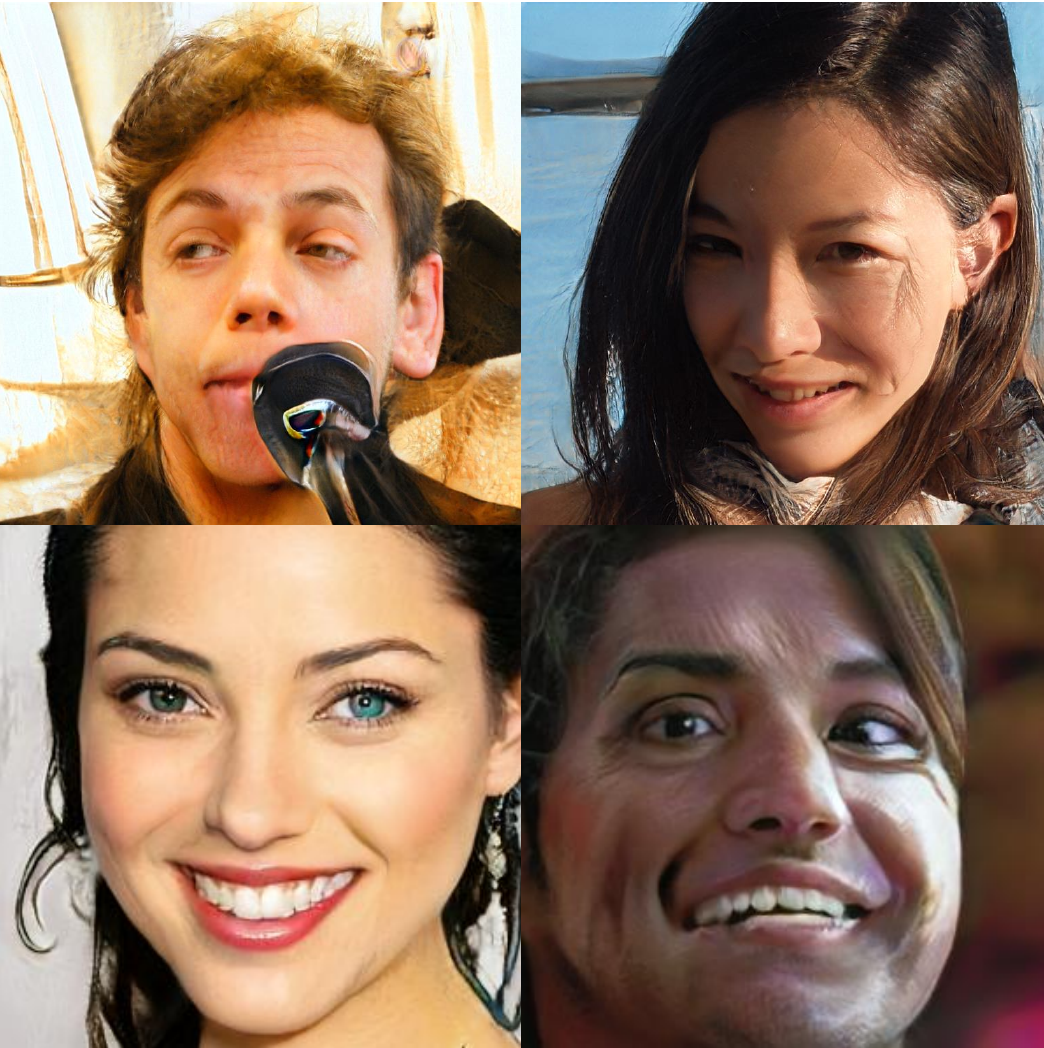} \tabularnewline
    \hline 
    \makecell[c]{FaceForensics++ \\ \citep{rossler2019FaceForensics++}}& \makecell[c]{PGGAN \\ \citep{karras2017progressive}} &  \makecell[c]{Celeb-DF \\ \citep{li2020celeb}} & \makecell[c]{StarGAN \\ \citep{miyato2018spectral}} & \makecell[c]{StyleGAN \\ \citep{karras2019style}} \tabularnewline
    \hline 
    \includegraphics[height=0.22\linewidth]{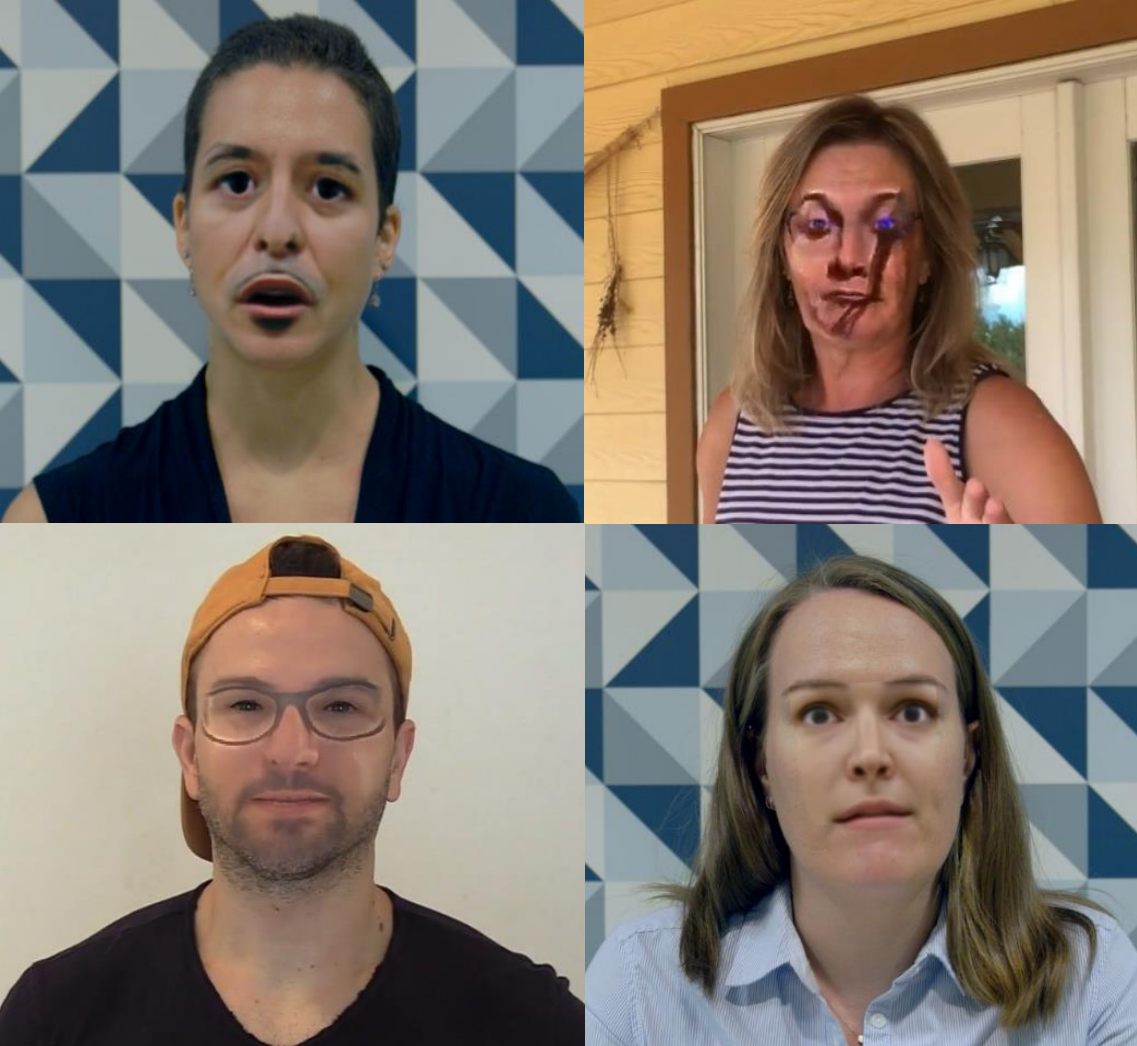} &  \includegraphics[height=0.22\linewidth]{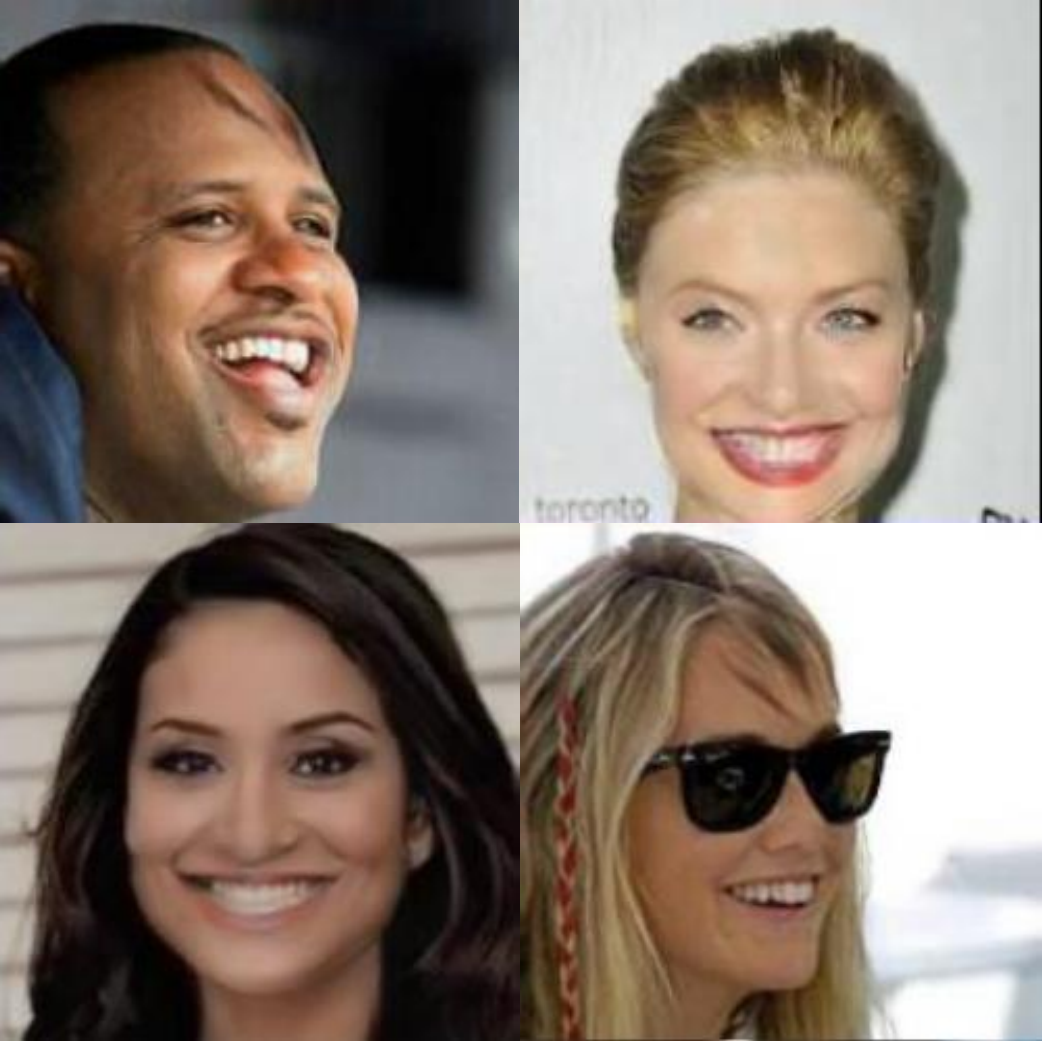} & \includegraphics[height=0.22\linewidth]{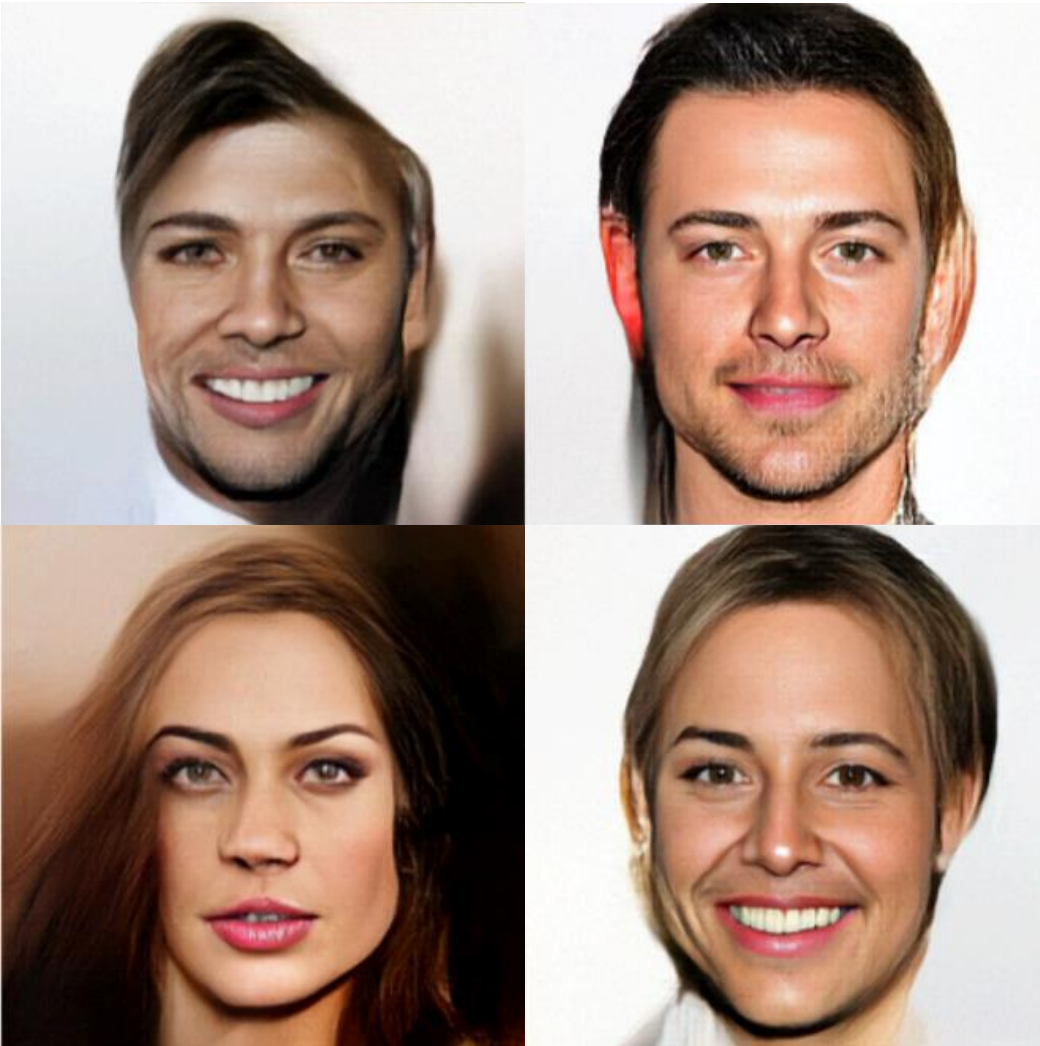} & \includegraphics[height=0.22\linewidth]{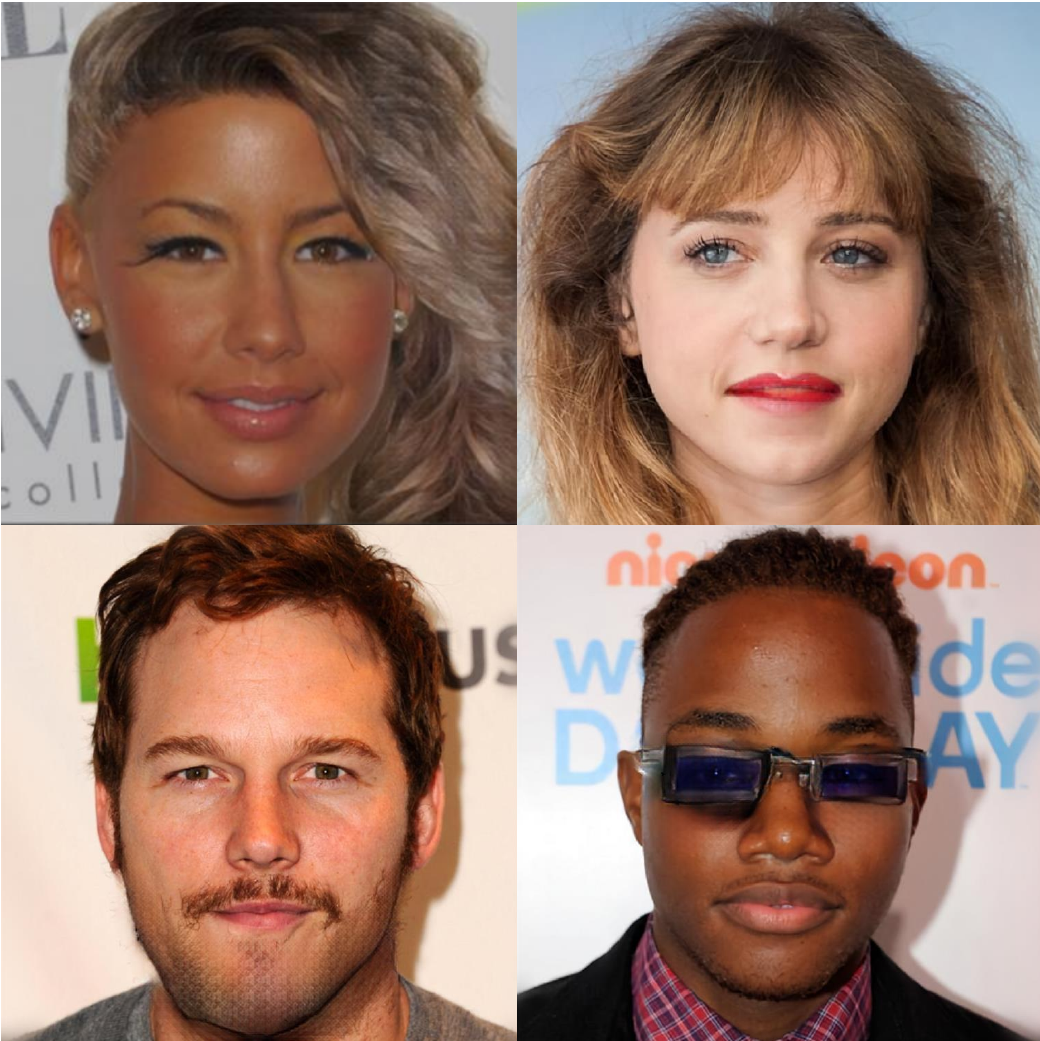} & \includegraphics[height=0.22\linewidth]{image/SAN_cropped.pdf} \tabularnewline
    \hline 
    \makecell[c]{DFDC \\ \citep{dolhansky2020deepfake}}& \makecell[c]{GDWCT \\ \citep{cho2019image}} &  \makecell[c]{Glow \\ \citep{kingma2018glow}}& \makecell[c]{SC-FEGAN \\ \citep{jo2019sc}}& \makecell[c]{SAN\\ \citep{dai2019second}} \tabularnewline
    \hline 
    \end{tabular}
\end{adjustbox}
\end{table*}

\begin{table}[!htbp]
\scriptsize
\centering
\setlength\tabcolsep{5pt} 
\caption{Information of real datasets.}
\vspace{-5pt}
\label{tab:real_database}
\begin{adjustbox}{width=\linewidth,center}
    \begin{tabular}{|c|c|c|c|c|}
    \hline 
    Database & Subjects & Images & \makecell[c]{Img per\\Subject} \tabularnewline
    \hline 
    \hline 
    CASIA-WebFace \citep{yi2014learning} & 10,575 & 494,414 & 46.75 \tabularnewline
    \hline 
    CelebA \citep{liu2015deep} & 10,000 & 200,000 & 20 \tabularnewline
    \hline 
    VGGFace2 \citep{cao2018vggface2} & 9,131 & 3,310,000 & 362.50 \tabularnewline
    \hline 
    FFHQ \citep{karras2019style} & unknown & 70,000 & unknown \tabularnewline
    \hline 
    VGGFace \citep{parkhi2015deep} & 2,622 & 2,600,000 & 991.61 \tabularnewline
    \hline 
    Ms-Celeb-1M \citep{guo2016ms} & 100,000 & 10,000,000 & 100 \tabularnewline
    \hline 
    \makecell[c]{MegaFace\\ \citep{kemelmacher2016megaface}} & 690,572 & 4,700,000 & 6.81 \tabularnewline
    \hline 
    LSUN \citep{yu2015lsun} & 30 & 1000,000 & 33333.33 \tabularnewline
    \hline 
    \end{tabular}
\end{adjustbox}
\vspace{-10pt}
\end{table}
\begin{table}[!htbp]
\centering
\scriptsize
\setlength\tabcolsep{3pt} 
\caption{Information of fake datasets.}
\vspace{-5pt}
\label{tab:fake_database}
\begin{adjustbox}{width=\linewidth,center}

\begin{tabular}{|c|c|c|c|c|c|}
\hline 
Database & \makecell[c]{Total Images\\/ Videos} & \makecell[c]{Real\\ Images} & \makecell[c]{Fake\\ Images} & \makecell[c]{Real\\ Videos} & \makecell[c]{Fake\\ Videos}\tabularnewline
\hline 
\hline 
UADFV \citep{li2018ictu} & 98 & - & - & 49 & 49\tabularnewline
\hline 
\makecell[c]{DeepFake-TIMIT\\ \citep{korshunov2018deepfakes}} & 620 & - & - & - & 620\tabularnewline
\hline 
\makecell[c]{DFDC Preview\\ \citep{dolhansky2019deepfake}} & 5,214 & - & - & 1,131 & 4,113\tabularnewline
\hline 
\makecell[c]{FaceForensics++ (FF++)\\ \citep{rossler2019FaceForensics++}} & 5,000 & - & - & 1,000 & 4,000\tabularnewline
\hline 
Celeb-DF \citep{li2020celeb} & 6,229 & - & - & 590 & 5,639\tabularnewline
\hline 
\makecell[c]{FakeCatcher\\ \citep{ciftci2020fakecatcher}} & 142 & - & - & - & 142\tabularnewline
\hline 
DFFD \citep{dang2020detection} & 303,039 & 58,703 & 240,336 & 1,000 & 3,000\tabularnewline
\hline 
\makecell[c]{Google DFD\\ \citep{dufour2019contributing}} & 3,431 & - & - & 363 & 3,068\tabularnewline
\hline 
\makecell[c]{DFDC \citep{dolhansky2020deepfake}} & 128,154 & - & - & 23,954 & 104,500\tabularnewline
\hline 
\makecell[c]{DeeperForensics\\ \citep{jiang2020deeperforensics}} & 60,000 & - & - & 50,000 & 10,000\tabularnewline
\hline 
\makecell[c]{Vox-DeepFake\\ \citep{dong2020identity}} & 2,171,215 & - & - & 1,125,429 & 1,045,786\tabularnewline
\hline 
\makecell[c]{WildDeepFake \citep{zi2020wilddeepfake}} & 7,314 & - & - & 3,805 & 3,509\tabularnewline
\hline 
\makecell[c]{ForgeryNet\\ \citep{he2021forgerynet}} & 3,117,309 & 1,438,201 & 1,457,861 & 99,630 & 121,617\tabularnewline
\hline 
\makecell[c]{DF-W\\ \citep{pu2021deepfake}} & 1,869  &-  & - & - & 1,869\tabularnewline
\hline 
\makecell[c]{FFIW$_{10K}$\\ \citep{zhou2021face}} & 20,000 & - & - & 10,000 & 10,000 \tabularnewline
\hline
\makecell[c]{OpenForensics\\ \citep{le2021openforensics}} & 115,325 & 45,473 & 70,325 & - & - \tabularnewline
\hline
\end{tabular}
\end{adjustbox}
\vspace{-10pt}
\end{table}

\begin{table*}[!htbp]
\scriptsize
\centering
\caption{Summary of the paper information of DeepFake generation methods and related datasets until August 1st, 2021. We mainly show the time, citation, days (from time of exposure), citation per day, method/dataset name, resolution, Elo rating/rank \citep{Elo} and project URL. From top to bottom, the table is comprised of three sections, and we show real datasets, fake datasets and generation methods in order. The selected time is when the camera-ready version of the paper is released, thus may be later than the time of the citation. For each of them, the papers are sorted by time. We also highlight the top-5 methods of citation, citation per day, and Elo rating. For resolution, we use ``aver'' to label the dataset which has images/videos of different sizes. For these datasets, we sample the dataset and show the average resolution. We also use ``align'' to label the datasets which usually be used after alignment and give the resolution of them. In total, 83 DeepFake generation methods are listed in the table.}
\vspace{-5pt}
\label{tab:paper_information}
\begin{adjustbox}{width=0.98\linewidth,center}
\setlength\tabcolsep{2.5pt} 

\begin{tabular}{|c|c|c|c|c|c|c|c|c|}
\hline 
\multirow{3}{*}{Time} & \multirow{3}{*}{Author} & \multirow{3}{*}{Citation} & \multirow{3}{*}{Days} & \multirow{3}{*}{\shortstack{Citation \\ Per Day}} & \multirow{3}{*}{Method/Dataset}  & \multirow{3}{*}{\shortstack{Resolution\\ of Images}} & \multirow{3}{*}{\shortstack{Elo Rating \\ / Rank}} & \multirow{3}{*}{\tabincell{c}{Project URL\\ ($\diamond$ 1st party project unavailable, 3rd party listed)}} \\
& & & & & & & & \tabularnewline
& & & & & & & & \tabularnewline
\hline 
\hline 
2014.11.28 & \cite{yi2014learning} & 1391 & 2438 & 0.57 & CASIA-WebFace & align(250,250,3) & N/A & \href{http://www.cbsr.ia.ac.cn/english/CASIA-WebFace-Database.html}{\tabincell{c}{cbsr.ia.ac.cn/english/CASIA-WebFace-Database}}\tabularnewline
\hline 
2015.09.07 & \cite{parkhi2015deep} & 4270 & 2155 & 1.98 & VGGFace & aver(569,395,3) & N/A & \href{https://www.robots.ox.ac.uk/\textasciitilde vgg/data/vgg\_face/}{www.robots.ox.ac.uk/\textasciitilde vgg/data/vgg\_face/}\tabularnewline
\hline 
2015.09.24 & \cite{liu2015deep} & 3956 & 2138 & 1.85 & CelebA & aver(613,507,3) & N/A & \href{http://mmlab.ie.cuhk.edu.hk/projects/CelebA.html}{mmlab.ie.cuhk.edu.hk/projects/CelebA}\tabularnewline
\hline 
2016.06.04 & \cite{yu2015lsun} & 772 & 1884 & 0.41 & LSUN & align(256,256,3) & N/A & \href{https://www.yf.io/p/lsun}{www.yf.io/p/lsun}\tabularnewline
\hline 
2016.06.27 & \cite{kemelmacher2016megaface} & 636 & 1861 & 0.34 & MegaFace & (112,112,3) & N/A & \href{http://megaface.cs.washington.edu/}{megaface.cs.washington.edu/}\tabularnewline
\hline 
2016.07.27 & \cite{guo2016ms} & 1044 & 1831 & 0.57 & Ms-Celeb-1M & align(112,112,3) & N/A & \href{https://www.msceleb.org/}{www.msceleb.org/}\tabularnewline
\hline 
2018.05.15 & \cite{cao2018vggface2} & 1209 & 1378 & 0.88 & VGGFace2 & aver(241,234,3) & N/A & \href{http://zeus.robots.ox.ac.uk/vgg\_face2/}{zeus.robots.ox.ac.uk/vgg\_face2/}\tabularnewline
\hline 
2019.03.29 & \cite{karras2019style} & 1994 & 856 & 2.33 & FFHQ & (1024,1024,3) & N/A & \href{https://github.com/NVlabs/ffhq-dataset}{github.com/NVlabs/ffhq-dataset}\tabularnewline
\hline 
\hline 
2018.03.24 & \cite{rossler2018FaceForensics} & 177 & 1226 & 0.14 & FaceForensics & aver(628,998,3) & 1319 / 78 & \href{https://niessnerlab.org/projects/roessler2019FaceForensicspp.html}{niessnerlab.org/projects/roessler2019FaceForensicspp}\tabularnewline
\hline 
2018.06.11 & \cite{li2018ictu} & 292 & 1147 & 0.25 & UADFV & aver(459,405,3) & 1311 / 80 & $\diamond$ \href{https://cutt.ly/Xh5dpZu}{cutt.ly/Xh5dpZu}\tabularnewline
\hline 
2018.12.20 & \cite{korshunov2018deepfakes} & 171 & 955 & 0.18 & DeepFake-TIMIT & (384,512,3) & 1386 / 66 & \href{https://www.idiap.ch/dataset/deepfaketimit}{www.idiap.ch/dataset/deepfaketimit}\tabularnewline
\hline 
2019.09.24 & \cite{dufour2019contributing} & 24 & 677 & 0.03 & Google DFD & (1080,1920,3) & 1342 / 75 & \href{https://ai.googleblog.com/2019/09/contributing-data-to-deepfake-detection.html}{\tabincell{c}{ai.googleblog.com/2019/09/contributing-data-to-deepfake-detection}}\tabularnewline
\hline 
2019.10.23 & \cite{dolhansky2019deepfake} & 123 & 648 & 0.19 & DFDC Preview & (1080,1920,3) & 1470 / 6 & \href{https://ai.facebook.com/datasets/dfdc/}{ai.facebook.com/datasets/dfdc/}\tabularnewline
\hline 
2019.10.29 & \cite{rossler2019FaceForensics++} & 430 & 642 & 0.67 & FaceForensics++ & aver(628,998,3) & 1170 / 83 & \href{https://github.com/ondyari/FaceForensics}{github.com/ondyari/FaceForensics}\tabularnewline
\hline 
2020.06.14 & \cite{li2020celeb} & 117 & 413 & 0.28 & Celeb-DF & aver(498,907,3) & 1463 / 7 & \href{http://www.cs.albany.edu/\textasciitilde lsw/celeb-deepfakeforensics.html}{www.cs.albany.edu/\textasciitilde lsw/celeb-deepfakeforensics}\tabularnewline
\hline 
2020.06.14 & \cite{dang2020detection} & 65 & 413 & 0.16 & DFFD & (1024,1024,3) & 1418 / 16 & \href{http://cvlab.cse.msu.edu/dffd-dataset.html}{cvlab.cse.msu.edu/dffd-dataset}\tabularnewline
\hline 
2020.07.19 & \cite{ciftci2020fakecatcher} & 73 & 378 & 0.19 & FakeCatcher & N/A & 1428 / 13 & \href{http://cs.binghamton.edu/\textasciitilde ncilsal2/DeepFakesDataset/}{cs.binghamton.edu/\textasciitilde ncilsal2/DeepFakesDataset/}\tabularnewline
\hline 
2020.10.12 & \cite{zi2020wilddeepfake} & 9 & 293 & 0.03 & WildDeepFake & N/A & 1400 / 24 & \href{https://github.com/deepfakeinthewild/deepfake-in-the-wild}{github.com/deepfakeinthewild/deepfake-in-the-wild}\tabularnewline
\hline
2020.10.28 & \cite{dolhansky2020deepfake} & 48 & 277 & 0.17 & DFDC & (1080,1920,3) & \cellcolor{royalblue!70}1621 / 1 & \href{https://ai.facebook.com/datasets/dfdc}{ai.facebook.com/datasets/dfdc}\tabularnewline
\hline 
2020.12.07 & \cite{dong2020identity} & 0 & 237 & 0 & Vox-DeepFake & (224,224,3) & 1443 / 10 & N/A\tabularnewline
\hline 
2020.12.11 & \cite{jiang2020deeperforensics} & 47 & 233 & 0.2 & DeeperForensics & aver(560,856,3) & 1430 / 11 & \href{https://liming-jiang.com/projects/DrF1/DrF1.html}{liming-jiang.com/projects/DrF1/DrF1}\tabularnewline
\hline 

2021.03.11 & \cite{he2021forgerynet} & 0 & 143 & 0 & ForgeryNet & aver(670,1145,3) & 1400 / 24 &
\href{https://yinanhe.github.io/projects/forgerynet}{yinanhe.github.io/projects/forgerynet}\tabularnewline
\hline 
2021.03.18 & \cite{zhou2021face} & 0 & 136 & 0 & FFIW$_{10K}$ & N/A & 1400 / 24 &
\href{https://github.com/tfzhou/FFIW}{github.com/tfzhou/FFIW}\tabularnewline
\hline 
2021.06.18 & \cite{pu2021deepfake} & 0 & 44 & 0 & DF-W & aver(751,1181,3) & 1400 / 24 &
\href{https://github.com/jmpu/webconf21-deepfakes-in-the-wild}{github.com/jmpu/webconf21-deepfakes-in-the-wild}\tabularnewline
\hline
2021.07.30 & \cite{le2021openforensics} & 2 & 2 & 1 & OpenForensics & (512,512,3) & 1400 / 24 &
\href{https://sites.google.com/view/ltnghia/research/openforensics}{sites.google.com/view/ltnghia/research/openforensics}\tabularnewline
\hline
\hline 
2016.01.07 & \cite{radford2015unsupervised} & \cellcolor{royalblue!70}9725 & 2033 & \cellcolor{royalblue!45}4.78 & DCGAN & (64,64,3) & 1457 / 8 & \href{https://github.com/Newmu/dcgan\_code}{github.com/Newmu/dcgan\_code}\tabularnewline
\hline 
2016.06.19 & \cite{Faceswap} & 3 & 1869 & 0 & FaceSwap & (720,1280) & 1400 / 24 & \href{https://github.com/MarekKowalski/FaceSwap/}{github.com/MarekKowalski/FaceSwap/}\tabularnewline
\hline 
2016.06.27 & \cite{thies2016face2face} & 868 & 1861 & 0.47 & Face2Face & (720,1280,3) & 1413 / 20 & \href{http://gvv.mpi-inf.mpg.de/projects/MZ/Papers/DemoF2F/page.html}{gvv.mpi-inf.mpg.de/projects/MZ/Papers/DemoF2F/page}\tabularnewline
\hline 
2016.11.13 & \cite{mao2016multi} & 173 & 1722 & 0.10 & LSGAN & (112,112,3) & 1345 / 74 & \href{https://github.com/xudonmao/LSGAN}{github.com/xudonmao/LSGAN}\tabularnewline
\hline 
2016.11.19 & \cite{perarnau2016invertible} & 432 & 1716 & 0.25 & IcGAN & (64,64,3) & 1306 / 81 & \href{https://github.com/Guim3/IcGAN}{github.com/Guim3/IcGAN}\tabularnewline
\hline 
2017.05.30 & \cite{bellemare2017cramer} & 195 & 1524 & 0.13 & CramerGAN & (160,160,3) & 1400 / 24 & \href{https://github.com/mbinkowski/MMD-GAN}{github.com/mbinkowski/MMD-GAN}\tabularnewline
\hline 
2017.05.31 & \cite{berthelot2017began} & 989 & 1523 & 0.65 & BEGAN & (128,128,3) & 1291 / 79 & $\diamond$ \href{https://github.com/carpedm20/BEGAN-tensorflow}{github.com/carpedm20/BEGAN-tensorflow}\tabularnewline
\hline 
2017.07.10 & \cite{iizuka2017globally} & 1089 & 1483 & 0.73 & G\&L & (256,256,3) & 1400 / 24 & \href{https://github.com/satoshiiizuka/siggraph2017\_inpainting}{github.com/satoshiiizuka/siggraph2017\_inpainting}\tabularnewline
\hline 
2017.07.10 & \cite{suwajanakorn2017synthesizing} & 533 & 1483 & 0.36 & A2V & (1080,1920,3) & 1400 / 24 & \href{https://github.com/supasorn/synthesizing\_obama\_network\_training}{\tabincell{c}{github.com/supasorn/synthesizing\_obama\_network\_training}}\tabularnewline
\hline
2017.07.22 & \cite{shu2017neural} & 203 & 1471 & 0.14 & NFE & (64,64,3) & 1400 / 24 & \href{https://github.com/zhixinshu/NeuralFaceEditing}{github.com/zhixinshu/NeuralFaceEditing}\tabularnewline
\hline 
2017.07.28 & \cite{chen2017photographic} & 676 & 1465 & 0.46 & CRN & (512,1024,3) & 1406 / 20 & \href{https://github.com/CQFIO/PhotographicImageSynthesis}{github.com/CQFIO/PhotographicImageSynthesis}\tabularnewline
\hline 
2017.09.13 & \cite{ding2018exprgan} & 106 & 1418 & 0.07 & ExprGAN & (128,128,3) & 1417 / 17 & \href{https://github.com/HuiDingUMD/ExprGAN}{github.com/HuiDingUMD/ExprGAN}\tabularnewline
\hline 
2017.10.22 & \cite{zhu2017unpaired} & \cellcolor{royalblue!57.5}9156 & 1379 & \cellcolor{royalblue!70}6.64 & CycleGAN & (256,256,3) & 1400 / 24 & \href{https://junyanz.github.io/CycleGAN/}{junyanz.github.io/CycleGAN/}\tabularnewline
\hline 
2017.12.06 & \cite{arjovsky2017wasserstein} & \cellcolor{royalblue!45}7289 & 1334 & \cellcolor{royalblue!57.5}5.46 & WGAN & (64,64,3) & 1368 / 72 & \href{https://github.com/martinarjovsky/WassersteinGAN}{github.com/martinarjovsky/WassersteinGAN}\tabularnewline
\hline 
2017.12.25 & \cite{gulrajani2017improved} & \cellcolor{royalblue!32.5}5147 & 1315 & \cellcolor{royalblue!32.5}3.91 & WGAN-GP & (128,128,3) & 1399 / 62 & \href{https://github.com/igul222/improved\_wgan\_training}{github.com/igul222/improved\_wgan\_training}\tabularnewline
\hline 
2018.01.28 & \cite{Faceswap-GAN} & N/A & 1281 & N/A & Faceswap-GAN & (256,256,3) & 1400 / 24 & \href{https://github.com/shaoanlu/faceswap-GAN}{github.com/shaoanlu/faceswap-GAN}\tabularnewline
\hline 
2018.02.16 & \cite{miyato2018spectral} & 2246 & 1262 & 1.78 & SNGAN & (128,128,3) & 1400 / 24 & \href{https://github.com/niffler92/SNGAN}{github.com/niffler92/SNGAN}\tabularnewline
\hline 
2018.02.26 & \cite{karras2017progressive} & \cellcolor{royalblue!20}3208 & 1252 & 2.56 & PGGAN & (1024,1024,3) & 1335 / 76 & \href{https://github.com/tkarras/progressive\_growing\_of\_gans}{github.com/tkarras/progressive\_growing\_of\_gans}\tabularnewline
\hline 
2018.03.21 & \cite{binkowski2018demystifying} & 340 & 1229 & 0.28 & MMDGAN & (160,160,3) & 1400 / 24 & \href{https://github.com/mbinkowski/MMD-GAN}{github.com/mbinkowski/MMD-GAN}\tabularnewline
\hline 
2018.03.21 & \cite{yu2018generative} & 1029 & 1229 & 0.84 & ContextAtten & (512,680,3) & 1430 / 12 & \href{https://github.com/JiahuiYu/generative\_inpainting}{github.com/JiahuiYu/generative\_inpainting}\tabularnewline
\hline 
2018.04.18 & \cite{natsume2018rsgan} & 52 & 1201 & 0.04 & RSGAN & (128,128,3) & 1400 / 24 & N/A \tabularnewline
\hline
2018.06.19 & \cite{chen2018learning} & 425 & 1139 & 0.37 & SITD & (4000,6000,3) & 1352 / 73 & \href{https://github.com/cchen156/Learning-to-See-in-the-Dark}{github.com/cchen156/Learning-to-See-in-the-Dark}\tabularnewline
\hline 
2018.07.10 & \cite{kingma2018glow} & 1110 & 1118 & 0.99 & Glow & (256,256,3) & \cellcolor{royalblue!45}1511 / 3 & \href{https://github.com/openai/glow}{github.com/openai/glow}\tabularnewline
\hline 
2018.09.08 & \cite{pumarola2018ganimation} & 346 & 1058 & 0.33 & GANimation & (128,128,3) & 1390 / 64 & \href{https://github.com/albertpumarola/GANimation}{github.com/albertpumarola/GANimation}\tabularnewline
\hline
2018.09.08 & \cite{zhang2018generative} & 92 & 1058 & 0.09 & SaGAN & (128,128,3) & 1400 / 24 & \href{https://github.com/elvisyjlin/SpatialAttentionGAN}{github.com/elvisyjlin/SpatialAttentionGAN}\tabularnewline
\hline 
2018.09.21 & \cite{choi2018stargan} & 1796 & 1045 & 1.72 & StarGAN & (256,256,3) & 1386 / 65 & \href{https://github.com/yunjey/stargan}{github.com/yunjey/stargan}\tabularnewline
\hline 
2019.02.25 & \cite{brock2018large} & 1941 & 888 & 2.19 & BigGAN & (256,256,3) & 1446 / 9 & \href{https://github.com/ajbrock/BigGAN-PyTorch}{github.com/ajbrock/BigGAN-PyTorch}\tabularnewline
\hline 
2019.04.04 & \cite{chen2018gated} & 59 & 850 & 0.07 & GatedGAN & (128,128,3) & 1368 / 71 & \href{https://github.com/xinyuanc91/Gated-GAN}{github.com/xinyuanc91/Gated-GAN}\tabularnewline
\hline 
2019.04.28 & \cite{thies2019deferred} & 300 & 826 & 0.36 & Neural-Texture & (512,512,3) & 1400 / 24 & $\diamond$ \href{https://github.com/SSRSGJYD/NeuralTexture}{github.com/SSRSGJYD/NeuralTexture}\tabularnewline
\hline 
2019.05.20 & \cite{he2019attgan} & 259 & 804 & 0.32 & AttGAN & (384,384,3) & 1426 / 15 & \href{https://github.com/LynnHo/AttGAN-Tensorflow}{github.com/LynnHo/AttGAN-Tensorflow}\tabularnewline
\hline 
2019.06.09 & \cite{cho2019image} & 45 & 784 & 0.06 & GDWCT & (256,256,3) & \cellcolor{royalblue!57.5}1532 / 2 & \href{https://github.com/WonwoongCho/GDWCT}{github.com/WonwoongCho/GDWCT}\tabularnewline
\hline 
2019.06.16 & \cite{karras2019style} & 1994 & 777 & \cellcolor{royalblue!20}2.57 & StyleGAN & (1024,1024,3) & 1384 / 68 & \href{https://github.com/NVlabs/stylegan}{github.com/NVlabs/stylegan}\tabularnewline
\hline 
2019.06.16 & \cite{liu2019stgan} & 116 & 777 & 0.15 & STGAN & (384,384,3) & 1390 / 63 & \href{https://github.com/csmliu/STGAN}{github.com/csmliu/STGAN}\tabularnewline
\hline 
2019.06.16 & \cite{dai2019second} & 383 & 777 & 0.49 & SAN & (392,392,3) & \cellcolor{royalblue!20}1486 / 5 & \href{https://github.com/daitao/SAN}{github.com/daitao/SAN}\tabularnewline
\hline 
2019.06.16 & \cite{yao2019attention} & 48 & 777 & 0.06 & AAMS & (512,512,3) & 1400 / 24 & \href{https://github.com/JianqiangRen/AAMS}{github.com/JianqiangRen/AAMS}\tabularnewline
\hline 
2019.06.16 & \cite{chen2019homomorphic} & 30 & 777 & 0.04 & HomointerpGAN & (128,128,3) & 1335 / 77 & \href{https://github.com/yingcong/HomoInterpGAN}{github.com/yingcong/HomoInterpGAN}\tabularnewline
\hline 
2019.06.16 & \cite{gu2019mask} & 38 & 777 & 0.05 & MaskPE & (256,256,3) & 1400 / 24 & \href{https://github.com/cientgu/Mask_Guided_Portrait_Editing}{github.com/cientgu/Mask\_Guided\_Portrait\_Editing}\tabularnewline
\hline 
2019.10.29 & \cite{li2019diverse} & 27 & 642 & 0.04 & IMLE & (256,512,3) & 1319 / 79 & \href{https://github.com/zth667/Diverse-Image-Synthesis-from-Semantic-Layout}{\tabincell{c}{github.com/zth667/Diverse-Image-Synthesis-from-Semantic-Layout}}\tabularnewline
\hline 
2019.10.29 & \cite{jo2019sc} & 102 & 642 & 0.16 & SC-FEGAN & (512,512,3) & \cellcolor{royalblue!32.5}1489 / 4 & \href{https://github.com/run-youngjoo/SC-FEGAN}{github.com/run-youngjoo/SC-FEGAN}\tabularnewline
\hline 
2019.10.29 & \cite{lin2019coco} & 53 & 642 & 0.08 & CocoGAN & (384,384,3) & 1400 / 24 & \href{https://github.com/hubert0527/COCO-GAN}{github.com/hubert0527/COCO-GAN}\tabularnewline
\hline 
2019.11.05 & \cite{park2019semantic} & 821 & 635 & 1.29 & GauGAN & (256,512,3) & 1383 / 69 & \href{https://github.com/NVlabs/SPADE}{github.com/NVlabs/SPADE}\tabularnewline
\hline 
2020.03.03 & \cite{durall2020watch} & 46 & 516 & 0.09 & WUCGAN & (1024,1024,3) & 1400 / 24 & \href{https://github.com/cc-hpc-itwm/UpConv}{github.com/cc-hpc-itwm/UpConv}\tabularnewline
\hline 
2020.04.26 & \cite{choi2020stargan} & 220 & 462 & 0.48 & StarGANv2 & (512,512,3) & 1400 / 24 & \href{https://github.com/clovaai/stargan-v2}{github.com/clovaai/stargan-v2}\tabularnewline
\hline 
2020.05.20 & \cite{petrov2020deepfacelab} & 36 & 438 & 0.08 & DeepFaceLab & (448,448,3) & 1427 / 14 & \href{https://github.com/iperov/DeepFaceLab}{github.com/iperov/DeepFaceLab}\tabularnewline
\hline 
2020.06.16 & \cite{karnewar2020msg} & 48 & 411 & 0.12 & MSGGAN & (1024,1024,3) & 1385 / 67 & \href{https://github.com/akanimax/msg-stylegan-tf}{github.com/akanimax/msg-stylegan-tf}\tabularnewline
\hline 
2020.06.16 & \cite{li2020advancing} & 14 & 411 & 0.03 & FaceShifter & (256,256,3) & 1402 / 21 & \href{https://lingzhili.com/FaceShifterPage/}{lingzhili.com/FaceShifterPage/}\tabularnewline
\hline 
2020.06.16 & \cite{li2020advancing} & 83 & 411 & 0.20 & Image2StyleGAN++
 & (1024,1024,3) & 1400 / 24 & N/A \tabularnewline
\hline 
2020.06.16 & \cite{li2020advancing} & 218 & 411 & 0.53 & InterFaceGAN
 & (1024,1024,3) & 1400 / 24 & \href{https://github.com/genforce/interfacegan}{github.com/genforce/interfacegan}\tabularnewline
\hline 
2020.06.16 & \cite{li2020advancing} & 35 & 411 & 0.08 & GANLocalEditing
 & (1024,1024,3) & 1400 / 24 & \href{https://github.com/cyrilzakka/GANLocalEditing}{github.com/cyrilzakka/GANLocalEditing}\tabularnewline
\hline 
2020.10.22 & \cite{viazovetskyi2020stylegan2} & 21 & 283 & 0.07 & StyleGAN 2 & (1024,1024,3) & 1415 / 18 & \href{https://github.com/NVlabs/stylegan2}{github.com/NVlabs/stylegan2}\tabularnewline
\hline 
2020.11.05 & \cite{zhu2020aot} & 6 & 269 & 0.02 & AOT & aver(628,998,3) & 1400 / 24 & \href{https://github.com/zhuhaozh/AOT}{github.com/zhuhaozh/AOT}\tabularnewline
\hline 
2020.12.03 & \cite{noroozi2020self} & 1 & 241 & 0 & slcGAN & (256,256,3) & 1400 / 24 & N/A\tabularnewline
\hline 
2020.12.05 & \cite{jung2020spectral} & 4 & 239 & 0.02 & SpectralGAN & (256,256,3) & 1400 / 24 & \href{https://github.com/steffen-jung/SpectralGAN}{\tabincell{c}{github.com/steffen-jung/SpectralGAN}}\tabularnewline
\hline 
2020.12.16 & \cite{liu2021self} & 0 & 228 & 0 & STIGAN & (1024,1024,3) & 1400 / 24 & \href{https://github.com/odegeasslbc/Self-Supervised-Sketch-to-Image-Synthesis-PyTorch}{\tabincell{c}{github.com/odegeasslbc/Self-Supervised-Sketch-to-Image-Synthesis-PyTorch}}\tabularnewline
\hline 
2020.12.17 & \cite{esser2021taming} & 30 & 227 & 0.13 & TTGAN & (1024,1840,3) & 1400 / 24 & \href{https://compvis.github.io/taming-transformers/}{compvis.github.io/taming-transformers/}\tabularnewline
\hline 
2020.12.23 & \cite{jiang2020focal} & 4 & 221 & 0.02 & FFL & (256,256,3) & 1400 / 24 & \href{https://github.com/EndlessSora/focal-frequency-loss}{github.com/EndlessSora/focal-frequency-loss}\tabularnewline
\hline
2021.06.18 & \cite{afifi2021histogan} & 1 & 44 & 0.02 & HistoGAN & (1024,1024,3) & 1400 / 24 & \href{https://github.com/mahmoudnafifi/HistoGAN}{github.com/mahmoudnafifi/HistoGAN}\tabularnewline
\hline 
2021.06.18 & \cite{wang2021one} & 13 & 44 & 0.30 & vid2vid & (1024,1024,3) & 1400 / 24 & \href{https://nvlabs.github.io/face-vid2vid/}{nvlabs.github.io/face-vid2vid}\tabularnewline
\hline 
2021.06.18 & \cite{tripathy2021facegan} & 1 & 44 & 0.02 & FACEGAN & (256,256,3) & 1400 / 24 & \href{https://tutvision.github.io/FACEGAN/}{tutvision.github.io/FACEGAN}\tabularnewline
\hline 
2021.06.18 & \cite{zhu2021one} & 0 & 44 & 0 & MegaFS & (1024,1024,3) & 1400 / 24 & \href{https://github.com/zyainfal/One-Shot-Face-Swapping-on-Megapixels}{github.com/zyainfal/One-Shot-Face-Swapping-on-Megapixels}\tabularnewline
\hline 
2021.06.18 & \cite{gao2021information} & 0 & 44 & 0 & InfoSwap & (512,512,3) & 1400 / 24 & \href{https://github.com/GGGHSL/InfoSwap-master}{github.com/GGGHSL/InfoSwap-master} \tabularnewline
\hline 
2021.06.18 & \cite{hyun2021self} & 0 & 44 & 0 & SVGAN & (64,64,3) & 1400 / 24 & N/A\tabularnewline
\hline 
2021.06.18 & \cite{xia2021tedigan} & 4 & 44 & 0.09 & TediGAN & (1024,1024,3) & 1400 / 24 & \href{https://github.com/IIGROUP/TediGAN}{github.com/IIGROUP/TediGAN}\tabularnewline
\hline 
2021.06.18 & \cite{li2021faceinpainter} & 0 & 44 & 0 & FaceInpainter & (512,512,3) & 1400 / 24 & N/A \tabularnewline
\hline 
2021.06.18 & \cite{zhang2021learning} & 1 & 44 & 0.02 & DLN & (224,224,3) & 1400 / 24 & N/A \tabularnewline
\hline
2021.06.18 & \cite{gao2021high} & 0 & 44 & 0 & HifaFace & (256,256,3) & 1400 / 24 & N/A \tabularnewline
\hline 
\end{tabular}
\end{adjustbox}
\end{table*}

\subsection{DeepFake Challenges}\label{DeepFake Challenges}
In recent years, there have been two famous DeepFake challenge: DeepFake Detection Challenge (DFDC) \citep{dolhansky2020deepfake} and DeeperForensics Challenge 2020 \citep{jiang2021deeperforensics}. 
The dataset DeeperForensics-1.0 \citep{jiang2020deeperforensics} and DFDC \citep{dolhansky2020deepfake} used by these two challenges are significantly larger than the previous datasets. They have 100,000 videos and 100,000,000 numbers of frames. 

The DeepFake Detection Challenge was hosted on the Kaggle platform\footnote{\scriptsize{\url{https://www.kaggle.com/c/deepfake-detection-challenge}}} by Facebook. During the course of the challenge, 2,114 teams participated. All final evaluations were tested on a private dataset, using a single V100 GPU. Submissions had to run over 10,000 videos within 90 hours. Of all of the scores on the private test set, 60\% of submissions had a log loss lower than or equal to 0.69, which is similar to the score if one were to predict a probability of 0.5 for every video. In contrast, the best models achieved very good detection performance on DFDC videos. In the top-5 winning solutions of 
DFDC, all of them were image-based detection methods. Three of the five methods used EfficientNet \citep{tan2019efficientnet} as the backbone model.

The DeeperForensics Challenge 2020 is hosted on the CodaLab platform\footnote{\scriptsize{\url{https://competitions.codalab.org/competitions/25228}}} in conjunction with ECCV 2020. During the course of the challenge, a total of 115 participants registered for the competition, and 25 teams made valid submissions. Similar to DFDC, the DeeperForensics Challenge uses binary cross-entropy loss (BCELoss) to evaluate the performance of detection models. The evaluation is conducted on a private test set, containing 3,000 videos. A total of two online evaluations (each with 7.5 hours of runtime limit) are allowed. Top-3 winning solutions achieved promising performance. Two of them used EfficientNet as the backbone model and all of them used augmentation in model training.

From the challenges, we can find two key points for building a powerful model. First, the backbone selection of the forgery detection models is important. The high-performance winning solutions are based on the state-of-the-art (SOTA) EfficientNet. Second, applying appropriate data augmentations may better simulate real-world scenarios and boost the model performance.

\subsection{Summary of DeepFake Generation Methods}\label{sec:sum_gen}
DeepFake generation methods have developed rapidly in recent years. Across the four main categories (entire face synthesis, attribute manipulation, identity swap, expression swap) and ``other'' generation methods, the high quality of the generated images has made it extremely hard for human eyes to distinguish between real and fake. Meanwhile, more and more real image datasets and fake image datasets also promote the development of generation and detection of the DeepFake research field.

However, we still think there's a large space to improve for the DeepFake generation methods. For example, the resolution of the generated images, manipulable face properties, the continuity of the video, \etc, could be further improved, which is introduced in detail in Section \ref{sec:horizon}.

To demonstrate the DeepFake generation methods, real datasets, and fake datasets in detail, we build four tables.

Table~\ref{tab:visualize} shows the images of top-5 generation methods based on battleground (Figure~\ref{fig:battle}) and Elo rating. In the first row, we show the DeepFake generation methods which are attempted the most by DeepFake detection methods according to Figure \ref{fig:battle}. In the second row, we show the DeepFake generation methods which have the highest Elo rating according to Table~\ref{tab:paper_information}.
Table~\ref{tab:real_database} and \ref{tab:fake_database} mainly introduce the information about real datasets and fake datasets. For real datasets, we have collected the number of images they contain and the diversity of the subjects. For the fake dataset, we have collected the number of images/videos they contain and show the number of real/fake ones clearly.

Table~\ref{tab:paper_information} is information statistics of the DeepFake generation methods, real datasets, and fake datasets. It contains the release time, the first author, the citations/days, the citation per day, the abbreviation of the method name, the resolution of images, the Elo rating, and the project URL. For each of the DeepFake generation methods, real datasets and fake datasets, we sort them in ascending chronological order. For the resolution of images in each real dataset, fake dataset, and generation method, we have collected the value from their paper. For those which have no exact value, we download the dataset and calculate the resolution of images/videos. For the value which has ``aver'' label, the resolution is calculated by taking the average resolution of dozens of images/videos. For the value which has ``align'' label, the resolution is recorded by the common resolution used in the DeepFake research field. For FakeCatcher, we could not find its resources and use ``N/A'' to label it. As shown in Table \ref{tab:paper_information}, according to citation, Elo rating and time, the top-3 models used for each type of DeepFake are DCGAN, GDWCT, StyleGAN2 (entire face synthesis), StarGAN, AttGAN, HifaFace (attribute manipulation), FaceForensics++, DFDC, OpenForensics (identity swap), Face2Face, SVGAN (expression swap).

Elo rating \citep{Elo} is widely used in chess and competitive games for calculating the players' ranking. For chess players, they may have different playing styles, which makes ranking difficult. Elo rating can give a relatively objective ranking according to the historical record. Similar to chess players, DeepFake generation methods also have different styles. Thus using the Elo rating to rank the detection difficulty of the generation method is also suitable. Although Elo rating can not give a very accurate ranking, it can give an efficient, intuitive, and objective ranking that is purely based on every single one-on-one battle between a detector and a generator. 

This paper mainly focuses on the battleground between DeepFake generation and DeepFake detection. For this purpose, we need to point out the DeepFake generation methods that are most difficult to detect by the existing DeepFake detection methods. Under the battleground, the desired metric should satisfy several requirements: (1) the metric can reflect the historical performance of a DeepFake generation method, (2) the metric should be flexible to reflect that a DeepFake generation method can be evaluated by more than one DeepFake detection methods, and across many research papers, (3) the metric should be as objective as possible to evaluate different DeepFake types (\ie, entire face synthesis, attribute manipulation, identity swap, expression swap). Existing metrics for evaluating deep generative approaches such as PSNR, SSIM are suitable for the tasks that ground truth image is available for the prediction image to compare with. The metrics such as FID, Inception score are suitable for comparing different image feature distributions. It is obvious that these metrics that evaluate the image quality or distributional closeness do not satisfy the three requirements mentioned above. To meet this challenge, we apply the Elo rating, a performance-related metric that is widely used in the ranking of chess or Go players. The Elo rating metric can simultaneously satisfy the three requirements.

For the calculating of Elo rating in DeepFake generation methods, we first set the Elo score of all the DeepFake generation methods to 1,400. Second, for each DeepFake detection method, we collected the generation method detected by them and sort the generation methods by the detection accuracy/AUC (\eg, if the detection accuracy on generation method A is lower than that on B, then we consider that the quality of A is higher than B, recorded as (A $>$ B)). Third, for the generation rank in each DeepFake detection method, we generate the strong and weak relationship between each pair of them (\eg, if A $>$ B $>$ C, then we generate three relationships A $>$ B, B $>$ C, A $>$ C). Fourth, we update the Elo score of each generation method by the strong and weak relationships. Consider the score of generation methods A and B are $\mathbf{score\_A}$ and $ \mathbf{score\_B}$, if A $>$ B, then the updated score $ \mathbf{score\_A'}$ and $ \mathbf{score\_B'}$ are calculated by below formula \eqref{score_A_adjust}-\eqref{score_B_update}. At last, we sort the generation methods by their Elo scores. Elo rating helps us to build strong and weak relationships with limited game information of different DeepFake generation methods. However, the Elo rating system has its inherent shortcoming. If a generation method only appears once but beats high-score competitors, its score will go up a lot. On the other hand, if a generation method has not been detected for a long time, its score will remain unchanged and cannot reflect its true difficulty.
\begin{align}
    & \mathbf{score\_A\_adjust} = \frac{1}{1+10^{\frac{\mathbf{score\_B}-\mathbf{score\_A}}{400}}} \label{score_A_adjust}\\
    & \mathbf{score\_B\_adjust} = \frac{1}{1+10^{\frac{\mathbf{score\_A}-\mathbf{score\_B}}{400}}} \\
    & \mathbf{score\_A'} = \mathbf{score\_A}+32*(1-\mathbf{score\_A\_adjust}) \\
    & \mathbf{score\_B'} = \mathbf{score\_B}-32*\mathbf{score\_B\_adjust}
    \label{score_B_update}
\end{align}

\section{Detection of DeepFakes}\label{sec:det}




In recent years, studies are continuously working on developing various techniques to identify whether a still image or video is synthesized with AI (especially manipulated with GANs and its variants) or produced naturally with a camera. 
Generalize to unseen synthesized techniques, robust against various attacks (\eg, adversarial attacks, image/video transformations), and providing explainable detection results are three critical factors for a detector practicality deployed in the wild. 
In this section, we mainly review recent studies on DeepFake detection based on their extracted features (\eg, spatial (Section~\ref{spatial_sec}), frequency (Section~\ref{frequency_sec}), and biological signals (Section~\ref{biological_sec})) and introduce their performance on the aforementioned three essential factors. In Section \ref{sec:DF_detection_other}, we detail the methods that can not be classified into the three typical DeepFake detection methods. To better present the DeepFake detection methods to readers, we use three tables (Table~\ref{tab:fake_detection_1}, \ref{tab:fake_detection_2}, and \ref{tab:fake_detection_3}) to summarize the existing DeepFake detection methods, a chord diagram (Figure \ref{fig:cd}) to show the performance among various DeepFake detectors in Section~\ref{sec:sum_det}, \revisedd{and a fishbone diagram (Figure~\ref{fig:fish_bone_detection}) to present the evolution of three typical detection techniques}. 




\begin{table*}[!htbp]
\scriptsize
\centering
\caption{Summary of existing DeepFake detection methods. We mainly show the time, author, the method type, employed classifier, the performance (\eg, worst and best performance), evaluation databases, evaluation object (\eg, image or video), compared baselines, and the detection capabilities in DeepFake detection. P, T, and B are employed for representing the baselines. P represents the method simply employs peer works for comparison, T indicates the method simply adopts some simple CNN models for comparison, B denotes the method leverages both peer works and conventional CNN models for comparison. G, R, and E are adopted for representing the capabilities of detection method. G represents whether the method has the generalization capabilities in tackling unseen DeepFakes, R indicates whether the method is robust against various attacks, especially image/ video transformations, E denotes whether the method provides explainable detection results. In total, there are $117$ methods listed in the table.}
\label{tab:fake_detection_1}
\begin{adjustbox}{width=0.97\linewidth,center}
\setlength\tabcolsep{3pt} 

\begin{tabular}{|c|c|c|c|c|c|c|c|c|c|c|c|c|c|}
\hline 
\multirow{2}*{Time} & \multirow{2}*{Author} & \multirow{2}*{Method} & \multirow{2}*{Classifier} & \multicolumn{2}{c|}{Performance} & \multirow{2}*{Databases} & \multicolumn{2}{c|}{Multimedia} & \multirow{2}*{B/L} & \multicolumn{3}{c|}{Capabilities}\tabularnewline   
 &   &  &   &   Worst & Best  &   & Img &  Vid  &  & G & R & E\tabularnewline
\hline 
\hline

2017.08.04 & \cite{zhang2017automated}  & Type \uppercase\expandafter{\romannumeral1}5 & SVM, RF, MLP & \tabincell{l}{ACC: 0.654} & \tabincell{l}{ACC: 0.9355} & 
Self-built & $\surd$ & $\times$ & N/A & $\times$ & $\times$ & $\times$\tabularnewline
\hline 

2018.03.29 & \cite{zhou2017two} & Type \uppercase\expandafter{\romannumeral1}2 & CNN & \tabincell{l}{N/A} & \tabincell{l}{AUC: 0.927} & 
Self-built & $\surd$ & $\times$ & B & $\times$ & $\times$ & $\times$\tabularnewline
\hline 

2018.04.10 & \cite{marra2018detection} & Type \uppercase\expandafter{\romannumeral1}2  & N/A & N/A & N/A & Self-built & $\surd$ & $\times$ & N/A & $\times$ & $\surd$ & $\times$\tabularnewline
\hline 

2018.06.11 & \cite{li2018ictu}  & Type \uppercase\expandafter{\romannumeral3}2 & CNN & \tabincell{l}{N/A} & \tabincell{l}{AUC: 0.99} & \tabincell{c}{UADFV}  & $\times$ & $\surd$ & N/A & $\times$ & $\times$ & $\surd$\tabularnewline
\hline

2018.06.20 & \cite{mo2018fake} & Type \uppercase\expandafter{\romannumeral1}2 & CNN & \tabincell{l}{N/A} & \tabincell{l}{ACC: 0.994 PGGAN} & Self-built & $\surd$ & $\times$ & N/A & $\times$ & $\times$ & $\times$\tabularnewline
\hline 

2018.08.29  & \cite{koopman2018detection} & Type \uppercase\expandafter{\romannumeral1}1 & N/A & \tabincell{l}{N/A} & \tabincell{l}{N/A} & Self-built & $\times$ & $\surd$ & N/A & $\times$ & $\times$ & $\times$\tabularnewline
\hline 

\multirow{2}*{2018.09.04} & \multirow{2}*{ \cite{afchar2018mesonet}}  & Type \uppercase\expandafter{\romannumeral1}2 & CNN & \tabincell{l}{ACC: 0.891} & \tabincell{l}{ACC: 0.984} & \tabincell{c}{FF++}  & $\times$ & $\surd$ & N/A & $\times$ & $\times$ & $\times$ \\ \cline{3-13}
& &\multicolumn{11}{c|}{ \href{https://github.com/DariusAf/MesoNet}{github.com/DariusAf/MesoNet}} \tabularnewline
\hline

2018.10.15 & \cite{tariq2018detecting} & Type \uppercase\expandafter{\romannumeral1}2 & CNN & \tabincell{l}{ACC: 0.9399} & \tabincell{l}{ACC: 0.9999} & Self-built & $\surd$ & $\times$ & T & $\times$ & $\times$ & $\times$\tabularnewline
\hline

2018.10.18 & \cite{hsu2018learning} & Type \uppercase\expandafter{\romannumeral1}2 & CNN & \tabincell{l}{ACC: 0.818 WGAN-GP} & \tabincell{l}{ACC: 0.947 LSGAN} & Self-built & $\surd$ & $\times$ & B & $\times$ & $\times$ & $\times$\tabularnewline
\hline

2018.11.12 & \cite{li2018can} & Type \uppercase\expandafter{\romannumeral4}1 & N/A & N/A & N/A & Self-built  & $\surd$ & $\times$ & N/A & $\surd$ & $\times$ & $\times$\tabularnewline
\hline

2018.11.27 & \cite{guera2018DeepFake} & Type \uppercase\expandafter{\romannumeral3}2 & RNN & \tabincell{l}{ACC: 0.967} & \tabincell{l}{ACC: 0.971} & \tabincell{c}{Self-built}  & $\times$ & $\surd$ & N/A & $\times$ & $\times$ & $\surd$\tabularnewline
\hline

2018.12.20  &  \cite{korshunov2018deepfakes} & Type \uppercase\expandafter{\romannumeral3}1 & CNN & \tabincell{l}{N/A} & \tabincell{l}{EER: 0.0333} & DeepFake-TIMIT & $\times$ & $\surd$ & B & $\times$ & $\times$ & $\surd$\tabularnewline
\hline

2019.01.07 & \cite{matern2019exploiting} & Type \uppercase\expandafter{\romannumeral3}2 & KNN, MLP, LR & \tabincell{l}{AUC: 0.843 Glow} & \tabincell{l}{AUC: 0.866} & FF & $\surd$ & $\times$ & N/A & $\times$ & $\times$ & $\surd$\tabularnewline
\hline

2019.02.26 & \cite{chen2019secure} & Type \uppercase\expandafter{\romannumeral4}1 & SVM & \tabincell{l}{ACC: 1.0} & \tabincell{l}{ACC: 1.0} & Self-built & $\surd$ & $\times$ & N/A & $\times$ & $\surd$ & $\times$\tabularnewline
\hline

2019.03.28 & \cite{marra2019gans} & Type \uppercase\expandafter{\romannumeral2}1 & N/A & \tabincell{l}{ACC: 0.99} & \tabincell{l}{ACC: 0.998} & Self-built & $\surd$ & $\times$ & N/A & $\surd$ & $\times$ & $\times$\tabularnewline
\hline

2019.03.30 & \cite{yang2019exposing_1} & Type \uppercase\expandafter{\romannumeral3}2 & SVM & \tabincell{l}{AUC: 0.83} & \tabincell{l}{AUC: 0.9413} & Self-built & $\surd$ & $\surd$ & T & $\times$ & $\times$ & $\surd$\tabularnewline
\hline

2019.05.12  & \cite{nguyen2019capsule} & Type \uppercase\expandafter{\romannumeral1}2 & CNN & \tabincell{l}{ACC: 0.8333} & \tabincell{l}{ACC: 0.9933} & FF & $\times$ & $\surd$ & T & $\times$ & $\surd$ & $\times$\tabularnewline
\hline

2019.05.12 & \cite{yang2019exposing} & Type \uppercase\expandafter{\romannumeral3}2 & SVM & \tabincell{l}{AUC: 0.843} & \tabincell{l}{AUC: 0.89} & \tabincell{c}{MFC, UADFV}  & $\times$ & $\surd$ & N/A & $\times$ & $\times$ & $\surd$\tabularnewline
\hline

2019.05.16 & \cite{sabir2019recurrent} & Type \uppercase\expandafter{\romannumeral1}2 & RNN & \tabincell{l}{ACC: 0.9843} & \tabincell{l}{ACC: 0.9959} & \tabincell{c}{FF++ }  & $\times$ & $\surd$ & T & $\times$ & $\times$ & $\times$\tabularnewline
\hline

2019.05.22  & \cite{li2018exposing} & Type \uppercase\expandafter{\romannumeral3}2 & CNN & \tabincell{l}{ACC: 0.932} & \tabincell{l}{ACC: 0.999} & \tabincell{c}{UADFV,\\ DeepFake-TIMIT} & $\times$ & $\surd$ & B & $\times$ & $\times$ & $\surd$\tabularnewline
\hline 

2019.06.17 & \cite{nguyen2019multi}  & Type \uppercase\expandafter{\romannumeral1}2 & CNN & \tabincell{l}{ACC: 0.5232} & \tabincell{l}{ACC: 0.9277} & FF++ & $\times$ & $\surd$ & P & $\surd$ & $\times$ & $\times$\tabularnewline
\hline

2019.06.18 & \cite{agarwal2019protecting}  & Type \uppercase\expandafter{\romannumeral1}3 & SVM & \tabincell{l}{AUC: 0.93} & \tabincell{l}{AUC: 1} & \tabincell{c}{Self-built}  & $\times$ & $\surd$ & N/A & $\times$ & $\times$ & $\surd$\tabularnewline
\hline

2019.06.18 & \cite{fernandes2019predicting}  & Type \uppercase\expandafter{\romannumeral3}3  & LSTM, VAE & \tabincell{l}{AUC: 0.93} & \tabincell{l}{AUC: 1} & \tabincell{c}{Self-built}  & $\surd$ & $\surd$ & N/A & $\times$ & $\times$ & $\surd$\tabularnewline
\hline

2019.06.18 & \cite{amerini2019DeepFake} & Type \uppercase\expandafter{\romannumeral1}2 & CNN & \tabincell{l}{ACC: 0.7546} & \tabincell{l}{ACC: 0.8161} & \tabincell{c}{FF++}  & $\times$ & $\surd$ & N/A & $\times$ & $\times$ & $\times$\tabularnewline
\hline

\multirow{2}*{2019.08.16}  & \multirow{2}*{ \cite{yu2019attributing}} & Type \uppercase\expandafter{\romannumeral2}1 & CNN & \tabincell{l}{ACC: 0.9766} & \tabincell{l}{ACC: 0.9950} & Self-built & $\surd$ & $\times$ & B & $\surd$ & $\times$ & $\surd$ \\ \cline{3-13}
&&\multicolumn{11}{c|}{ \href{https://github.com/ningyu1991/GANFingerprints}{github.com/ningyu1991/GANFingerprints}} \tabularnewline
\hline

2019.09.01 & \cite{dang2019face}  & Type \uppercase\expandafter{\romannumeral1}2 & CNN & \tabincell{l}{AUC: 0.90} & \tabincell{l}{AUC: 0.934} & Self-built & $\surd$ & $\times$ & B & $\times$ & $\times$ & $\times$\tabularnewline
\hline

2019.09.22 & \cite{he2019detection}  & Type \uppercase\expandafter{\romannumeral1}2 & CNN & \tabincell{l}{ACC: 0.9987} & \tabincell{l}{ACC: 1.0} & Self-built & $\surd$ & $\times$ & P & $\times$ & $\surd$ & $\times$\tabularnewline
\hline 

2019.09.22 & \cite{mccloskey2019detecting}  & Type \uppercase\expandafter{\romannumeral2}1 & SVM & \tabincell{l}{AUC: 0.61} & \tabincell{l}{AUC: 0.92} & Self-built & $\surd$ & $\times$ & N/A & $\times$ & $\times$ & $\times$\tabularnewline
\hline 

2019.10.03 & \cite{nataraj2019detecting}  & Type \uppercase\expandafter{\romannumeral1}1 & CNN & \tabincell{l}{ACC: 0.9937 StarGAN} & \tabincell{l}{ACC: 9971 cycleGAN} & Self-built & $\surd$ & $\times$ & B & $\surd$ & $\surd$ & $\times$\tabularnewline
\hline 

2019.10.06 & \cite{marra2019incremental} & Type \uppercase\expandafter{\romannumeral1}2 & \tabincell{c}{Multi-task Incremental\\ Learning} & \tabincell{l}{ACC: 0.6771} & \tabincell{l}{ACC: 0.9937} & \tabincell{c}{Self-built}  & $\surd$ & $\times$ & B & $\times$ & $\times$ & $\times$\tabularnewline
\hline

2019.10.12  & \cite{songsri2019complement} & Type \uppercase\expandafter{\romannumeral1}4 & CNN & \tabincell{l}{N/A} & \tabincell{l}{ACC: 0.9926} & \tabincell{c}{FF++}  & $\surd$ & $\surd$ & T & $\times$ & $\times$ & $\times$\tabularnewline
\hline

\multirow{2}*{2019.10.15} & \multirow{2}*{ \cite{zhang2019detecting}}  & Type \uppercase\expandafter{\romannumeral2}1 & CNN & \tabincell{l}{ACC: 0.786} & \tabincell{l}{ACC: 1} & 
Self-built & $\surd$ & $\times$ & B & $\surd$ & $\surd$ & $\surd$ \\ \cline{3-13}
&&\multicolumn{11}{c|}{ \href{https://github.com/ColumbiaDVMM/AutoGAN}{github.com/ColumbiaDVMM/AutoGAN}} \tabularnewline
\hline

2019.10.29  & \cite{nguyen2019use} & Type \uppercase\expandafter{\romannumeral1}2 & CNN & N/A & \tabincell{l}{ACC: 0.9311} & FF++ & $\times$ & $\surd$ & T & $\times$ & $\times$ & $\times$\tabularnewline
\hline

2019.11.11 &  \cite{sohrawardi2019poster}  & Type \uppercase\expandafter{\romannumeral1}3 & LSTM & \tabincell{l}{ACC: 0.86} & \tabincell{l}{ACC: 0.95} & FF++ & $\times$ & $\surd$ & P & $\times$ & $\surd$ & $\times$\tabularnewline
\hline 

2019.11.17 &  \cite{fernando2019exploiting}  & Type \uppercase\expandafter{\romannumeral1}2 & HMN & \tabincell{l}{ACC: 0.8412} & \tabincell{l}{ACC: 0.9997} & FF++ & $\surd$ & $\surd$ & B & $\surd$ & $\surd$ & $\times$\tabularnewline
\hline 

2019.11.27 & \cite{cozzolino2018forensictransfer}  & Type \uppercase\expandafter{\romannumeral1}2 & CNN & \tabincell{l}{ACC: 0.7062} & \tabincell{l}{ACC: 1.0} & \tabincell{c}{Self-built}  & $\surd$ & $\times$ & B & $\surd$ & $\times$ & $\times$\tabularnewline
\hline

2019.12.10 & \cite{xuan2019generalization}  & Type \uppercase\expandafter{\romannumeral1}2 & CNN & \tabincell{l}{ACC: 0.6055} & \tabincell{l}{ACC: 0.9545} & Self-built & $\surd$ & $\times$ & N/A & $\surd$ & $\times$ & $\times$\tabularnewline
\hline

2019.12.12 &  \cite{li2019zooming}  & Type \uppercase\expandafter{\romannumeral1}4 & CNN & \tabincell{l}{ACC: 0.9835} & \tabincell{l}{AUC: 0.9918} & FF++ & $\times$ & $\surd$ & T & $\times$ & $\times$ & $\times$\tabularnewline
\hline

2019.12.21 &  \cite{yu2019detecting}  & Type \uppercase\expandafter{\romannumeral1}2 & CNN & \tabincell{l}{ACC: 0.805} & \tabincell{l}{ACC: 0.981} & \tabincell{c}{FF++}  & $\surd$ & $\surd$ & T & $\times$ & $\times$ & $\times$\tabularnewline
\hline

2020.01.03 & \cite{hsu2020deep} & Type \uppercase\expandafter{\romannumeral1}2 & CNN & \tabincell{l}{Pre.: 0.929 DCGAN} & \tabincell{l}{Pre.: 0.988 WGAN} & Self-built & $\surd$ & $\times$ & P & $\times$ & $\times$ & $\times$\tabularnewline
\hline

2020.01.21 & \cite{kumar2020detecting} & Type \uppercase\expandafter{\romannumeral1}2 & CNN & \tabincell{l}{ACC.: 0.9120} & \tabincell{l}{ACC.: 0.9996} & FF & $\times$ & $\surd$ & B & $\times$ & $\surd$ & $\times$\tabularnewline
\hline

2020.02.11 & \cite{tarasiou2020extracting}  & Type \uppercase\expandafter{\romannumeral1}2 & CNN & \tabincell{l}{ACC: 0.8876} & \tabincell{l}{ACC: 0.9803} & \tabincell{c}{Google DFD,\\ FF++, Celeb-DF} & $\times$ & $\surd$ & N/A & $\times$ & $\times$ & $\times$\tabularnewline
\hline 

2020.03.03 & \cite{durall2020watch}  & Type \uppercase\expandafter{\romannumeral2}2 & CNN & \tabincell{l}{ACC: 0.85} & \tabincell{l}{ACC: 0.90} & \tabincell{c}{FF++} & $\times$ & $\surd$ & N/A & $\times$ & $\times$ & $\surd$\tabularnewline
\hline 

\multirow{2}*{2020.03.04} & \multirow{2}*{ \cite{durall2019unmasking}}  & Type \uppercase\expandafter{\romannumeral2}2 & \tabincell{c}{SVM, K-Means, LR} & \tabincell{l}{ACC: 0.9} & \tabincell{l}{ACC: 1} & \tabincell{c}{FF++} & $\surd$ & $\surd$ & N/A & $\times$ & $\times$ & $\times$ \\ \cline{3-13} 
&&\multicolumn{11}{c|}{ \href{https://github.com/cc-hpc-itwm/DeepFakeDetection}{github.com/cc-hpc-itwm/DeepFakeDetection}} \tabularnewline
\hline

2020.03.19  &  \cite{liu2020global} & Type \uppercase\expandafter{\romannumeral1}2 & CNN & \tabincell{l}{ACC: 0.9854} & \tabincell{l}{ACC: 0.991} & Self-built & $\surd$ & $\times$ & B & $\surd$ & $\surd$ & $\times$\tabularnewline
\hline 

2020.03.19 & \cite{kumar2020detecting} & Type \uppercase\expandafter{\romannumeral1}2 & CNN & \tabincell{l}{AUC: 0.8273} & \tabincell{l}{AUC: 0.997} & \tabincell{c}{FF++, Celeb-DF }  & $\times$ & $\surd$ & B & $\times$ & $\times$ & $\times$\tabularnewline
\hline 

2020.03.27  &  \cite{mansourifar2020one} & Type \uppercase\expandafter{\romannumeral1}2 & CNN & N/A & \tabincell{l}{ACC: 0.81} & Self-built  & $\surd$ & $\times$ & P & $\times$ & $\times$ & $\times$\tabularnewline
\hline 

\multirow{2}*{2020.04.06} & \multirow{2}*{ \cite{dogonadze2020deep}}  & Type \uppercase\expandafter{\romannumeral1}2 & CNN & \tabincell{l}{N/A} & \tabincell{l}{ACC: 0.748} & \tabincell{c}{FF++}  & $\surd$ & $\surd$ & B & $\times$ & $\times$ & $\times$ \\ \cline{3-13}
&&\multicolumn{11}{c|}{ \href{https://github.com/Megatvini/DeepFaceForgeryDetection}{github.com/Megatvini/DeepFaceForgeryDetection/}} \tabularnewline
\hline 

2020.04.16  &  \cite{bonettini2021use} & Type \uppercase\expandafter{\romannumeral2}2 & RF & \tabincell{l}{ACC: 0.9813} & \tabincell{l}{ACC: 1} & Self-built  & $\surd$ & $\times$ & B & $\times$ & $\surd$ & $\times$\tabularnewline
\hline 

\multirow{2}*{2020.04.16}  & \multirow{2}*{ \cite{bonettini2021video}} & Type \uppercase\expandafter{\romannumeral1}2 & CNN & \tabincell{l}{AUC: 0.8785} & \tabincell{l}{AUC: 0.9444} & \tabincell{c}{DFDC, FF++}  & $\times$ & $\surd$ & T & $\times$ & $\times$ & $\times$ \\ \cline{3-13}
&&\multicolumn{11}{c|}{ \href{https://github.com/polimi-ispl/icpr2020dfdc}{github.com/polimi-ispl/icpr2020dfdc}} \tabularnewline
\hline

2020.04.19 & \cite{li2020face}  & Type \uppercase\expandafter{\romannumeral1}4 & \tabincell{c}{HRNet\\ \citep{sun2019high}} & \tabincell{l}{AUC: 0.7115} & \tabincell{l}{AUC: 0.9540} & \tabincell{c}{Celeb-DF,\\ DFDC, FF++,\\ Google DFD}  & $\times$ & $\surd$ & B & $\surd$ & $\times$ & $\surd$\tabularnewline
\hline

2020.04.22 & \cite{guarnera2020DeepFake}  & Type \uppercase\expandafter{\romannumeral1}5 & KNN, SVM, LDA & \tabincell{l}{ACC: 0.8840 GDWCT} & \tabincell{l}{ACC: 0.9981 StyleGAN2} & Self-built  & $\surd$ & $\times$ & N/A & $\times$ & $\times$ & $\surd$\tabularnewline
\hline 

2020.04.29  & \cite{agarwal2020detecting} & Type \uppercase\expandafter{\romannumeral1}3 & CNN & \tabincell{l}{AUC: 0.824} & \tabincell{l}{AUC: 0.989} & \tabincell{c}{Celeb-DF,\\ DFDC-Preview}  & $\times$ & $\surd$ & B & $\surd$ & $\times$ & $\times$\tabularnewline
\hline

2020.05.04 & \cite{wu2020sstnet}  & Type \uppercase\expandafter{\romannumeral1}2 & CNN+RNN & \tabincell{l}{ACC: 0.9011} & \tabincell{l}{ACC: 0.9857} & \tabincell{c}{FF++}  & $\times$ & $\surd$ & B & $\times$ & $\times$ & $\times$\tabularnewline
\hline

2020.05.12 & \cite{hulzebosch2020detecting}  & Type \uppercase\expandafter{\romannumeral1}5 & CNN & \tabincell{l}{ACC: 0.983} & \tabincell{l}{ACC: 0.998} & Self-built  & $\surd$ & $\times$ & N/A & $\surd$ & $\surd$ & $\surd$\tabularnewline
\hline

2020.05.17  & \cite{sambhu2020detecting} & Type \uppercase\expandafter{\romannumeral1}2 & CNN & \tabincell{l}{ACC: 0.993} & \tabincell{l}{ACC: 0.996} & \tabincell{c}{FF++}  & $\surd$ & $\surd$ & T & $\times$ & $\times$ & $\times$\tabularnewline
\hline 

2020.05.19 & \cite{ding2020swapped}  & Type \uppercase\expandafter{\romannumeral1}2 & CNN & \tabincell{l}{ACC: 0.9197} & \tabincell{l}{ACC: 1} & Self-built & $\surd$ & $\times$ & N/A & $\times$ & $\times$ & $\times$\tabularnewline
\hline

2020.06.01 & \cite{chugh2020not}  & Type \uppercase\expandafter{\romannumeral3}1 & N/A & \tabincell{l}{ACC: 0.916} & \tabincell{l}{ACC: 0.983} & \tabincell{c}{DFDC,\\ DeepFake-TIMIT}  & $\times$ & $\surd$ & B & $\times$ & $\times$ & $\surd$\tabularnewline
\hline

2020.06.09 & \cite{huang2020fakelocator} & Type \uppercase\expandafter{\romannumeral1}4 & CNN & N/A & N/A & Self-built & $\surd$ & $\times$ & N/A & $\surd$ & $\surd$ & $\surd$\tabularnewline
\hline

2020.06.16 & \cite{mas2020DeepFakes} & Type \uppercase\expandafter{\romannumeral1}2 & RNN, CNN & \tabincell{l}{-} & \tabincell{l}{log-likelihood err: 0.321} & \tabincell{c}{DFDC}  & $\times$ & $\surd$ & B & $\surd$ & $\times$ & $\times$\tabularnewline
\hline

\end{tabular}
\end{adjustbox}
\end{table*}


\begin{table*}[!htbp]
\scriptsize
\centering
\caption{Continued Table~\ref{tab:fake_detection_1}.}\label{tab:fake_detection_2}
\begin{adjustbox}{width=0.97\linewidth,center}
\setlength\tabcolsep{3pt} 

\begin{tabular}{|c|c|c|c|c|c|c|c|c|c|c|c|c|c|}
\hline 
\multirow{2}*{Time} & \multirow{2}*{Author} & \multirow{2}*{Method} & \multirow{2}*{Classifier} & \multicolumn{2}{c|}{Performance} & \multirow{2}*{Databases} & \multicolumn{2}{c|}{Multimedia} & \multirow{2}*{B/L} & \multicolumn{3}{c|}{Capabilities}\tabularnewline   
 & &  & &    Worst & Best  &   & Img &  Vid  &  & G & R & E\tabularnewline
\hline 
\hline

2020.06.16 &  \cite{Agarwal_2020_CVPR_Workshops}  & Type \uppercase\expandafter{\romannumeral3}1 & CNN & \tabincell{l}{ACC: 0.928} & \tabincell{l}{ACC: 0.97} & \tabincell{c}{Self-built}  & $\times$ & $\surd$ & N/A & $\times$ & $\times$ & $\surd$\tabularnewline
\hline

2020.06.16 &  \cite{tursman2020towards}  & Type \uppercase\expandafter{\romannumeral3}2 & \tabincell{c}{Hierarchical Clustering} & \tabincell{l}{ACC: 0.60} & \tabincell{l}{ACC: 0.98} & \tabincell{c}{Self-built}  & $\times$ & $\surd$ & T & $\times$ & $\times$ & $\surd$\tabularnewline
\hline

\multirow{2}*{2020.06.16} & \multirow{2}*{ \cite{wang2020cnn}} & Type \uppercase\expandafter{\romannumeral1}2 & CNN & \tabincell{l}{Pre.: 0.529} & \tabincell{l}{Pre.: 1.0} & \tabincell{c}{FF++}  & $\surd$ & $\surd$ & P & $\surd$ & $\surd$ & $\times$ \\ \cline{3-13} 
&&\multicolumn{11}{c|}{ \href{https://github.com/peterwang512/CNNDetection}{github.com/peterwang512/CNNDetection}} \tabularnewline
\hline

2020.06.16 & \cite{khalid2020oc} & Type \uppercase\expandafter{\romannumeral1}2 & One-class VAE & \tabincell{l}{ACC: 0.712} & \tabincell{l}{ACC: 0.982} & \tabincell{c}{FF++}  & $\times$ & $\surd$ & B & $\times$ & $\times$ & $\times$\tabularnewline
\hline

2020.06.19 & \cite{bai2020fake} & Type \uppercase\expandafter{\romannumeral2}2 & CNN & \tabincell{l}{ACC: 0.9319} & \tabincell{l}{ACC: 0.9768} & Self-built  & $\surd$ & $\times$ & B & $\times$ & $\times$ & $\times$\tabularnewline
\hline 

\multirow{2}*{2020.06.26}  & \multirow{2}*{ \cite{frank2020leveraging}} & Type \uppercase\expandafter{\romannumeral2}2 & CNN & \tabincell{l}{ACC: 0.9780} & \tabincell{l}{ACC: 0.9991} & Self-built  & $\surd$ & $\times$ & T & $\times$ & $\surd$ & $\surd$ \\ \cline{3-13}
&&\multicolumn{11}{c|}{ \href{https://github.com/RUB-SysSec/GANDCTAnalysis}{github.com/RUB-SysSec/GANDCTAnalysis}} \tabularnewline
\hline 

\multirow{2}*{2020.06.26} & \multirow{2}*{ \cite{de2020deepfake}}  & Type \uppercase\expandafter{\romannumeral1}2 & CNN & \tabincell{l}{ACC: 0.7625} & \tabincell{l}{ACC: 0.9826} & \tabincell{c}{Celeb-DF}  & $\times$ & $\surd$ & B & $\times$ & $\times$ & $\times$ \\ \cline{3-13}
&&\multicolumn{11}{c|}{ \href{https://github.com/oidelima/Deepfake-Detection}{github.com/oidelima/Deepfake-Detection}} \tabularnewline
\hline

\multirow{2}*{2020.06.28}  & \multirow{2}*{ \cite{trinh2021interpretable}} & Type \uppercase\expandafter{\romannumeral1}2 & DNN & \tabincell{l}{ACC: 0.9037} & \tabincell{l}{ACC: 0.9625} & \tabincell{c}{FF++, Celeb-DF}  & $\times$ & $\surd$ & B & $\times$ & $\times$ & $\surd$ \\ \cline{3-13}
&&\multicolumn{11}{c|}{ \href{https://github.com/loc-trinh/DPNet}{github.com/loc-trinh/DPNet}} \tabularnewline
\hline

2020.07.02 &  \cite{tolosana2020deepfakes}  & Type \uppercase\expandafter{\romannumeral4}1 & N/A & N/A & N/A & \tabincell{c}{FF++, Celeb-DF,\\ DFDC, UADFV}   & $\times$ & $\surd$ & N/A & $\times$ & $\times$ & $\times$\tabularnewline
\hline 

2020.07.15 & \cite{pishori2020detecting}  & Type \uppercase\expandafter{\romannumeral4}1 & \tabincell{c}{LSTM, Eye Blink,\\ Grayscale Histograms} & \tabincell{l}{ACC: 0.8167} & \tabincell{l}{ACC: 0.8571} & \tabincell{c}{DFDC} & $\times$ & $\surd$ & B & $\times$ & $\times$ & $\surd$\tabularnewline
\hline

2020.07.16 & \cite{wang2020fakespotter}  & Type \uppercase\expandafter{\romannumeral1}5 & CNN & \tabincell{l}{ACC: 0.682 DFDC \\ ACC: 0.88 StarGAN} & \tabincell{l}{ACC: 0.985 FF++ \\ ACC: 0.986 PGGAN} & \tabincell{c}{Celeb-DF \\ FF++, DFDC} & $\surd$ & $\surd$ & B & $\surd$ & $\times$ & $\surd$\tabularnewline
\hline

2020.07.19 & \cite{ciftci2020fakecatcher}  & Type \uppercase\expandafter{\romannumeral3}3 & CNN & \tabincell{l}{ACC: 0.9107} & \tabincell{l}{ACC: 0.96} & \tabincell{c}{ FF++, Celeb-DF,\\ FF, FakeCatcher}  & $\times$ & $\surd$ & B & $\surd$ & $\times$ & $\surd$\tabularnewline
\hline

2020.07.20 &  \cite{goebel2020detection}  & Type \uppercase\expandafter{\romannumeral1}4 & CNN & \tabincell{l}{N/A} & \tabincell{l}{ACC: 0.9916} & \tabincell{c}{Self-built}  & $\surd$ & $\times$ & B & $\surd$ & $\surd$ & $\surd$\tabularnewline
\hline

2020.07.30 &  \cite{bonomi2020dynamic}  & Type \uppercase\expandafter{\romannumeral1}3 & SVM & \tabincell{l}{ACC: 0.8595} & \tabincell{l}{ACC: 0.9024} & \tabincell{c}{FF++}  & $\times$ & $\surd$ & B & $\surd$ & $\surd$ & $\surd$\tabularnewline
\hline

\multirow{2}*{2020.08.01} & \multirow{2}*{ \cite{mittal2020emotions}}  & Type \uppercase\expandafter{\romannumeral3}1 & N/A & \tabincell{l}{ACC: 0.844} & \tabincell{l}{ACC: 0.966} & \tabincell{c}{DFDC,\\ DeepFake-TIMIT}  & $\times$ & $\surd$ & B & $\surd$ & $\times$ & $\surd$ \\ \cline{3-13}
&&\multicolumn{11}{c|}{ \href{https://gamma.umd.edu/deepfakes/}{gamma.umd.edu/deepfakes/}} \tabularnewline
\hline

2020.08.07 &  \cite{guarnera2020fighting} & Type \uppercase\expandafter{\romannumeral1}3 & Random Forest & \tabincell{l}{ACC: 0.7219} & \tabincell{l}{ACC: 0.9964} & \tabincell{c}{Self-built}  & $\surd$ & $\surd$ & B & $\times$ & $\surd$ & $\surd$\tabularnewline
\hline

\multirow{2}*{2020.08.10} & \multirow{2}*{ \cite{jeon2020fdftnet}}  & Type \uppercase\expandafter{\romannumeral1}2 & SVM, RF, MLP & \tabincell{l}{ACC: 0.654} & \tabincell{l}{ACC: 0.9355} & 
Self-built & $\surd$ & $\surd$ & N/A & $\times$ & $\times$ & $\times$ \\ \cline{3-13}
&&\multicolumn{11}{c|}{ \href{https://github.com/cutz-j/FDFtNet}{github.com/cutz-j/FDFtNet}} \tabularnewline
\hline

\multirow{2}*{2020.08.10} & \multirow{2}*{\cite{jeon2020t}}  & Type \uppercase\expandafter{\romannumeral1}2 & CNN & \tabincell{l}{AUC: 0.7969} & \tabincell{l}{AUC: 0.9839} & \tabincell{c}{Self-built}  & $\surd$ & $\times$ & B & $\surd$ & $\times$ & $\times$ \\ \cline{3-13}
&&\multicolumn{11}{c|}{ \href{https://github.com/cutz-j/T-GD}{github.com/cutz-j/T-GD}} \tabularnewline
\hline

2020.08.11 & \cite{li2020sharp}  & Type \uppercase\expandafter{\romannumeral1}2 & S-MIL & \tabincell{l}{ACC: 0.7535} & \tabincell{l}{ACC: 1.0} & \tabincell{c}{Celeb-DF,\\ FF++, DFDC}  & $\times$ & $\surd$ & B & $\times$ & $\times$ & $\times$\tabularnewline
\hline

2020.08.11 & \cite{wang2020exposing}  & Type \uppercase\expandafter{\romannumeral1}1 & AdaBoost & \tabincell{l}{ACC: 0.565} & \tabincell{l}{ACC: 0.991} & \tabincell{c}{FF++}  & $\times$ & $\surd$ & B & $\surd$ & $\surd$ & $\surd$\tabularnewline
\hline

\multirow{2}*{2020.08.24} & \multirow{2}*{ \cite{chai2020makes}} & Type \uppercase\expandafter{\romannumeral1}3 & CNN & \tabincell{l}{Pre.: 0.9138} & \tabincell{l}{Pre.: 1} & \tabincell{c}{FF++}  & $\surd$ & $\surd$ & T & $\surd$ & $\times$ & $\surd$ \\ \cline{3-13}
& &\multicolumn{11}{c|}{ \href{https://github.com/chail/patch-forensics}{github.com/chail/patch-forensics}} \tabularnewline
\hline


2020.08.26 & \cite{qi2020deeprhythm} & Type \uppercase\expandafter{\romannumeral3}3 & CNN & \tabincell{l}{ACC: 0.641} & \tabincell{l}{ACC: 1.0} & \tabincell{c}{FF++, DFDC}  & $\times$ & $\surd$ & B & $\times$ & $\surd$ & $\surd$\tabularnewline
\hline

2020.08.26 & \cite{ciftci2020hearts}  & Type \uppercase\expandafter{\romannumeral3}2 & CNN & \tabincell{l}{ACC: 0.8662} & \tabincell{l}{ACC: 0.9466} & \tabincell{c}{FF++, Celeb-DF}  & $\times$ & $\surd$ & T & $\surd$ & $\times$ & $\surd$\tabularnewline
\hline

2020.08.27 & \cite{nirkin2020DeepFake} & Type \uppercase\expandafter{\romannumeral1}3 & CNN & \tabincell{l}{AUC: 0.66} & \tabincell{l}{AUC: 0.997} & \tabincell{c}{Celeb-DF,\\ FF++, DFDC}  & $\times$ & $\surd$ & B & $\surd$ & $\times$ & $\surd$\tabularnewline
\hline

2020.08.31 & \cite{li2020identification} & Type \uppercase\expandafter{\romannumeral1}1 & one-class & \tabincell{l}{ACC: 0.9795} & \tabincell{l}{ACC: 1} & Self-built & $\surd$ & $\times$ & B & $\surd$ & $\times$ & $\surd$\tabularnewline
\hline 

2020.09.04 & \cite{masi2020two} &  Type \uppercase\expandafter{\romannumeral1}2 & RNN & \tabincell{l}{AUC: 0.8659} & \tabincell{l}{AUC: 0.9912} & \tabincell{c}{Celeb-DF,\\ FF++, DFDC}  & $\times$ & $\surd$ & B & $\surd$ & $\times$ & $\times$\tabularnewline
\hline

2020.09.13 & \cite{feng2020deep} & Type \uppercase\expandafter{\romannumeral1}2  & CNN & \tabincell{l}{AUC: 0.999} & \tabincell{l}{AUC: 0.999} & \tabincell{c}{UADFV,\\ Celeb-DF, FF++}  & $\times$ & $\surd$ & B & $\surd$ & $\times$ & $\times$\tabularnewline
\hline

2020.09.16 & \cite{tariq2020convolutional} &  Type \uppercase\expandafter{\romannumeral1}2 & \tabincell{c}{Convolutional LSTM} & \tabincell{l}{ACC: 0.7412} & \tabincell{l}{ACC: 0.995} & \tabincell{c}{FF++}  & $\times$ & $\surd$ & B & $\surd$ & $\times$ & $\times$\tabularnewline
\hline

2020.09.20 & \cite{du2019towards} &  Type \uppercase\expandafter{\romannumeral4} & CNN & \tabincell{l}{ACC: 0.5905} & \tabincell{l}{ACC: 0.9991} & \tabincell{c}{Self-built}  & $\surd$ & $\surd$ & B & $\surd$ & $\times$ & $\times$\tabularnewline
\hline

\multirow{2}*{2020.10.02} & \multirow{2}*{ \cite{barni2020cnn}} &  Type \uppercase\expandafter{\romannumeral2}2 & CNN & \tabincell{l}{-} & \tabincell{l}{ACC: 0.997} & \tabincell{c}{Self-built}  & $\surd$ & $\times$ & N/A & $\surd$ & $\surd$ & $\surd$ \\ \cline{3-13}
&&\multicolumn{11}{c|}{ \href{https://github.com/ehsannowroozi/FaceGANdetection}{github.com/ehsannowroozi/FaceGANdetection}} \tabularnewline
\hline 

2020.10.12 & \cite{hu2021exposing} & Type \uppercase\expandafter{\romannumeral3}2 & N/A & \tabincell{l}{N/A} & \tabincell{l}{AUC: 0.94} & \tabincell{c}{Self-built}  & $\surd$ & $\times$ & N/A & $\times$ & $\times$ & $\surd$\tabularnewline
\hline

2020.10.22 & \cite{ganiyusufoglu2020spatio} & Type \uppercase\expandafter{\romannumeral1}2 & 3D CNN & \tabincell{l}{Pre.: 0.9429} & \tabincell{l}{Pre.: 0.9929} & \tabincell{c}{FF++, DFDC}  & $\times$ & $\surd$ & T & $\surd$ & $\times$ & $\times$\tabularnewline
\hline

2020.10.24 & \cite{dang2020detection} &  Type \uppercase\expandafter{\romannumeral1}4 & CNN & \tabincell{l}{ACC: 0.712} & \tabincell{l}{ACC: 0.984} & \tabincell{c}{Celeb-DF,\\ UADFV, DFFD} & $\surd$ & $\times$ & B & $\surd$ & $\times$ & $\surd$\tabularnewline
\hline 

2020.10.27 & \cite{qian2020thinking} &  Type \uppercase\expandafter{\romannumeral2}2 & CNN & \tabincell{l}{ACC: 0.9043} & \tabincell{l}{ACC: 0.9999} & \tabincell{c}{FF++}  & $\times$ & $\surd$ & B & $\times$ & $\surd$ & $\times$\tabularnewline
\hline

2020.10.27 & \cite{yu2020mining} &  Type \uppercase\expandafter{\romannumeral2}2 & CNN & \tabincell{l}{ACC: 0.9895} & \tabincell{l}{ACC: 0.9975} & \tabincell{c}{FF++, DFDC}  & $\surd$ & $\surd$ & B & $\surd$ & $\times$ & $\times$\tabularnewline
\hline

2020.10.28 & \cite{chen2020attentive} & Type \uppercase\expandafter{\romannumeral1}2  & CNN & \tabincell{l}{ACC: 0.9575} & \tabincell{l}{ACC: 0.9998} & \tabincell{c}{FF++, Celeb-DF}  & $\surd$ & $\surd$ & B & $\surd$ & $\times$ & $\surd$\tabularnewline
\hline

2020.11.16 & \cite{bondi2020training} & Type \uppercase\expandafter{\romannumeral4}1  & CNN & \tabincell{l}{AUC: 0.922} & \tabincell{l}{AUC: 0.998} & \tabincell{c}{Celeb-DF,\\ FF++, DFDC}  & $\times$ & $\surd$ & T & $\surd$ & $\surd$ & $\times$\tabularnewline
\hline

2020.11.19 & \cite{zhu2021face} & Type \uppercase\expandafter{\romannumeral1}2  & CNN & \tabincell{l}{ACC: 0.8731} & \tabincell{l}{ACC: 0.9972} & \tabincell{c}{FF++, DFDC,\\ Google DFD}  & $\times$ & $\surd$ & B & $\surd$ & $\surd$ & $\surd$\tabularnewline
\hline

2020.11.19 & \cite{wang2020face} & Type \uppercase\expandafter{\romannumeral1}3  & CNN & \tabincell{l}{ACC: 0.9438} & \tabincell{l}{ACC: 0.9935} & \tabincell{c}{FF++}  & $\times$ & $\surd$ & P & $\surd$ & $\times$ & $\surd$\tabularnewline
\hline

2020.12.04 & \cite{cozzolino2020id} & Type \uppercase\expandafter{\romannumeral1}2  & CNN & \tabincell{l}{AUC: 0.863} & \tabincell{l}{AUC: 0.960} & \tabincell{c}{FF++, DFDC,\\ Google DFD}  & $\times$ & $\surd$ & T & $\times$ & $\surd$ & $\times$\tabularnewline
\hline

2020.12.07 & \cite{dong2020identity} & Type \uppercase\expandafter{\romannumeral4}1  & CNN & \tabincell{l}{AUC: 0.9061} & \tabincell{l}{AUC: 0.9854} & \tabincell{c}{Google DFD,\\ FF++, Celeb-DF,\\ Vox-DeepFake}  & $\times$ & $\surd$ & B & $\surd$ & $\surd$ & $\surd$\tabularnewline
\hline

2020.12.07 & \cite{pu2020noisescope} & Type \uppercase\expandafter{\romannumeral4}1  & CNN & \tabincell{l}{F1: 0.9014} & \tabincell{l}{F1: 0.9963} & \tabincell{c}{Self-built}  & $\surd$ & $\times$ & P & $\surd$ & $\surd$ & $\surd$\tabularnewline
\hline

2020.12.08 & \cite{kukanov2020cost} & Type \uppercase\expandafter{\romannumeral4}1  & CNN & \tabincell{l}{EER: 7.16} & \tabincell{l}{EER: 6.03} & \tabincell{c}{DeepFake-TIMIT,\\ FF++}  & $\surd$ & $\times$ & P & $\times$ & $\times$ & $\times$\tabularnewline
\hline

2020.12.14 & \cite{hernandez2020deepfakeson} &  Type \uppercase\expandafter{\romannumeral3}3 & \tabincell{c}{Convolutional\\Attention Net} & \tabincell{l}{ACC: 0.944} & \tabincell{l}{ACC: 0.987} & \tabincell{c}{Celeb-DF,\\ DFDC-Preview}  & $\times$ & $\surd$ & P & $\times$ & $\times$ & $\surd$\tabularnewline
\hline

2020.12.14 & \cite{haliassos2021lips} & Type \uppercase\expandafter{\romannumeral3}2  & CNN & \tabincell{l}{AUC: 0.735} & \tabincell{l}{AUC: 0.997} & \tabincell{c}{ DeeperForensics,\\ Celeb-DF,\\ FF++, DFDC}  & $\surd$ & $\times$ & B & $\surd$ & $\surd$ & $\surd$\tabularnewline
\hline

2020.12.16 & \cite{zhao2020learning} & Type \uppercase\expandafter{\romannumeral1}3   & CNN & \tabincell{l}{AUC: 0.9438} & \tabincell{l}{AUC: 0.9998} & \tabincell{c}{Google DFD,\\ Celeb-DF,\\ DFDC, FF++,\\ DFDC-Preview,\\ DeeperForensics}  & $\times$ & $\surd$ & P & $\surd$ & $\surd$ & $\surd$\tabularnewline
\hline

2020.12.16 & \cite{guo2020fake} & Type \uppercase\expandafter{\romannumeral1}2 & CNN & \tabincell{l}{ACC: 0.9102} & \tabincell{l}{ACC: 0.9852} & \tabincell{c}{FF++}  & $\surd$ & $\surd$ & T & $\surd$ & $\surd$ & $\times$\tabularnewline
\hline 

2020.12.19 & \cite{sun2020identifying} &Type \uppercase\expandafter{\romannumeral1}2  & CNN & \tabincell{l}{AUC: 0.681} & \tabincell{l}{AUC: 0.738} & \tabincell{c}{FF++, Celeb-DF}  & $\times$ & $\surd$ & B & $\times$ & $\surd$ & $\surd$\tabularnewline
\hline

\multirow{2}*{2021.06.18} & \multirow{2}*{ \cite{zhao2021multi}} &  Type \uppercase\expandafter{\romannumeral1}2 & N/A & \tabincell{l}{AUC:0.904} & \tabincell{l}{AUC:0.993} & \tabincell{c}{Celeb-DF\\FF++, DFDC}  & $\surd$ & $\surd$ & B & $\surd$ & $\surd$ & $\surd$ \\ \cline{3-13}
&&\multicolumn{11}{c|}{ \href{https://github.com/yoctta/multiple-attention}{github.com/yoctta/multiple-attention}} \tabularnewline
\hline

\end{tabular}
\end{adjustbox}
\end{table*}

\begin{table*}[!htbp]
\scriptsize
\centering
\caption{Continued Table~\ref{tab:fake_detection_2}.}\label{tab:fake_detection_3}
\begin{adjustbox}{width=0.97\linewidth,center}
\setlength\tabcolsep{3pt} 

\begin{tabular}{|c|c|c|c|c|c|c|c|c|c|c|c|c|c|}
\hline 
\multirow{2}*{Time} & \multirow{2}*{Author} & \multirow{2}*{Method} & \multirow{2}*{Classifier} & \multicolumn{2}{c|}{Performance} & \multirow{2}*{Databases} & \multicolumn{2}{c|}{Multimedia} & \multirow{2}*{B/L} & \multicolumn{3}{c|}{Capabilities}\tabularnewline   
 & &  & &    Worst & Best  &   & Img &  Vid  &  & G & R & E\tabularnewline
\hline 
\hline

2021.06.18 & \cite{liu2021spatial} &Type \uppercase\expandafter{\romannumeral2}2  & CNN & \tabincell{l}{AUC: 0.828} & \tabincell{l}{AUC: 0.953} & \tabincell{c}{Celeb-DF, FF++, DFDC}  & $\surd$ & $\surd$ & B & $\surd$ & $\surd$ & $\surd$\tabularnewline
\hline

2021.06.18 & \cite{agarwal2021detecting} &Type \uppercase\expandafter{\romannumeral3}1  & N/A & \tabincell{l}{AUC: 0.70} & \tabincell{l}{AUC: 0.98} & \tabincell{c}{Self-build}  & $\times$ & $\surd$ & N/A & $\surd$ & $\times$ & $\surd$\tabularnewline
\hline

2021.06.18 & \cite{schwarcz2021finding} &Type \uppercase\expandafter{\romannumeral1}2  & CNN & \tabincell{l}{AUC: 0.586} & \tabincell{l}{AUC: 0.965} & \tabincell{c}{Celeb-DF, FF++, DFDC}  & $\surd$ & $\surd$ & N/A & $\surd$ & $\times$ & $\surd$\tabularnewline
\hline

2021.06.18 & \cite{wang2021representative} &Type \uppercase\expandafter{\romannumeral1}2  & CNN & \tabincell{l}{AUC: 0.934} & \tabincell{l}{AUC: 1.0} & \tabincell{c}{DFFD, Celeb-DF}  & $\surd$ & $\surd$ & T & $\surd$ & $\times$ & $\surd$\tabularnewline
\hline

2021.06.18 & \cite{li2021frequency} &Type \uppercase\expandafter{\romannumeral2}2  & CNN & \tabincell{l}{ACC: 0.890} & \tabincell{l}{ACC: 0.994} & \tabincell{c}{FF++}  & $\surd$ & $\surd$ & B & $\surd$ & $\surd$ & $\surd$ \tabularnewline
\hline

2021.06.18 & \cite{luo2021generalizing} &Type \uppercase\expandafter{\romannumeral2}2  & CNN & \tabincell{l}{AUC: 0.497} & \tabincell{l}{AUC: 0.995} & \tabincell{c}{DeeperForensics, FF++\\ Google DFD, DFDC, Celeb-DF}  & $\surd$ & $\surd$ & B & $\surd$ & $\surd$ & $\surd$\tabularnewline
\hline

\multirow{2}*{2021.06.18} & \multirow{2}*{ \cite{kim2021fretal}} &  Type \uppercase\expandafter{\romannumeral1}2 & N/A & \tabincell{l}{ACC:0.731} & \tabincell{l}{ACC:0.870} & \tabincell{c}{FF++}  & $\surd$ & $\surd$ & B & $\surd$ & $\surd$ & $\times$ \\ \cline{3-13}
&&\multicolumn{11}{c|}{ \href{https://github.com/alsgkals2/FReTAL}{github.com/alsgkals2/FReTAL}} \tabularnewline
\hline

\multirow{2}*{2021.06.18} & \multirow{2}*{\cite{chandrasegaran2021closer}} &  Type \uppercase\expandafter{\romannumeral2}2 & KNN & \tabincell{l}{ACC:0.677} & \tabincell{l}{ACC:0.999} & \tabincell{c}{Self-build}  & $\surd$ & $\surd$ & N/A & $\surd$ & $\surd$ & $\surd$ \\ \cline{3-13}
&&\multicolumn{11}{c|}{ \href{https://keshik6.github.io/Fourier-
Discrepancies-CNN-Detection}{keshik6.github.io/Fourier-
Discrepancies-CNN-Detection}} \tabularnewline
\hline

\end{tabular}
\end{adjustbox}
\end{table*}

\subsection{Spatial based Detection}\label{spatial_sec}

\begin{figure}
	\centering 
	\includegraphics[width=\linewidth]{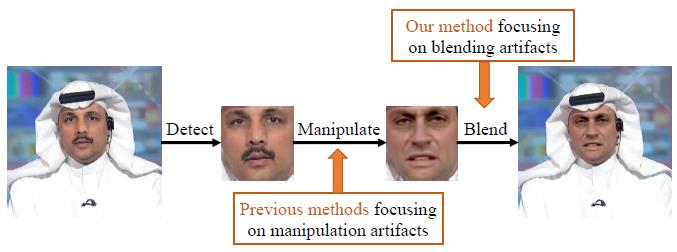}
	\caption{The difference between real and fake from the spatial domain, especially the discrepancies across the blending the boundary~\citep{li2020face}.}
	\label{fig:spatial_sample}
	\vspace{-10pt}
\end{figure}

Recently, detecting DeepFakes on the spatial domain is the most popular techniques adopted by existing studies. They observe various visible or invisible artifacts on the spatial domain for distinguishing real and fake. Figure~\ref{fig:spatial_sample} shows the potential of working on spatial domain for DeepFake detection.

\subsubsection{Image forensics based detection}

The traditional forensics-based techniques inspect the disparities in pixel-level, which is investigated by recent studies for DeepFake detection. They provide explainable clues in the detection and introduce the differences between real and fake. However, these works suffer the robustness issues when the images or videos are manipulated by simple transformations.

\cite{li2020identification} observe that the differences between synthesized faces and real faces are revealed in the chrominance components, especially in the residual domain. They propose to train a one-class classifier on real faces by leveraging the differences in the chrominance components for tackling the unseen GANs. However, their performance against perturbation attacks like image transformations is unknown.

Photo response non uniformity (PRNU) pattern is a noise pattern in a digital image caused by the light sensor in camera, which could be applied for distinguishing DeepFakes from authentic videos \citep{koopman2018detection}. Others explore utilizing the co-occurrence matrices for differentiating real and fake faces \citep{nataraj2019detecting}. The insight behind these works is obvious, but their effectiveness in tackling challenging high-quality DeepFakes is not clear.

Similarly, in tackling the fake videos, researchers also borrow the ideas from the traditional video forensic by leveraging the local motion features captured from real videos to spot the abnormality of manipulated videos \citep{wang2020exposing}. Leveraging the image/video forensic techniques is a direct idea for fighting against DeepFake by focusing on the low-level features, but they are not practical to be deployed in the wild where DeepFakes suffer known and unknown degradations.

\subsubsection{DNN-based detection}

These methods are totally data-driven by utilizing existing or designing new DNN-based models by extracting spatial features to improve the effectiveness and generalization ability of detection. 
However, these DNN-based detection methods all suffer from the adversarial attacks with additive noises and all the studies failed in evaluating their effectiveness in tackling adversarial noise attacks~\citep{carlini2020evading}. 
The existing studies by leveraging DNN to identify DeepFakes can be classified into the following three categories.

\textbf{Improving generalization abilities.} Conventional DNNs have been widely applied in detecting fake faces, but they will overfit to specific manipulation types and suffer the transferability issues where the capabilities of unseen manipulation methods are lacking. Thus, motivated by the social perception and social cognition processes of the human brain, a novel hierarchical memory network (HMN) is employed for detecting fake faces to address the transferability issues and improve the effectiveness in tacking unknown GANs \citep{fernando2019exploiting}.

Gram-Net \citep{liu2020global} improves the robustness and generalization ability to existing CNNs in discriminating synthesized fake faces by leveraging global texture features. Experimental results indicate that Gram-Net shows strong robustness against perturbation attacks like downsampling, JPEG compression, blur, and noise. Additionally, Gram-Net claims its generalization ability in tackling different GANs, which has shown promising applications in the wild.

\cite{wang2020cnn} observe that a binary classifier by leveraging a simple ResNet-50 as the backbone which is a pre-trained model with ImageNet shows strong generalization capabilities in detecting GAN-synthesized still images. Their classifier is merely trained on PGGAN database and could be well generalized to other GAN architectures like StyleGAN, StarGAN, \etc. Experiments show that they are robust against perturbation attacks by incorporating various data augmentation strategies into the training process. Obviously, detecting still images synthesized by SOTA GANs like StyleGAN is not hard.

OC-FakeDect \citep{khalid2020oc} proposes training on real faces with a one-class variational autoencoder (VAE) to detect synthesized fake faces. OC-FakeDect is vastly different from all the existing detectors which is highly dependent on collected fake faces by using binary classifiers. They claim that the approach has a good performance in generalizing across different DeepFake methods, but their robustness against perturbation attacks is unclear.

\textbf{Investigating the artifact clues.} In order to focus on the intrinsic forensics clues, image preprocessing by using smoothing filtering or noise is employed for destroying the low-level unstable artifacts in GAN-synthesized images \citep{xuan2019generalization}. Investigating the intrinsic clues could significantly improve the generalization ability of the CNN model in identifying unknown GANs.


CLRNet \citep{tariq2020convolutional} proposes a convolutional \revised{long short-term memory (LSTM)} based residual network for capturing the temporal information in consecutive frames. Transfer learning is employed for improving the generalization ability in tackling fake videos created with different synthetic methods.

SSTNet \citep{wu2020sstnet} incorporates the spatial, steganalysis, and temporal features for detecting DeepFakes. Specifically, a deep model XceptionNet is employed for extracting spatial features, a simplified XceptionNet with a constraint on the conventional filter to learn statistical characteristics of image pixels for steganalysis feature extraction, \revised{recurrent neural network (RNN)} is applied for extracting temporal features. SSTNet reveals that combing multi-modal features from a wide range of levels is a promising idea for developing powerful DeepFake detectors.

Instead of using various CNN models for addressing general object recognition tasks, researchers observed that the face recognition systems focus on learning the representation of faces. Thus, \cite{nhu2018forensics} employ a deep face recognition system to extract the face representation for building a binary classifier to detect real and fake faces. DPNet \citep{trinh2021interpretable} captures the temporal dynamic features with a carefully designed DNN to build an \textit{interpretable} DeepFake detection framework to explain why a video is predicted as fake or real. 

In DeepFake videos, the temporal artifacts across frames indicate the abnormal face in the video. A recurrent convolutional network is exploited to capture the temporal discrepancies in fake videos \citep{sabir2019recurrent}. \cite{de2020deepfake} have employed the temporal information by leveraging various typical CNNs for DeepFake detection.

\textbf{Empowering CNN models.} FDFtNet \citep{jeon2020fdftnet} provides a reusable fine-tuning network to improve the capabilities of existing CNN models (\eg, SqueezeNet, ShallowNetV3, ResNetV2, and Xception) in detecting fake images effectively. A fine-tune transformer (FTT) is designed with self-attention to extract different features from the image, then an MBblockV3 adopts different convolution and structure techniques to extract features. FDFtNet outperforms the baselines by using various CNNs. However, their robustness in tacking with unseen GANs and the evaluation on perturbation attacks is still unclear.

Inspired by the advances of deep learning, various DNN-based approaches are continuously proposed for distinguishing synthesized fake faces, such as deep transfer learning \citep{ding2020swapped,dogonadze2020deep,jeon2020t}, customized CNN \citep{dang2019face,marra2020full,cozzolino2018forensictransfer}, CNN with local features \citep{tarasiou2020extracting}, CNN with optical flow \citep{amerini2019DeepFake}, ensembled CNNs \citep{bonettini2021video,tariq2018detecting}, light-weight CNN \citep{sambhu2020detecting}, 3D CNNs \citep{ganiyusufoglu2020spatio}, a two-stream CNN using RGB space and multi-scale retinex space \citep{chen2019attention}, two-stream Faster R-CNN with features from the RGB image and noise features by using steganalysis \citep{zhou2018learning}, two-stream neural network with GoogLeNet for observing artifacts and a patch-based triplet network for capturing local noise residual \citep{zhou2017two}, multistream deep learning network for capturing the artifacts by using Face2Face reenactment \citep{kumar2020detecting}, incorporating CNN with image segmentation \citep{yu2019detecting}, capsule networks \citep{nguyen2019capsule, nguyen2019use}, pairwise learning \citep{hsu2020deep}, metric learning \citep{kumar2020detecting}, incremental learning \citep{marra2019incremental}, multiple instance learning \citep{li2020sharp}, few-shot learning \citep{mansourifar2020one}, enhanced MesoNet \citep{kawa2020note,afchar2018mesonet}, a combination with CNN and RNN \citep{guera2018DeepFake,mas2020DeepFakes}, DNN with contrastive loss \citep{hsu2018learning}, \revised{multi-attentional network \citep{zhao2021multi}}.

Though, numerous studies are working on proposing various DNN-based detection methods to discern fake faces. However, they are not robust to be deployed in dealing with real-world scenarios according to a recent study \citep{hulzebosch2020detecting}. Leveraging the power of CNN models as the backbone is a promising idea for detecting DeepFakes in the wild, however, the biggest obstacle is that the DNN models are susceptible to adversarial noise attacks.

\subsubsection{Obvious artifacts clues}

Due to the limitation of existing AI techniques, the generated DeepFakes exhibit some obvious artifacts which could be leveraged for detection by using some simple DNN models.
\cite{chai2020makes} investigate that local patches have redundant artifacts which could be used for differentiating fake faces. A fully convolutional approach is applied for training classifiers to focus on image patches. This approach can be well generalized across different network architectures, image datasets, \etc.
The discrepancy between faces and their context is another artifact clue for detecting fake faces \citep{nirkin2020DeepFake}. A face identification network is trained by using the face region to identify the person, while a context recognition network is trained by utilizing the face context like hair, ear to identify the person. Two vectors from the aforementioned two networks are compared for detecting the identity-to-identity discrepancies. This approach also has a good generalization ability across GANs.
For each individual that is speaking, their facial and head movements are always in distinct pattern \citep{agarwal2019protecting,agarwal2020detecting}. This could be exploited to protect celebrities with large historic training data.
These approaches simply leverage the artifact clues for detection without introducing any new DNN models, thus they will be invalid when the GAN is updated or the artifacts are fixed in the new version.

\subsubsection{Detection and localization}

Beyond DeepFake detection, some researchers are working on locating the manipulated regions which provides evidence for forensics and inspires future work to develop more powerful DeepFake detectors by focusing on the manipulated regions.
FakeLocator \citep{huang2020fakelocator} investigates the architecture of existing GANs and observed that the imperfection of upsampling methods exhibits obvious clues for detection and forgery localization where the manipulated area could be precisely marked. They employ an encoder-decoder network to extract the fake texture with devised gray-scale prediction map for better detection and localization. FakeLocator performs well across different GANs and shows strong generalization capabilities in unknown synthetic techniques. Furthermore, the robustness against perturbation attacks (\eg, compression, blur, noise, and low-resolution) is also promising. Locating the manipulated area provides clear explanations why the image is identified as fake.

\cite{dang2020detection} also study the localization of forgery area in fake faces by estimating an image-specific attention map. However, the estimation of the attention map fails to work in a totally unsupervised manner. The inverse intersection non-containment (IINC), a novel metric, is proposed for evaluating the performance of facial forgery localization. They also claim that forgery detection can work well in both seen and unseen synthetic techniques. The evaluation of the robustness against perturbation attacks is still lacking. The proposed attention map predicts the manipulated pixels, which provides a direct decision for determining fake faces.

Multi-task learning could also be used for classifying and locating the manipulated facial images \citep{nguyen2019multi}. Formulating fake forensics as a segmentation task to localize the manipulated region in synthesized faces is another interesting idea in fighting against DeepFakes \citep{li2019zooming,chen2020attentive}. Combing deep learning and co-occurrence matrices could also be used for the detection, attribution, and localization of GAN images \citep{goebel2020detection}.

Face X-ray \citep{li2020face} observes that a manipulated facial image always blends into an existing background image. Thus, the discrepancies across the blending the boundary could be used as a signal for detecting manipulated fake faces. Face X-ray is designed for working as both DeepFake detection and manipulated region localization.

\cite{songsri2019complement} release the first forensic localization dataset with labeled corresponding binary masks. The dataset contains real facial images, generated facial images, and partially manipulated facial images. ManTra-Net \citep{wu2019mantra} proposes an end-to-end fully convolutional network for addressing various types of image forgery, such as splicing, copy-move, removal, enhancement, and even unknown types. ManTra-Net formulates the localization problem as local anomaly detection and has a board of applications than the existing image forgery localization methods.

\subsubsection{Facial image preprocessing}\label{sec:fi_pre}

Some studies propose preprocessing the facial images before sending them to binary classifiers for discrimination. These works hope that the preprocessed DeepFakes could expose their fake textures to \revised{simple }classifiers, \revised{such as, shallow neural networks, conventional machine learning models (\eg SVM, KNN)}.

FakeSpotter \citep{wang2020fakespotter} observes that the layer-by-layer neuron behaviors provide more subtle features for capturing the differences between real and fake faces. This work provides a new insight for spotting fake faces by monitoring third-party DNN-based neuron behaviors, which could be extended to other fields like fake speech detection \citep{deepsonar}. Experimental results also show its robustness against four common perturbation attacks and its capabilities in detecting DeepFake videos. However, the generalization ability of unseen techniques is still unclear. FakeSpotter simply receives facial images as input, thus the detection result is lacking explainability.

The EM algorithm is employed for extracting the local features to represent the convolutional traces in the generated facial images \citep{guarnera2020DeepFake}. Then, some naive classifiers like \revised{K-nearest neighbors (KNN)}, \revised{support vector machine (SVM)}, and \revised{latent Dirichlet allocation (LDA)} could easily classify real and fake faces. Actually, some dimension reduction algorithms like T-SNE could non-linearly separate the real and fake faces. However, the robustness against perturbation attacks and generalization ability in different GANs are unclear.

In the real-world scenario, videos always suffer various degradations such as compression, blurring, \etc. ARENnet \citep{guo2020fake} aims at highlighting the tampered artifacts by suppressing the image content to build a practical DeepFake detector. An adaptive residuals extraction network (AREN) is designed for suppressing image content to learn prediction residuals via an adaptive convolution layer. Then, a fake face detector ARENnet is constructed by integrating AREN with CNN to deal with the fake videos suffering degradations. ARENnet claims the robustness against perturbation attacks and generalization ability in unseen GANs.

\cite{chen2021attentive} also study to improve the quality of training dataset by employing dataset preprocessing techniques to remove the false face detected in videos. They observe that preprocessing the training dataset can significantly increase the detection performance in comparison with baselines.

\cite{zhang2017automated} detect the key points in facial images and applied a descriptor to represent them for capturing the local information, then a linear classifier is applied for effective detection. This approach could be integrated into face verification systems to enhance their security.

Studies have shown that the preprocessed DeepFakes could obviously improve the detection performance. However, attackers can use other preprocessing techniques to remove the artifacts which could be used for DeepFake detection, which poses potential threats to the community.

\subsubsection{\revisedd{Technical Evolution of Spatial-based Detection}}\label{spatial_summary}

\revisedd{In this subsection, we introduce the evolution of the spatial-based DeepFake detection techniques and present the strength and weakness in detecting DeepFakes as well. Due to the low quality faces generated by the early DeepFakes, researchers first investigate the differences of real and fake faces in the spatial domain since 2017. Investigating on the spatial domain is a straightforward idea for distinguishing real and fake faces, which could borrow ideas from the traditional digital media forensics.}

\revisedd{The spatial based DeepFake detection methods aim at leveraging the power of DNN models to capture the subtle differences between real and fake in the spatial domain. The detection task can be simply formulated as a binary classification problem. Most of these studies are working towards two directions, observing more visual artifacts and developing powerful DNN models which could work in an end-to-end manner.}

\revisedd{The simple approach for detecting DeepFakes is employing the traditional image forensics techniques by inspecting the disparities at the pixel-level, such as studying the PRNU pattern which caused by the light sensor in camera~\citep{koopman2018detection} and the chrominance components~\citep{li2020identification} on real and fake faces. However, such solution usually suffers from performance decline when the DeepFakes' quality is degraded. Due to the significant progress of DNN models in various cutting-edge fields, powerful DNN models are designed or employed for spotting DeepFakes, such as FDFtNet \citep{jeon2020fdftnet} or ResNet-50 \citep{wang2020cnn}. In addition, some researchers employ shallow machine learning models like KNN and SVM to detect DeepFakes with hand-craft features. For example, \cite{guarnera2020DeepFake} apply EM to extract local features and use KNN and SVM for classification. 
Beyond DeepFake detection, \cite{huang2020fakelocator} aim to localize the GAN-based manipulated region with gray-scale prediction map, which is helpful for fine-grained DeepFake forensics. 
The early DeepFakes present abnormal visual artifacts like discrepancy between faces and their context \citep{nirkin2020DeepFake}, providing effective artifact clues for detection. We believe more and more interesting studies on the spatial domain will be proposed by our community.}

\revisedd{Overall, the spatial based DeepFakes detection is one of the most popular solutions. It works well when DeepFakes exhibit obvious visual artifacts. However, it will be not a promising way when the DeepFakes become realistic in the near future. Nevertheless, there are two critical challenges via the spatial based solution to fight against DeepFakes, \ie, poor generalization capability against unknown synthetic techniques and low robustness to adversarial attacks \citep{carlini2020evading}.}

\subsection{Frequency based Detection}\label{frequency_sec}

\begin{figure}
	\centering 
	\includegraphics[width=\linewidth]{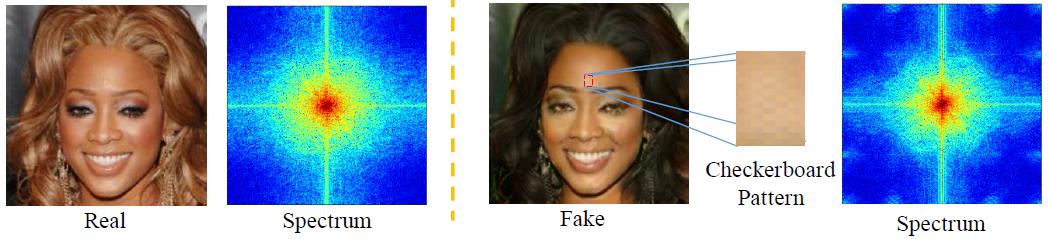}
	\caption{The difference between real and fake from the frequency domain, especially noticing the difference in their spectra \citep{zhang2019detecting}.}
	\label{fig:frequency_sample}
	\vspace{-10pt}
\end{figure}

\revised{Beyond distinguishing real and fake from the spatial domain, some studies are working on exploiting the differences between real and fake from the frequency domain. Figure~\ref{fig:frequency_sample} represents the potentials of employing the frequency artifacts for detecting DeepFakes, where the GAN-based manipulation introduces invisible artifacts in the frequency domain.}

\subsubsection{GAN-based artifacts}\label{gan_artifacts}

Instead of examining the visual artifacts, some researchers are working on investigating the imperfection design of existing GANs, which provides obvious signals for differentiating real and fake faces. They are normally working on the frequency domain, \revised{but there are generalized artifacts of existing GANs}.

\cite{mccloskey2019detecting} investigate the architecture of the generator model and observed that the internal value of the generator is normalized which limits the frequency of saturated pixels. Then, a simple SVM-based classifier is trained to measure the frequency of saturated and under-exposed pixels in each facial image for discriminating fake faces.

AutoGAN \citep{zhang2019detecting} identifies a unique artifact in GANs which is introduced due to the upsampling design of common GAN pipelines. Then, a GAN simulator is proposed for simulating the artifacts without accessing pre-trained GANs to improve the generalization ability of existing detectors. The artifacts are manifested as replications of spectra in the frequency domain. Finally, a classifier is trained by using the frequency spectrum for discriminating GAN-synthesized fake faces. They claim that the observed GAN-based artifacts could generalize well in unseen synthetic techniques with similar architectures. However, their robustness against perturbation attacks is not explored.

\cite{yu2019attributing} first introduce the GAN fingerprints for classifying the images as real or synthesized with GANs. The GAN fingerprints could be further utilized for predicting the source of images. Study has shown that small differences in GAN training could result in distinct GAN fingerprints. However, the fingerprints could be easily destroyed by simple perturbation attacks like blur, JPEG compression, \etc.
Other studies \citep{marra2019gans} also leverage the GAN fingerprints for discriminating GAN-synthesized fake faces. The GAN artifacts are a promising clue for detection, however the artifacts could be easily corrupted with some simple image transformations like shallow reconstruction with \revised{principal component analysis (PCA)} \citep{huang2020fakepolisher}.

\subsubsection{\revised{Frequency domain feature}} \label{frequency_domain}

The differences between real and synthesized fake faces could also be revealed in the frequency domain. \revised{Here, we mainly introduce the studies simply employing frequency domain as features for differentiating real and fake. These methods are often failed in tackling unknown GAN-synthesized DeepFakes, which are vast different from the aforementioned methods in subsection \ref{gan_artifacts}.}

\cite{frank2020leveraging} comprehensively investigate the artifacts revealed in the frequency domain across different GAN architectures and datasets. They observe that severe artifacts are introduced due to the upsampling techniques in GANs. Experiments demonstrate that a classifier with a simple linear model and a CNN-based model could both achieve promising results on the entire frequency spectrum. Furthermore, the classifier trained on the frequency domain is robust against common perturbation attacks (\eg blurring, cropping) and tackles the future unseen GANs~\citep{frank2020leveraging}.

FGPD-FA \citep{bai2020fake} extracts three types of features (\eg, statistical, oriented gradient, and blob) in the frequency domain for differentiating real and fake faces. F$^3$-Net \citep{qian2020thinking} considers two complementary frequency-aware clues for detection, namely frequency-aware pattern from frequency-aware image decomposition, and local frequency statistics. Specifically, \revised{discrete cosine transform (DCT)} is applied for frequency-domain transformation. Finally, a two-stream collaborative learning framework collaboratively learns the two frequency clues and achieves considerable performance in low-quality DeepFake video detection.

\cite{barni2020cnn} propose exploiting the inconsistencies among spectral bands, then a CNN model is trained with cross-band co-occurrence matrices and pixel co-occurrence matrices for discriminating real and fake faces.
\cite{yu2020mining} explore the channel difference image (CDI) and spectrum image (SI) to work as intrinsic clues for distinguishing images generated with a camera and manipulated with AI techniques. Octave convolution (OctConv) \citep{chen2019drop} is leveraged for capturing the frequency domain to learn the intrinsic feature from CDI and SI to determine fake faces. These two intrinsic clues are claimed to generalize well in unseen manipulations. \revised{To improve the transferability of face forgery detection method across unseen 
synthetic techniques, \cite{liu2021spatial} combine spatial image and phase spectrum to capture the up-sampling artifacts in existing GANs for aiding detection.} \revisedd{\cite{masi2020two} propose a two-branch network for amplifying the artifacts in the synthesized faces by combing clues from the color domain and frequency domain. Furthermore, the two-branch network has claimed a good performance across datasets.}
%
%
Actually, leveraging the frequency domain to distinguish real and fake faces is widely applied in recent studies \citep{durall2019unmasking, bonettini2021use,guarnera2020preliminary}.

\subsubsection{Technical Evolution of Frequency-based Detection}\label{frequency_summary}

\revisedd{In this subsection, we introduce the evolution of the frequency-based DeepFake detection techniques and present the strength and weakness in detecting DeepFakes as well. Beyond detecting DeepFakes via the spatial information, exploring the artifacts of DeepFakes in the frequency domain is another effective solution. The frequency based DeepFake detection methods identify DeepFakes via artifacts in the frequency domain, benefiting higher generalization ability.}

The frequency based DeepFake detection methods mainly rely on two kinds of information, \ie, \textit{the artifacts in the spectra introduced by GAN} and \textit{frequency domain features of real or fake faces}.
For the first kind solution, researchers observe that the GAN-synthesized facial images exhibit obvious artifacts in the spectra, which provides effective clues for detection with high generalization. Figure~\ref{fig:frequency_sample} visualizes the spectra maps of real and fake face \citep{zhang2019detecting}, respectively. 
Then, a series of studies try to mine GAN-based artifacts of fake faces effectively and achieve better generalization capabilities when addressing unknown synthetic techniques.
In particular, \cite{mccloskey2019detecting} measure the frequency of saturated and under-exposed pixels to discriminate real and fake faces. \cite{zhang2019detecting} identify the artifacts introduced by GAN due to the common used upsampling operation. The above two methods claim their competitive generalization capabilities in unknown DeepFake techniques. 
For the second solution, features in the frequency domain is leveraged as clues for detection. For example, \cite{frank2020leveraging} employ the entire frequency spectrum as features for differentiating fake. \cite{barni2020cnn} exploit the inconsistencies among spectrum bands to discriminate fake.

The frequency based DeepFake detection method could generalize well on unknown synthetic techniques, but they are not robust to various image degradations, such as image compression, reconstruction, \etc~\citep{huang2020fakepolisher}. Thus, they are less practical in the real-wold scenario where known and unknown image degradations are common. As a result, more robust frequency based methods should be developed, which is critical for detectors to be further deployed in the wild.

\subsection{Biological Signal based Detection}\label{biological_sec}

\begin{figure}
	\centering 
	\includegraphics[width=\linewidth]{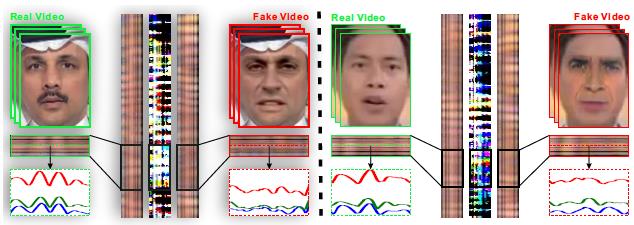}
	\caption{The difference between real and fake from the biological signal domain, especially the colorful motion-magnified spatial-temporal (MMST) maps between them~\citep{qi2020deeprhythm}.}
	\label{fig:biological_sample}
	\vspace{-10pt}
\end{figure}

Real still facial images and videos are produced with cameras, which are natural compared to the synthesized fake faces. \revised{The biological signal exhibits a clear signal for distinguishing real and fake. In general, the biological signals are existed in both real videos and synthesized fake videos. However, the biological signals revealed in the real faces videos are natural and realistic. Unfortunately, in the fake videos, the biological signals are generated with low-quality and most of time the perceptual biological signals are disappeared, such as the consistency between visual and audio. Figure~\ref{fig:biological_sample}} shows the sample of biological signals which could served as clues for DeepFake detection. These biological signals can be classified into the following categories.

\subsubsection{Visual-audio inconsistency}

For DeepFake video, combining visual and audio to identify the inconsistency in fake faces is a new insight for distinguishing DeepFakes. These methods can well explain why the video is fake.
A Siamese network is employed for modeling the visual and audio in videos with a combination of two triplet loss functions for measuring the similarity \citep{mittal2020emotions}. Specifically, one loss function is designed for computing the similarity between visual and audio, the other loss function is devised for calculating the effect cues, like perceived emotion. Experiments show that it outperforms conventional DNN-based methods in detecting fake videos.
Lip-sync is a typical DeepFake by generating a person's mouth to be consistent with a person's speech. With the basic insight that the dynamics of mouth shape are sometimes inconsistent with a spoken phoneme due to the highly compelling circumstances. Specifically, the lips have to be closed when spoken some words that begin with $M, B, P$. However, this is violated in fake videos. Researchers leveraged this clue for detecting lip-sync DeepFakes \citep{Agarwal_2020_CVPR_Workshops}.
Modality dissonance score (MDS) is proposed for measuring the audio-visual dissonance in videos \citep{chugh2020not}. Specifically, MDS is based on \emph{contrastive loss} which enforces the distance between visual and audio to be closer for real, and further for synthesized fake video. Additionally, MDS can be used for temporal forgery localization which identifies the tampered segment in the video.
However, \cite{korshunov2018deepfakes} investigate several baselines for evaluating existing studies in DeepFake detection including lip-sync inconsistency detection. They observe that detecting the inconsistency of lip-sync is not effective for fighting DeepFakes. They also release a public dataset for the community.

\subsubsection{Visual inconsistency}

Visual inconsistency indicates that the synthesized faces are not natural, especially the shape, facial features, and landmarks of faces. 
\cite{li2018exposing} observe that the synthesized fake faces are always in fixed sizes due to the limitation of computation resources and the production time of DeepFake algorithms. The fixed size of synthesized faces leaves artifacts in warping to match the source face, which can be employed for DeepFake detection. Then, a CNN model is trained for detecting the artifacts. The lack of eye blinking is another telltale sign for exposing DeepFakes \citep{li2018ictu}. A CNN combined with a recursive neural network is trained for distinguishing the eye state. The mismatched facial landmarks in fake faces are invisible to human eyes, but they can be easily revealed from head poses estimated from 2D landmarks \citep{yang2019exposing, yang2019exposing_1}. A naive SVM classifier is finally trained for capturing the differences between estimated head poses, which is further employed for DeepFake detection. Visual artifacts such as eyes, teeth, facial contours will be an important clue for exposing DeepFakes \citep{matern2019exploiting}. The inconsistent corneal specular highlight between two eyes is another clue for exposing the GAN-synthesized faces \citep{hu2021exposing}. This inconsistency is mainly due to the lack of physical/physiological constraints in the existing popular GANs.
These methods are all based on the observation that the fake faces exhibit obvious artifacts to human eyes, especially the inconsistencies that appeared in the face compared with real faces. They provide strong guarantees to explain the decision in distinguishing real or fake, but they will be invalid when advanced GANs are proposed. Furthermore, their robustness against perturbation attacks is unclear.

\subsubsection{Biological signal in video}

Biological signals in the video are not easily replicable. In FakeCatcher, six different biological signals are extracted to exploit the spatial and temporal coherence for authenticating real videos taken by the camera \citep{ciftci2020fakecatcher}.
Studies have shown that the heart rate could be used for detecting fake videos, however, obtaining the heart rate from videos is a time-consuming task. \cite{fernandes2019predicting} use neural ordinary differential equation (Neural-ODE) \citep{chen2018neural} trained on the original videos to predict the heart rate of testing videos.
DeepRhythm also exposes DeepFake videos by monitoring the heartbeat rhythms \citep{qi2020deeprhythm}. Specifically, they develop motion-magnified spatial-temporal representation (MMSTR) to video for highlighting the heart rhythm signals. Finally, a dual-spatial-temporal attentional network is designed for detecting fake video based on the output of MMSTR. 
DeepFakesON-Phys \citep{hernandez2020deepfakeson} also leverages heart rate for DeepFake detection by using remote photoplethysmography (rPPG) to illustrate the presence of blood flow by observing the subtle color changes in human skins. A convolutional attention network (CAN) is proposed for extracting the spatial and temporal information from video frames to detect DeepFake video. Beyond the DeepFake detection, PPG could be used for discovering the generative model which is used for generating DeepFake \citep{ciftci2020hearts}.
In detecting DeepFake videos, the biological signal exposed by the heart rate provides promising clues for detection. This will be a promising idea for dealing with future advanced GANs, since the subtle biological characteristics are a challenge for synthesis.

\subsubsection{\revisedd{Technical Evolution of Biological Signal based Detection}}\label{biological_summary}

In this subsection, we introduce the evolution of the biological signal based DeepFake detection techniques and present the strength and weakness in detecting DeepFakes as well. With the rapid development of deep synthesis techniques, the fake images would be perfectly synthesized without exposing any artifacts in both the spatial and frequency domains in the near future, which would pose more challenges for DeepFake detection. 
Recently, some researchers are working to explore the biological signals in the facial videos to serve as effective fake indicators since the signals are not natural and unrealistic in fake videos.
%
%

The existing biological signal based DeepFake detection methods utilize biological signals that are broken and could not be easily replicated by state-of-the-art DeepFake techniques. 
Early works study the irregular eye blinking~\citep{li2018ictu}, the mismatch facial landmarks~\citep{yang2019exposing, yang2019exposing_1}, and the fixed size of synthesized faces~\citep{li2018exposing}, \etc. 
Nevertheless, the above visual inconsistencies could be easily removed in the advanced DeepFakes. 
In addition to the visual information, the audio of the video is also an important clue for DeepFake detection. 
Specifically, the inconsistency between visual and audio is common in fake videos. However, \cite{korshunov2018deepfakes} observed that the simple lip-sync is not enough for accurate DeepFake detection. 
Then, some studies are working on how to measure the similarity between visual and audio \citep{mittal2020emotions,chugh2020not} and further explore strong visual-audio inconsistency signals for DeepFake detection \citep{Agarwal_2020_CVPR_Workshops}. Moreover, some works also exploit the subtle color changes in human skins introduced by the normal heartbeat to authentic real videos \citep{qi2020deeprhythm,hernandez2020deepfakeson}. We believe that more and more interesting and robust biological signals will be observed for discriminating DeepFakes in the wild.

Overall, In the near future, the DeepFake could be realistic where the spatial and frequency based detection methods could hardly exhibit noticeable and detectable artifacts by human eyes and machines.
As a result, the biological signals would be a more effective solution for fighting against DeepFake that could be deployed in the real world. 
Nevertheless, the solution might be invalid when the biological signals are enhanced manually, and exploring more informative biological signals would be the most promising one for the future detection.

\subsection{Other DeepFake Detectors}\label{sec:DF_detection_other}

Besides the aforementioned three types, some studies cannot be classified into any of them. Here, we introduce them with an independent subsection.
\cite{fraga2020fake} provide a comprehensive overview by leveraging distributed ledger technologies (DLT) to combat digital deception. \cite{hasan2019combating} also leverage blockchain to trace and track the source of multimedia which provides insight for combating DeepFake videos.
Instead of a focus on the multimedia self, FakeET \citep{gupta2020eyes} explores to leverage the user behavior clues for DeepFake detection, specifically the {eye-gaze}.
\cite{tolosana2020deepfakes} explore the role of different facial regions in contributing to the DeepFake detection. They find that the artifacts which exist in the specific facial region could improve the detection performance by a large margin than the entire face. Similarly, \cite{du2019towards} observe that concentrating on the forgery region could help for DeepFake detection.
\cite{maurer2000authentication} approaches the DeepFake detection as a hypothesis testing problem and presents a generalizable statistical framework based on the information-theoretic study of authentication.

\subsection{Summary of DeepFake Detection Methods}\label{sec:sum_det}

In this section, we use a long table and a chord diagram to summarize the existing DeepFake detection methods \revisedd{and a fishbone diagram to show the evolution of the three DeepFake detection techniques}.

\begin{figure}
\tiny
\setlength\tabcolsep{3pt} 
\caption{(L) Summary of various types of DeepFake detection methods, including the type ID and the name of each type. (R) The proportion of various types of DeepFake detection methods in our collected DeepFake detection papers.} 
\vspace{-5pt}
\label{tab:method_type}
\begin{minipage}{0.48\linewidth}
    \centering
    \begin{tabular}{|l|l|}
    \hline 
    Type ID & Name \tabularnewline
    \hline 
    \hline 
    \cellcolor{googleblue!60}Type \uppercase\expandafter{\romannumeral1}1 & Image Forensics based Detection \tabularnewline
    \hline
    \cellcolor{googleblue!60}Type \uppercase\expandafter{\romannumeral1}2 & DNN-based Detection \tabularnewline
    \hline
    \cellcolor{googleblue!60}Type \uppercase\expandafter{\romannumeral1}3 & Obvious Artifacts Clues \tabularnewline
    \hline
    \cellcolor{googleblue!60}Type \uppercase\expandafter{\romannumeral1}4 & Detection and Localization\tabularnewline
    \hline
    \cellcolor{googleblue!60}Type \uppercase\expandafter{\romannumeral1}5 & Facial Image Preprocessing\tabularnewline 
    \hline\hline
    \cellcolor{googlegreen!60}Type \uppercase\expandafter{\romannumeral2}1 & GAN-based Artifacts \tabularnewline
    \hline
    \cellcolor{googlegreen!60}Type \uppercase\expandafter{\romannumeral2}2 & Frequency Domain\tabularnewline 
    \hline\hline
    \cellcolor{googleyellow!60}Type \uppercase\expandafter{\romannumeral3}1 & Visual-audio Inconsistency\tabularnewline
    \hline
    \cellcolor{googleyellow!60}Type \uppercase\expandafter{\romannumeral3}2 & Visual Inconsistency\tabularnewline
    \hline
    \cellcolor{googleyellow!60}Type \uppercase\expandafter{\romannumeral3}3 & Biological Signal in Video \tabularnewline 
    \hline\hline
    \cellcolor{googlered!60}Type \uppercase\expandafter{\romannumeral4}1 & Others \tabularnewline
    \hline
    \end{tabular}
\end{minipage}
\hfill
\begin{minipage}{0.48\linewidth}
    \centering
    \includegraphics[width=\linewidth]{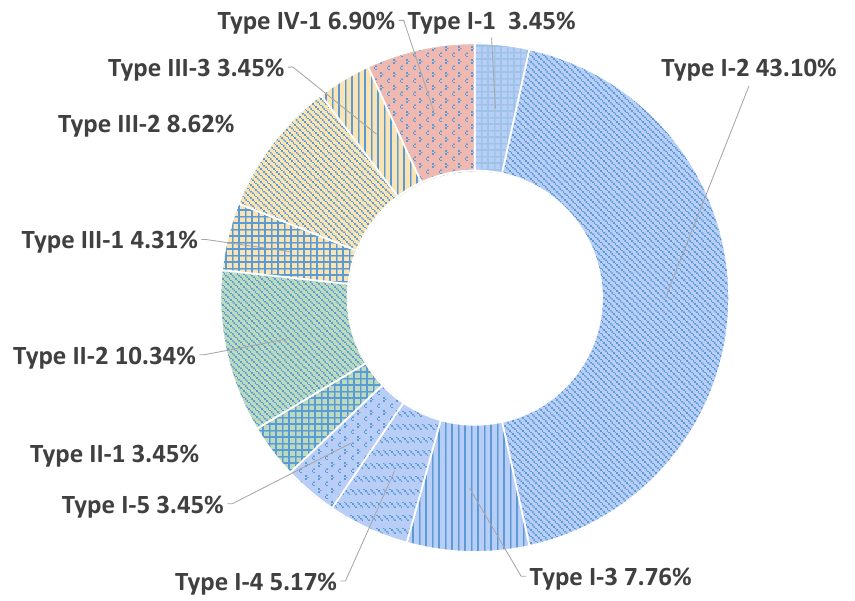}
\end{minipage}
\vspace{-10pt}
\end{figure}

Tables \ref{tab:fake_detection_1}, \ref{tab:fake_detection_2}, and \ref{tab:fake_detection_3} tabulate the summary of DeepFake detection methods, where Figure \ref{tab:method_type} gives the meaning of type ID and the proportions. In these tables, we mainly show the method type, the adopted classifier, the claimed performance, compared baselines, and its capabilities with regard to the generalization capabilities in tackling unseen DeepFakes, the robustness against various attacks, and whether provides explainable detection results.

In analyzing the Tables \ref{tab:fake_detection_1}, \ref{tab:fake_detection_2}, and \ref{tab:fake_detection_3}, we can gather the following interesting findings. Due to the powerful capabilities of DNN model, CNN models are served as the most popular backbone in the DeepFake detection classifiers. However, linear machine learning models like KNN are rarely employed in detection. In employing the evaluation metrics, ACC and AUC are the two popular metrics for evaluating the performance of DeepFake detection methods. Compared with DeepFake videos, the still images are easier to be detected by various DeepFake detectors. Researchers tend to evaluate their method on public DeepFake videos, rather than build their own synthesized-images datasets for evaluation due to the lack of public fake image datasets. The existing studies claimed their effectiveness in detecting DeepFakes with high confidence, however most of them failed in evaluating their effectiveness in tackling unseen DeepFakes and their robustness against image/video transformations, which is critical for a detector deployed in the wild. Additionally, these methods failed in providing evidence to introduce the differences between real and fake, thus the explainability is limited in existing studies.

\begin{figure*}
	\centering 
    \includegraphics[width=\linewidth]{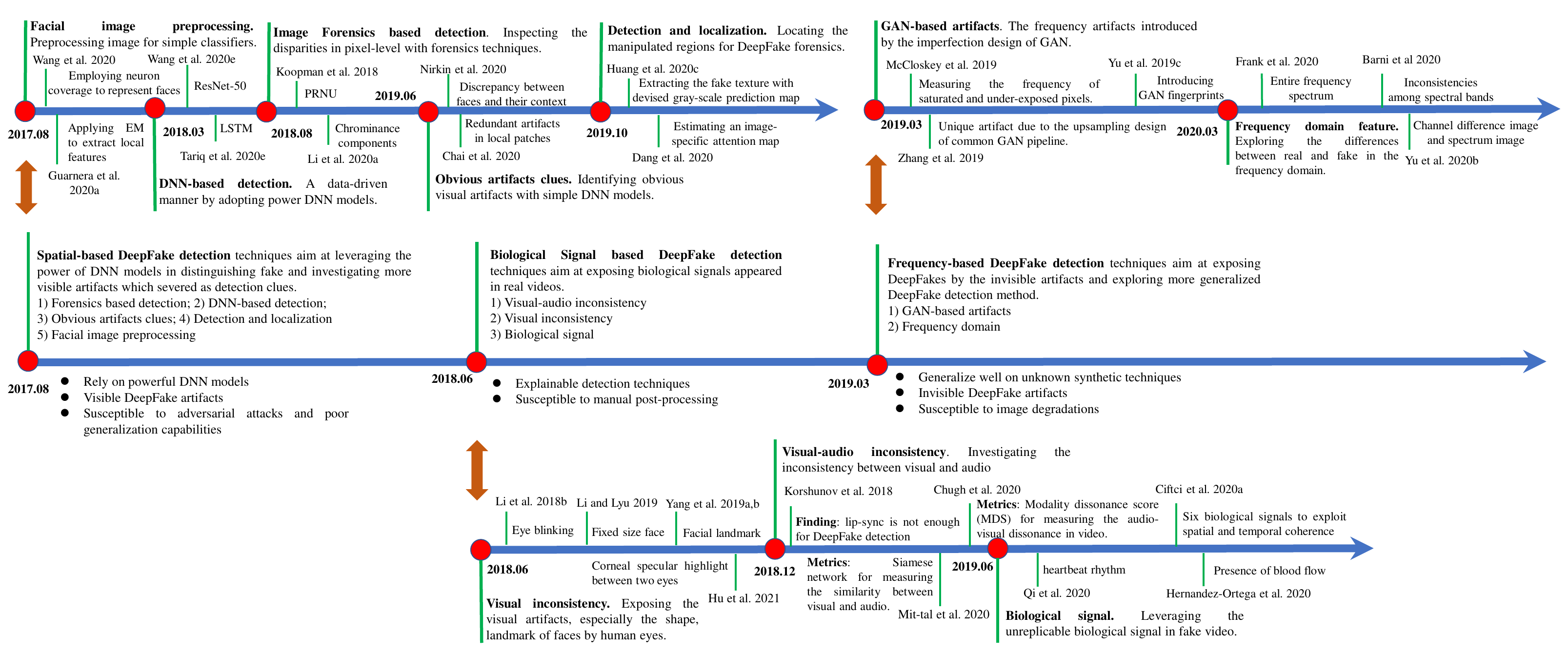}
	\caption{\revised{The evolution of DeepFake detection techniques with a fishbone diagram. In the main fishbone, the weakness and strengths of each detection method are presented as well. For each DeepFake detection method in the sub-fishbone diagram, the milestone studies are added for presenting the significant progress, especially their novelty on technical, the problem addressed, and new insight for defending DeepFakes.}}
	\label{fig:fish_bone_detection}
\end{figure*}

\revisedd{Figure~\ref{fig:fish_bone_detection} presents the milestone studies of DeepFake detection with a fishbone diagram. In investigating the three classical DeepFake detection techniques, we observed that there are two critical challenges that should be addressed for the future DeepFake detection techniques. The first challenge is that the fake textures for DeepFake detection might be corrupted or intentionally removed. The second one is that the quality of synthesized images would be further improved with the development of synthetic techniques. As a result, the community should develop more robust models against various degradations to capture the subtle differences between real and fake faces and investigate more long-standing clues to detect unknown DeepFake synthetic techniques.}


\section{Battleground}\label{sec:battle}

\begin{figure*}[tbp]
	\centering 
    \setlength{\belowcaptionskip}{-0.3cm}  
    \includegraphics[width=0.98\linewidth]{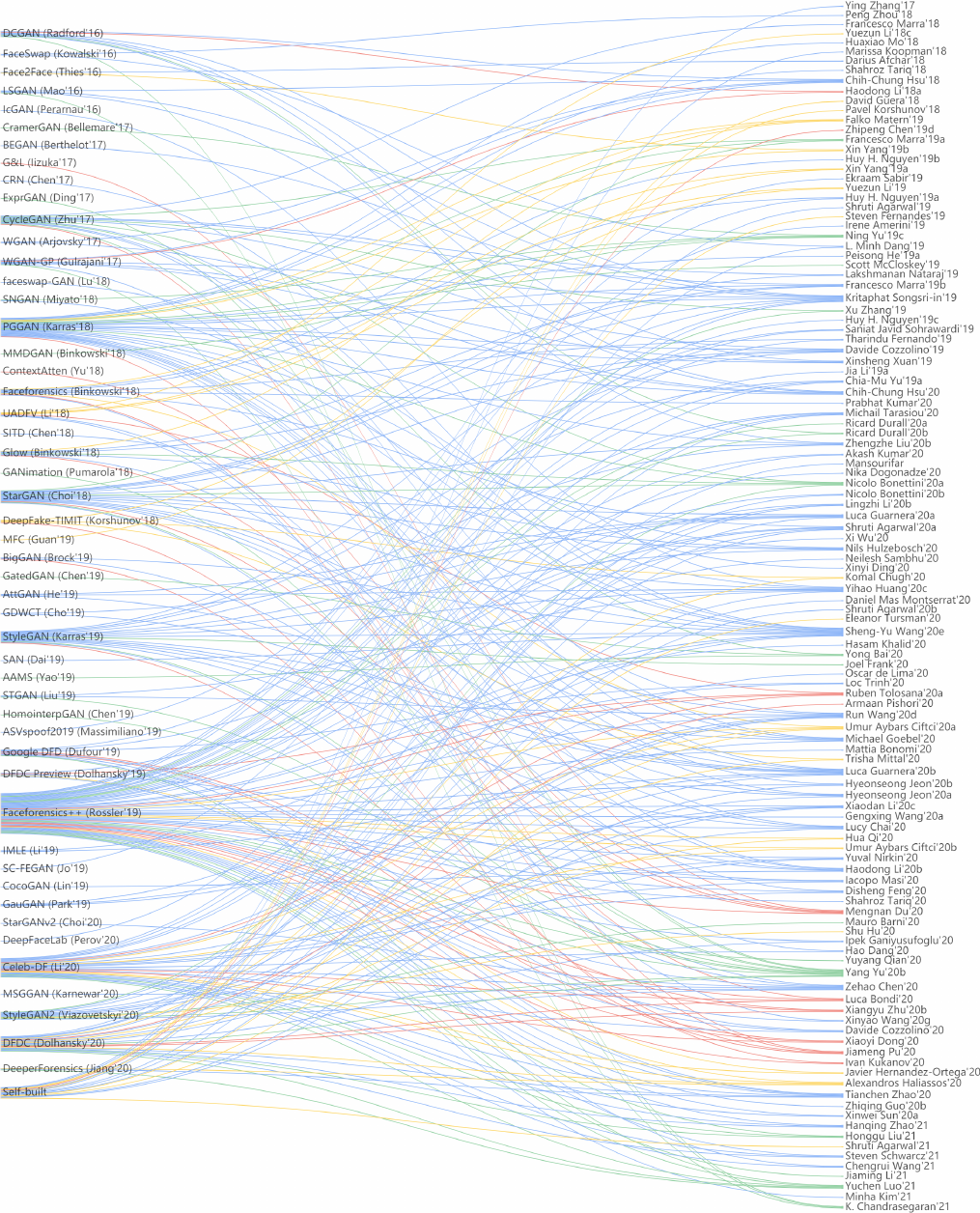}
	\caption{Battleground diagram between DeepFake generation and detection. The Sankey diagram shows the interactions between various DeepFake detection methods (right column) and various DeepFake generation methods (left column). Both of the generation and detection methods are sorted by the release time and labeled with the corresponding years (same as the order in Tables \ref{tab:paper_information}, \ref{tab:fake_detection_1}, \ref{tab:fake_detection_2} and \ref{tab:fake_detection_3}). Four colors represent the different types of detection methods introduced in Tables \ref{tab:fake_detection_1}, \ref{tab:fake_detection_2}, and \ref{tab:fake_detection_3}: \textcolor{googleblue!80}{Blue is Type-I (spatial based) methods}, \textcolor{googlegreen!80}{green is Type-II (frequency based) methods}, \textcolor{googleyellow!80}{yellow is Type-III (biological signal based) methods}, and \textcolor{googlered!80}{red is Type-IV (others) methods}. Interactive diagram is available at \scriptsize{\url{http://www.xujuefei.com/dfsurvey}}.}
	\label{fig:battle}
\end{figure*}

In the previous two sections, we have discussed recent advances in DeepFake generation methods (Section~\ref{sec:gen}) and DeepFake detection methods (Section~\ref{sec:det}), respectively. The two parties naturally form a battleground, where the ``offenders'' or the ``adversaries'' are the DeepFake generation methods, and the ``defenders'' are the DeepFake detection methods. By illustrating and visualizing the battleground, we hope to gain insights and knowledge about the most current battling landscape and interactions between DeepFake generation and detection methods. 
We believe that incremental but continuous scientific progresses can be made through the competition between adversaries and defenders, and new observations can be obtained when defeating the other side. It is the unceasing battling between the two parties that will most likely make the meaningful progress to push the field (\ie, high-fidelity generation of DeepFakes as well as high-performance detection of DeepFakes) forward possible. 

Among all $318$ DeepFake-related papers surveyed so far, we have kept the important ones in tables across Section~\ref{sec:gen} and \ref{sec:det}. As previously tabulated in Table~\ref{tab:paper_information}, we have surveyed $83$ DeepFake generation methods in Section~\ref{sec:gen}. As tabulated in Table~\ref{tab:fake_detection_1}, \ref{tab:fake_detection_2} and \ref{tab:fake_detection_3}, we have surveyed $117$ DeepFake detection methods in Section~\ref{sec:det}. In order to create a full map of the battleground, for each of the DeepFake detection methods, we 
aim to know which DeepFake generation method the detector attempted to counter, \ie, to perform DeepFake detection on. 
In the Sankey diagram \citep{sankeywiki} shown in Figure~\ref{fig:battle}, we have chronologically arranged various surveyed DeepFake generation methods (including datasets) on the left column and the surveyed DeepFake detection methods on the right column. A curve connecting the node $A$ on the left and the node $B$ on the right means that DeepFake detection method $B$ has evaluated and reported detection results on the DeepFake generation method $A$ in its paper.
After all the nodes are connected by traversing the $83\times117$ generation-detection relationships, Figure~\ref{fig:battle} now presents the status of the DeepFake generation-detection battleground. 
As the out degree for each node shown in the figure, we can tell how popular each DeepFake generation method or DeepFake detection method is. For example, the FaceForensics++ has a large out degree, which means that it is evaluated by a large number of DeepFake detection methods. 
Similarly, a big clustered connections on the right side indicates that a particular DeepFake detection method (\eg, Sheng-Yu Wang's method) has been evaluated extensively across various DeepFake generation methods.
The colorful curves represent the different types of detection methods introduced in Tables \ref{tab:fake_detection_1}, \ref{tab:fake_detection_2}, and \ref{tab:fake_detection_3} as well as tabulated in Figure~\ref{tab:method_type}: \textcolor{googleblue!80}{Blue is Type-I (\ie, spatial based) methods}, \textcolor{googlegreen!80}{green is Type-II (\ie, frequency-based) methods}, \textcolor{googleyellow!80}{yellow is Type-III (\ie, biological signal based) methods}, and \textcolor{googlered!80}{\ie, red is Type-IV (others) methods}. The `self-built' methods on the left-side bottom of the battleground represent the nameless methods 
addressed by the detection methods on the right side.



\begin{figure}
	\centering 
	\includegraphics[width=0.49\linewidth]{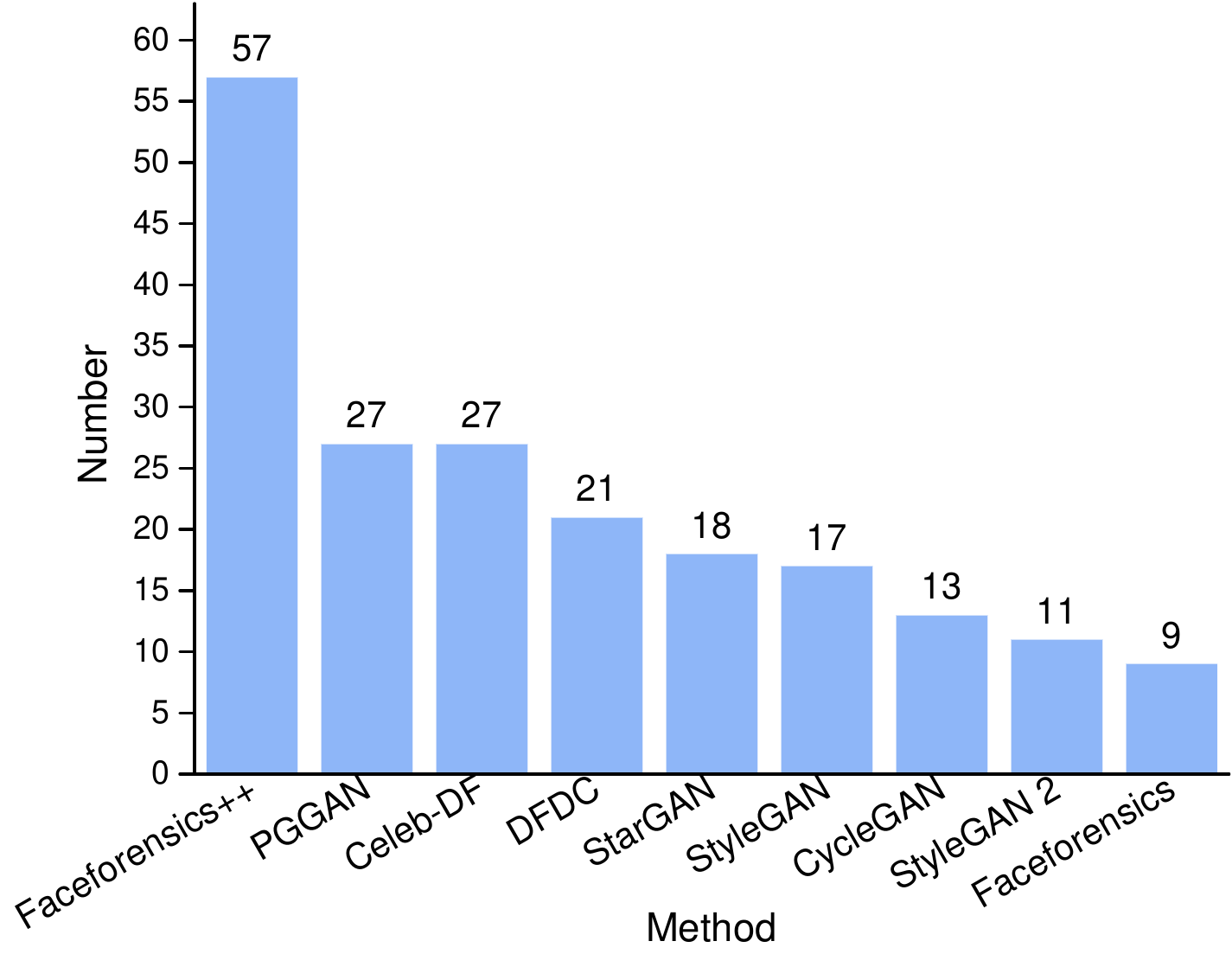}
	\includegraphics[width=0.49\linewidth]{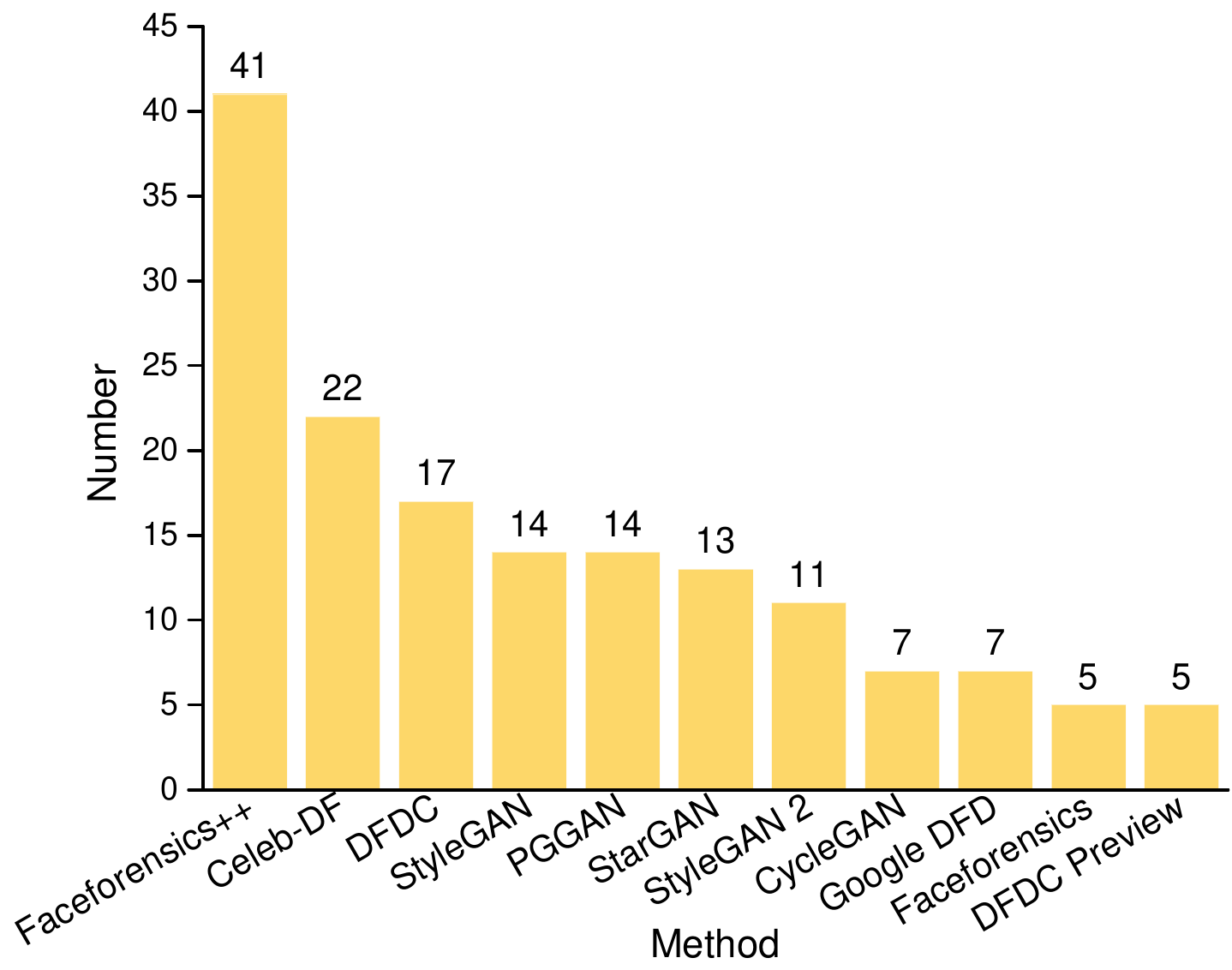}
	\caption{(L) Top-9 most popular DeepFake generation methods or datasets based on the battleground. (R) 2020's Top-11 most popular DeepFake generation methods or datasets based on the battleground.}
	\label{fig:battle_generation_rank}
\end{figure}

Figure~\ref{fig:battle_generation_rank} shows the top-9 most popular DeepFake generation methods or datasets based on the topology of the battleground figure, as well as the top-11 most popular DeepFake generation methods or datasets in the year 2020 alone. As expected, systemically organized DeepFake datasets such as FaceForensics++ \citep{rossler2019FaceForensics++}, Celeb-DF \citep{li2020celeb}, DFDC \citep{dolhansky2020deepfake} as well as open-sourced high-fidelity face generation methods such as PGGAN \citep{karras2017progressive}, StarGAN \citep{choi2018stargan}, StyleGAN \citep{karras2019style}, \etc, are on the top of the list.


Based on the above discussion about the most popular or most widely evaluated DeepFake generation methods or datasets, we have made some interesting observations: (1) the surveyed DeepFake detectors perform more detection experiments on DeepFake images than on DeepFake videos; (2) only a tiny portion of the surveyed detection methods work on both DeepFake image and video detection tasks; (3) for those detectors for both DeepFake image and video detection, most of them focus on the latest high-fidelity image-based DeepFakes while on the less state-of-the-art video-based DeepFakes, although both modalities are concurrently accessible. 
This can be partially attributed to the fact that video-based DeepFake datasets are more scarce, and/or the latest ones are much more challenging to tackle. 

\begin{figure}
	\centering 
	\includegraphics[width=\linewidth]{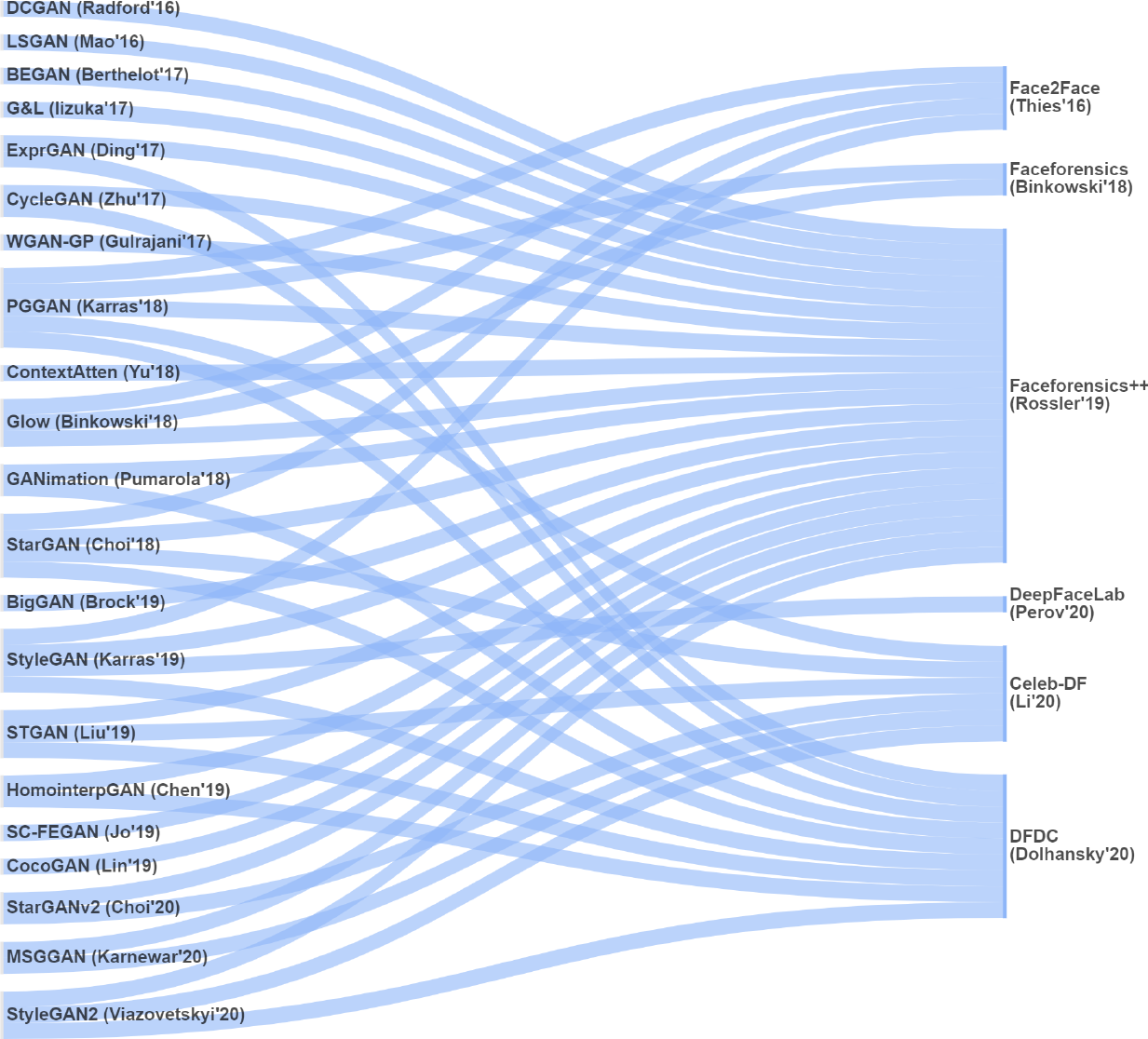}
	\caption{Relation pairs of the image- and video-based DeepFake generation methods that are simultaneously evaluated by some DeepFake detection methods. Interactive diagram is available at \scriptsize{\url{http://www.xujuefei.com/dfsurvey}}.}
	\label{fig:Image-Video-Relation}
\end{figure}

We try to capture this phenomenon through the Sankey diagram in Figure~\ref{fig:Image-Video-Relation}, where only $10$ of the surveyed $117$ DeepFake detection methods have attempted the DeepFake detection on both the image and video modalities. A curve connecting a node $A$ on the left column and a node $B$ on the right column means that a particular DeepFake detector evaluated on the image-based DeepFake generation method $A$ has also been evaluated on video-based DeepFake generation method $B$, as reported in its paper.
%
%
\begin{figure*}[tbp]
	\centering 
    \setlength{\belowcaptionskip}{-0.3cm}  
	\includegraphics[width=0.8\linewidth]{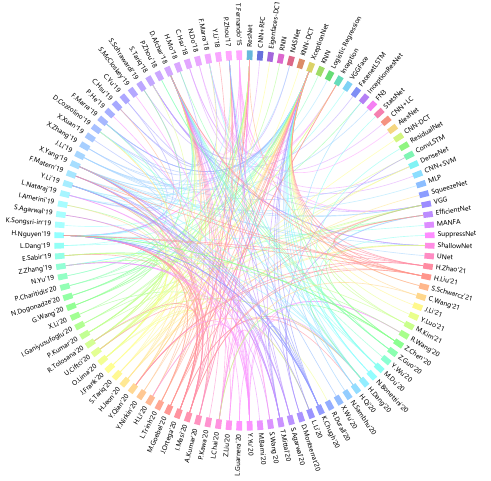}
	\caption{A chord diagram represents the comparison among the existing detection methods. The node indicates the method for DeepFake detection and the link represents that one of the work is served as the baseline in the evaluation. The baselines include typical CNN models and the works with/without the peer review. An interactive diagram is available at \scriptsize{\url{http://www.xujuefei.com/dfsurvey}}.}
	\label{fig:cd}
\end{figure*}

\begin{figure}
	\centering 
	\includegraphics[width=0.49\linewidth]{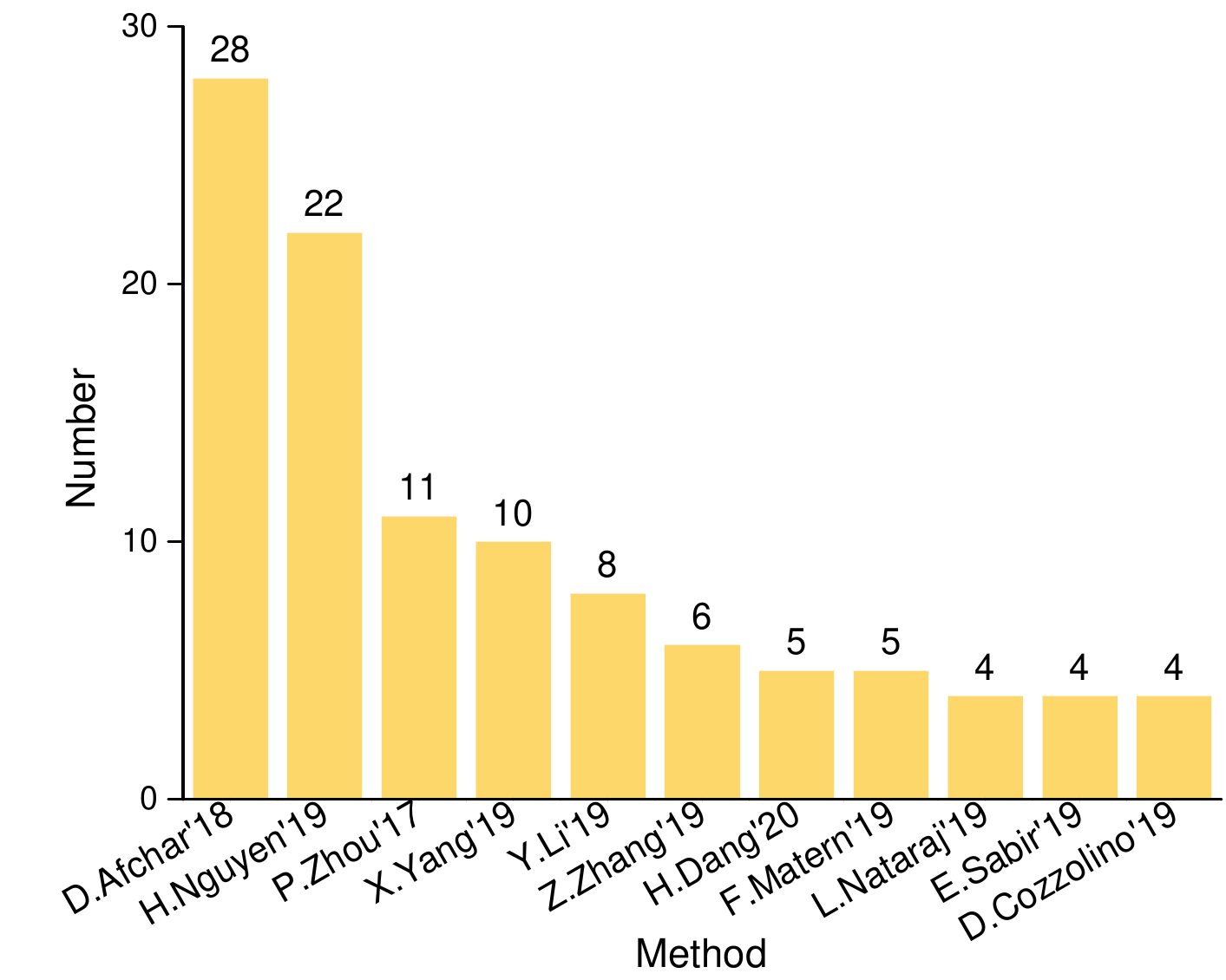}
	\includegraphics[width=0.49\linewidth]{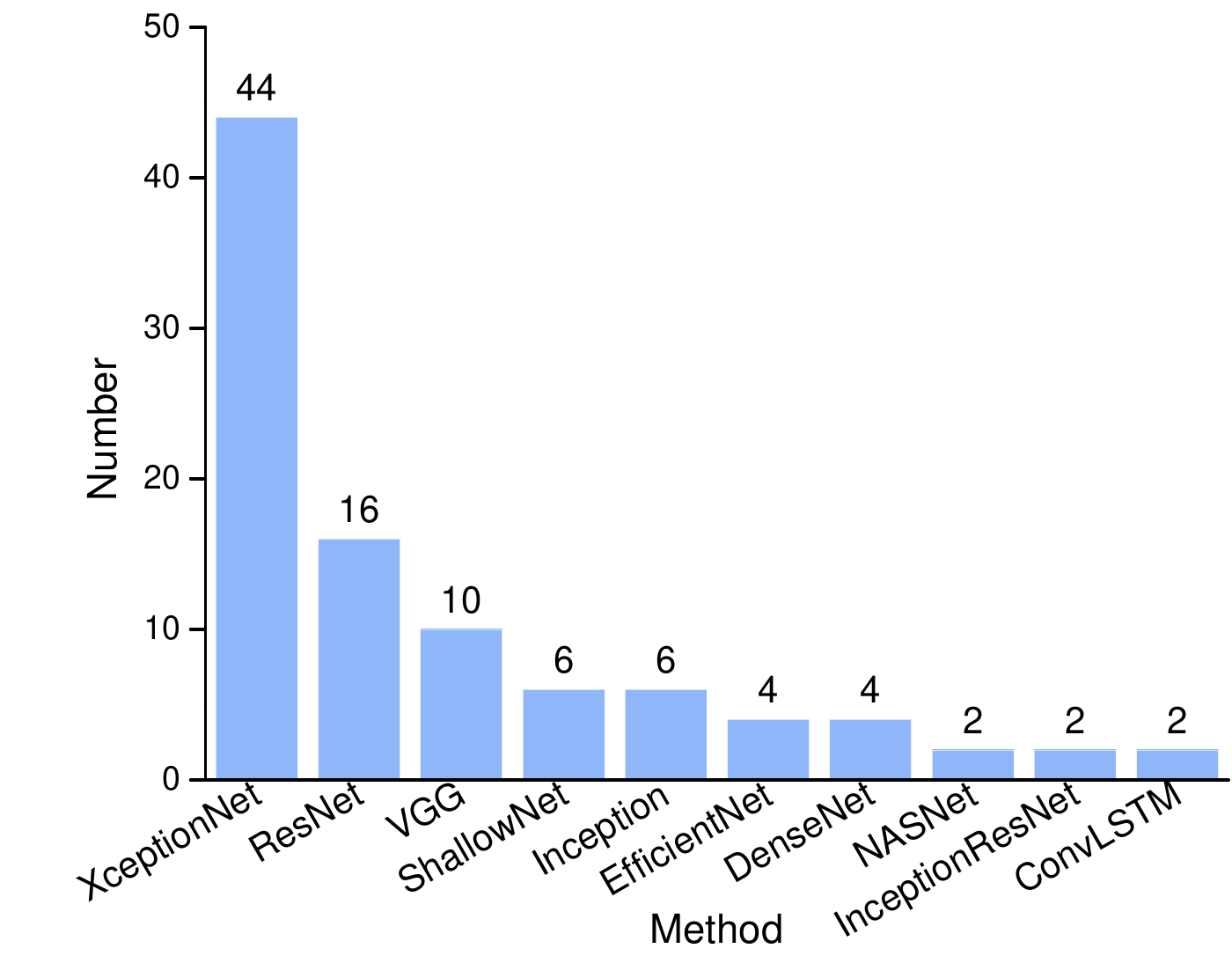}
	\caption{(L) Top-11 most popular DeepFake detection methods chosen as baselines. (R) Top-10 most popular ML-based methods chosen as baselines.}
	\label{fig:det_popular_baseline_rank}
\end{figure}
\begin{figure}
	\centering 
	\includegraphics[width=0.49\linewidth]{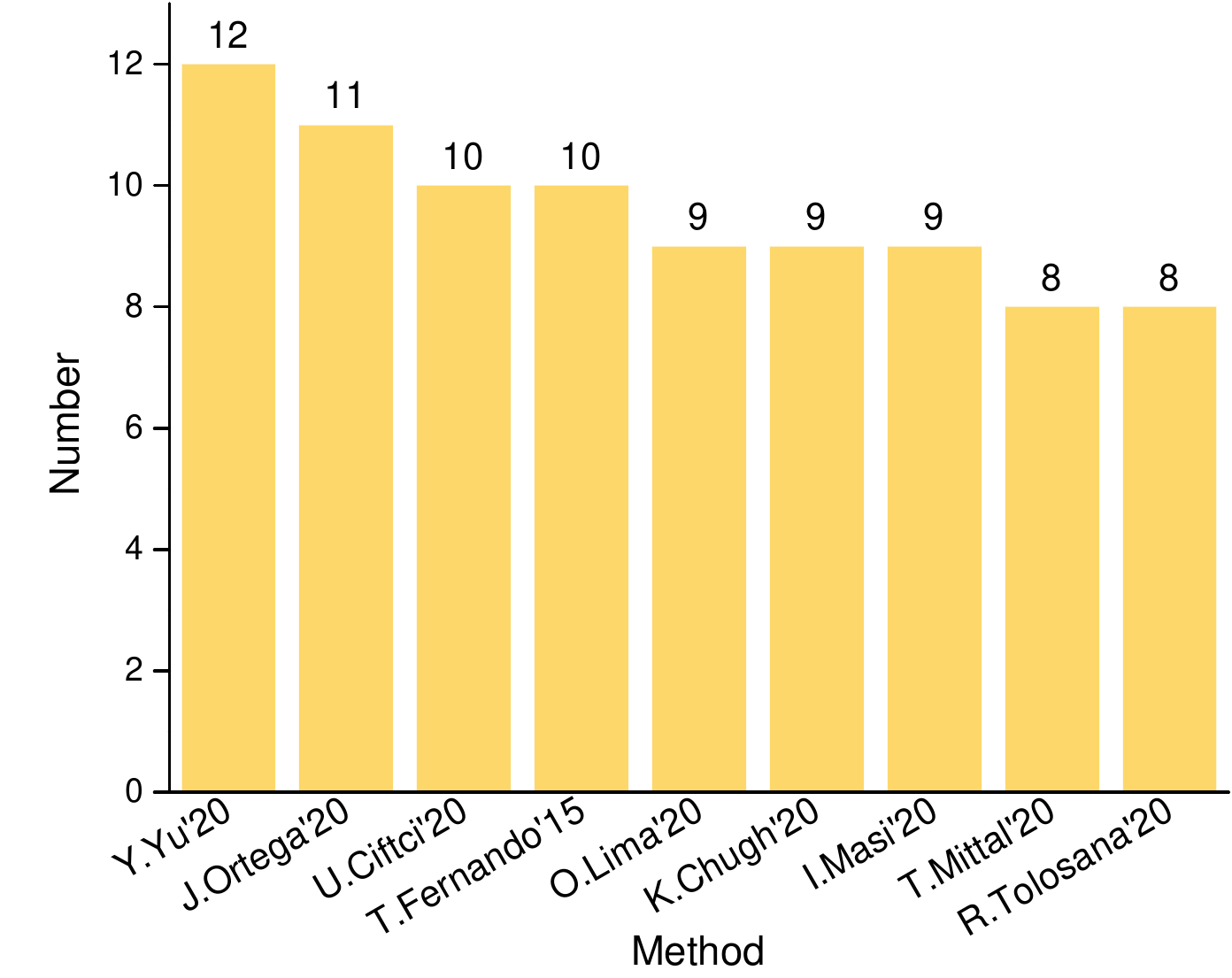}
	\includegraphics[width=0.49\linewidth]{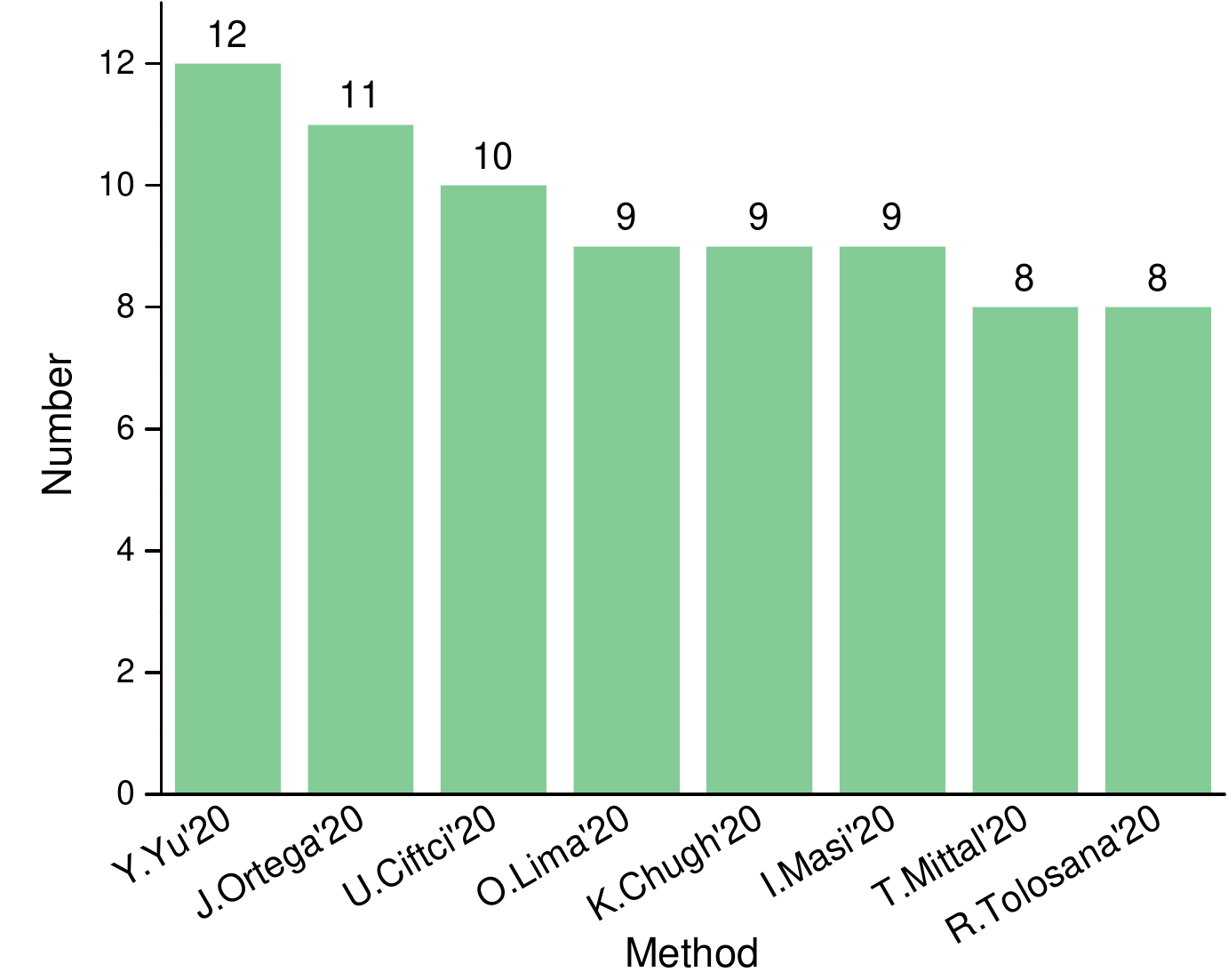}
	\caption{(L) Top-9 DeepFake detection methods that benchmark against the most number of baselines. (R) Top-8 DeepFake detection methods that benchmark against the most number of baselines in 2020.}
	\label{fig:det_extensive_rank}
\end{figure}

Moreover, we try to understand for a particular DeepFake detection method listed on the right column of Figure~\ref{fig:battle} , which previously published detectors has it benchmarked against.
Figure~\ref{fig:cd} presents a chord diagram to show the `competition' among later detectors and earlier ones. In the chord diagram, each node represents a DeepFake detection method, and a link connecting a node $A$ and a node $B$ means that the method $A$ has been compared with the method $B$ as a baseline in $A$'s paper, which infers that the method $A$ comes after $B$. 
We also notice that many of the DeepFake detectors are benchmarked against common machine learning (ML) based classifiers such as KNN and logistic regression, or popular DNNs such as the ResNet \citep{he2016deep}, \etc. Therefore, we also list out 30 popular ML-based methods in Figure~\ref{fig:cd}, and a link between a DeepFake detection method $A$ and an ML-based method $B$ can be established when the method $B$ is compared by the $A$'s paper. We provide an interactive diagram\footnote{\scriptsize{\url{http://www.xujuefei.com/dfsurvey}}} to facilitate the interpretation of the graph.
The top-$5$ popular baselines adopted in the evaluation are XceptionNet \citep{chollet2017xception}, \cite{afchar2018mesonet}, \cite{nguyen2019multi}, ResNet \citep{he2016deep}, and \cite{yang2019exposing_1}. XceptionNet, ResNet, and VGG are the top-$3$ CNN models that are employed as the baselines for comparison. In particular, XceptionNet is the most popular baseline and more than one-third studies compare with it. 
Figure~\ref{fig:det_popular_baseline_rank} (L) shows the top-11 most popular DeepFake detection methods chosen as baselines and Figure~\ref{fig:det_popular_baseline_rank} (R) shows the top-10 most popular ML-based methods chosen as baselines by various DeepFake detectors (See Figure~\ref{fig:cd}). We also identify the DeepFake detection methods that conduct the most extensive comparison experiments, that is, the number of baselines used by these methods are ranked in the top 9 according to the chord diagram. Figure~\ref{fig:det_extensive_rank} (L) shows the top-9 DeepFake detection methods that benchmark against the most number of baselines, and Figure~\ref{fig:det_extensive_rank} (R) shows the top-8 DeepFake detection methods that benchmark against the most number of baselines in 2020.

\begin{figure}
	\centering 
	\includegraphics[width=0.49\linewidth]{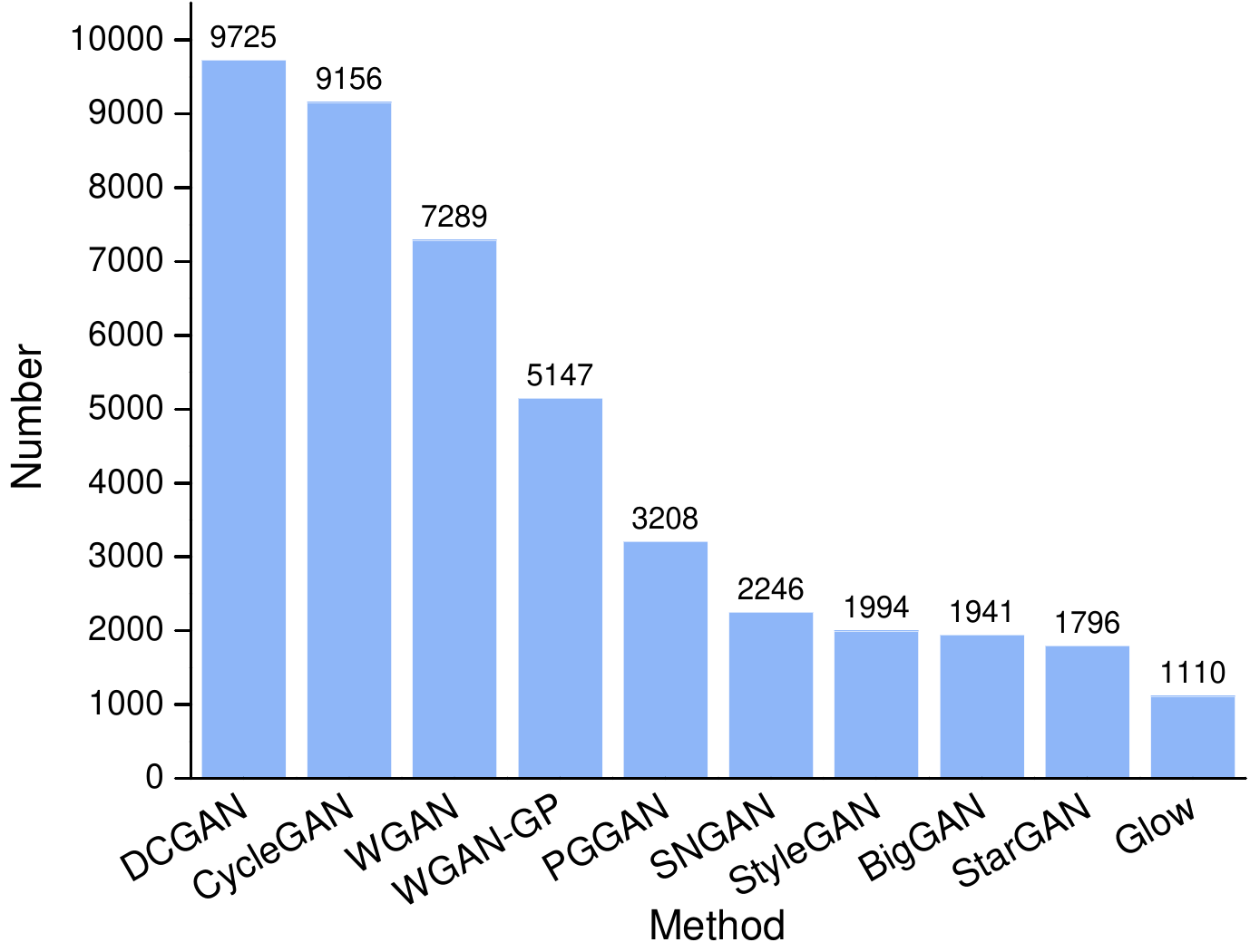}
	\includegraphics[width=0.49\linewidth]{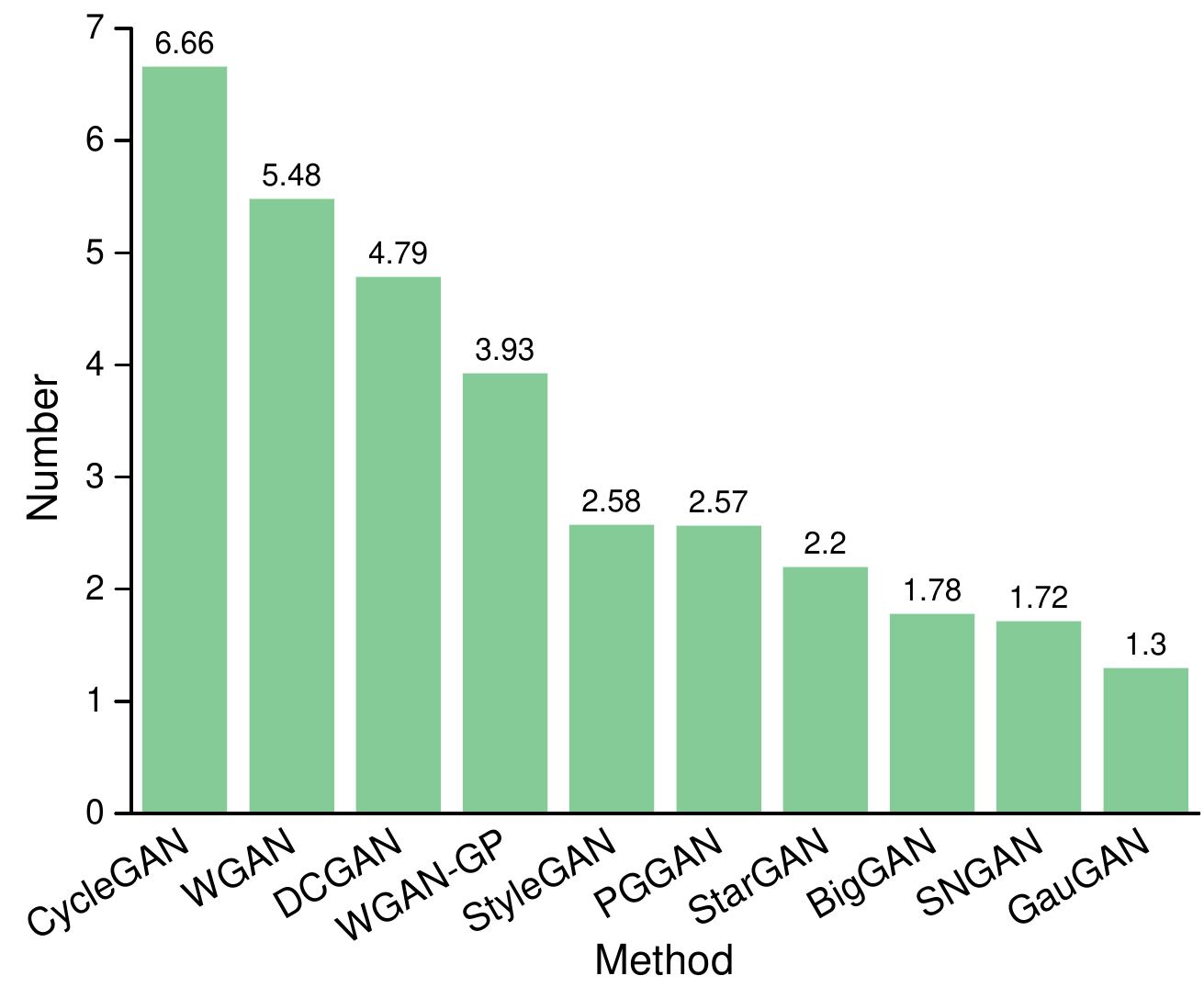}
	\caption{(L) Top-10 DeepFake generation methods or datasets based on citations. (R) Top-10 DeepFake generation methods or datasets based on normalized citations.}
	\label{fig:citation_generation_rank}
\end{figure}
\begin{figure}
	\centering 
	\includegraphics[width=0.49\linewidth]{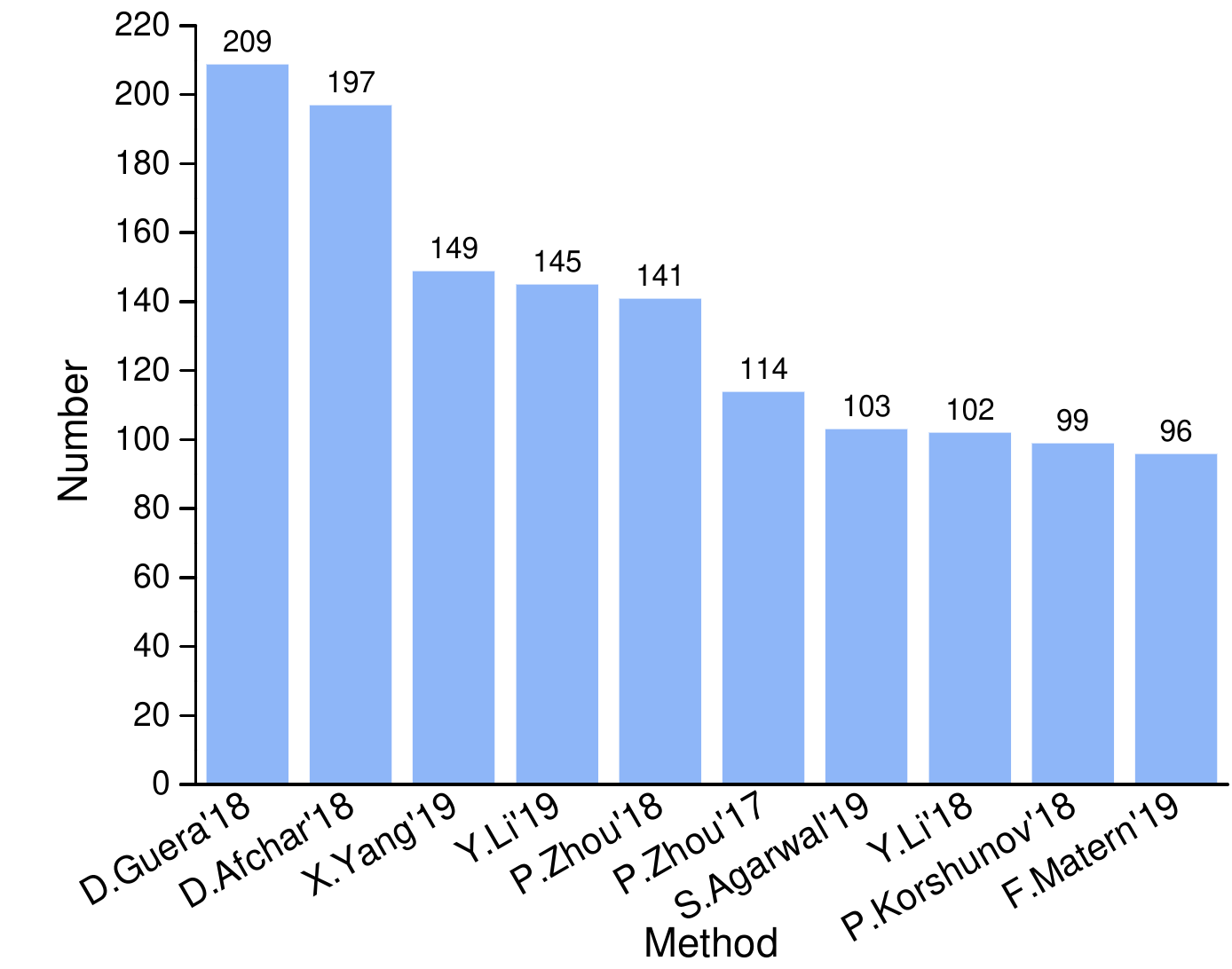}
	\includegraphics[width=0.49\linewidth]{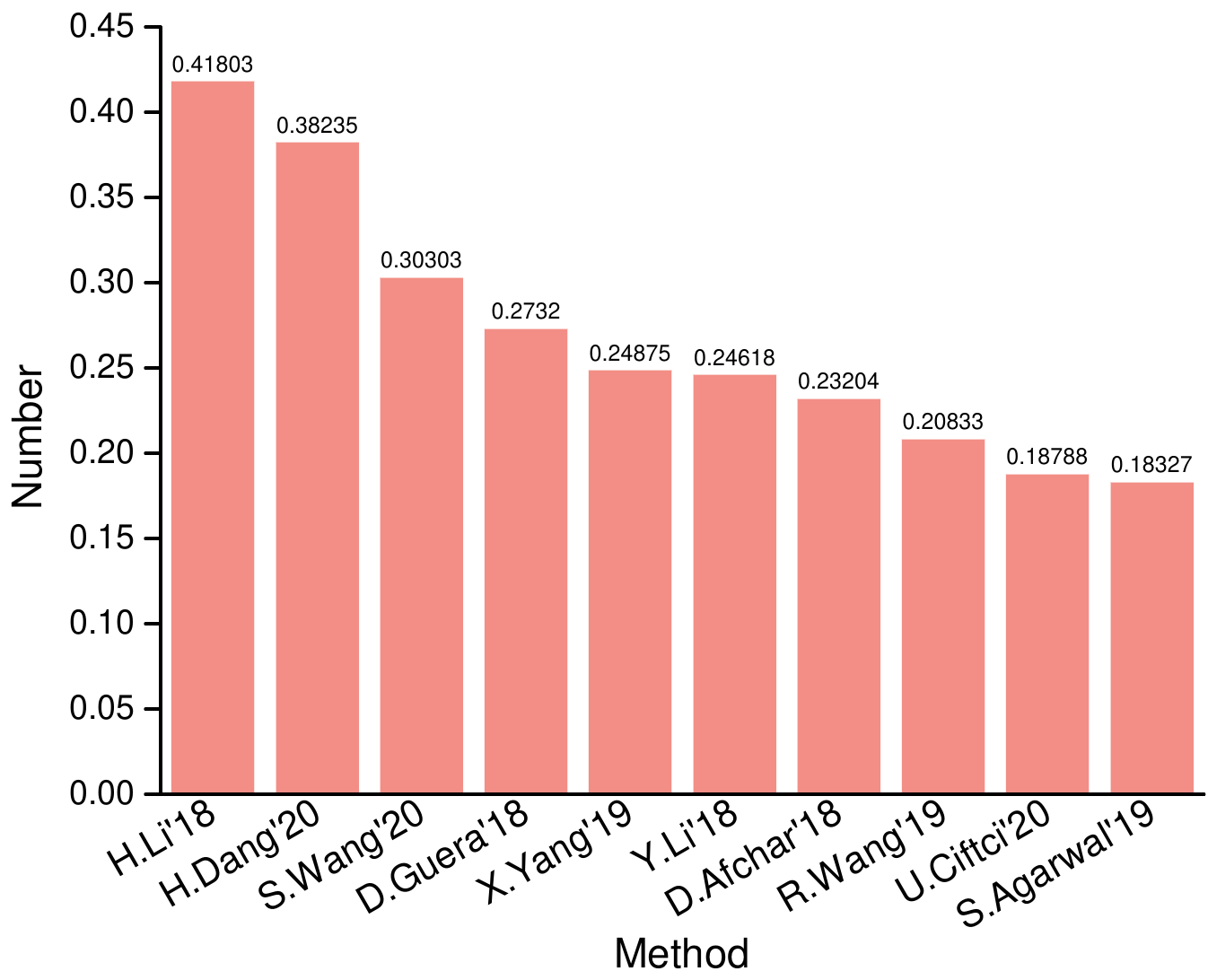}
	\caption{(L) Top-10 DeepFake detection methods based on citations. (R) Top-10 DeepFake detection methods based on normalized citations.}
	\label{fig:citation_detection_rank}
\end{figure}
\begin{figure}
	\centering 
	\includegraphics[width=0.49\linewidth]{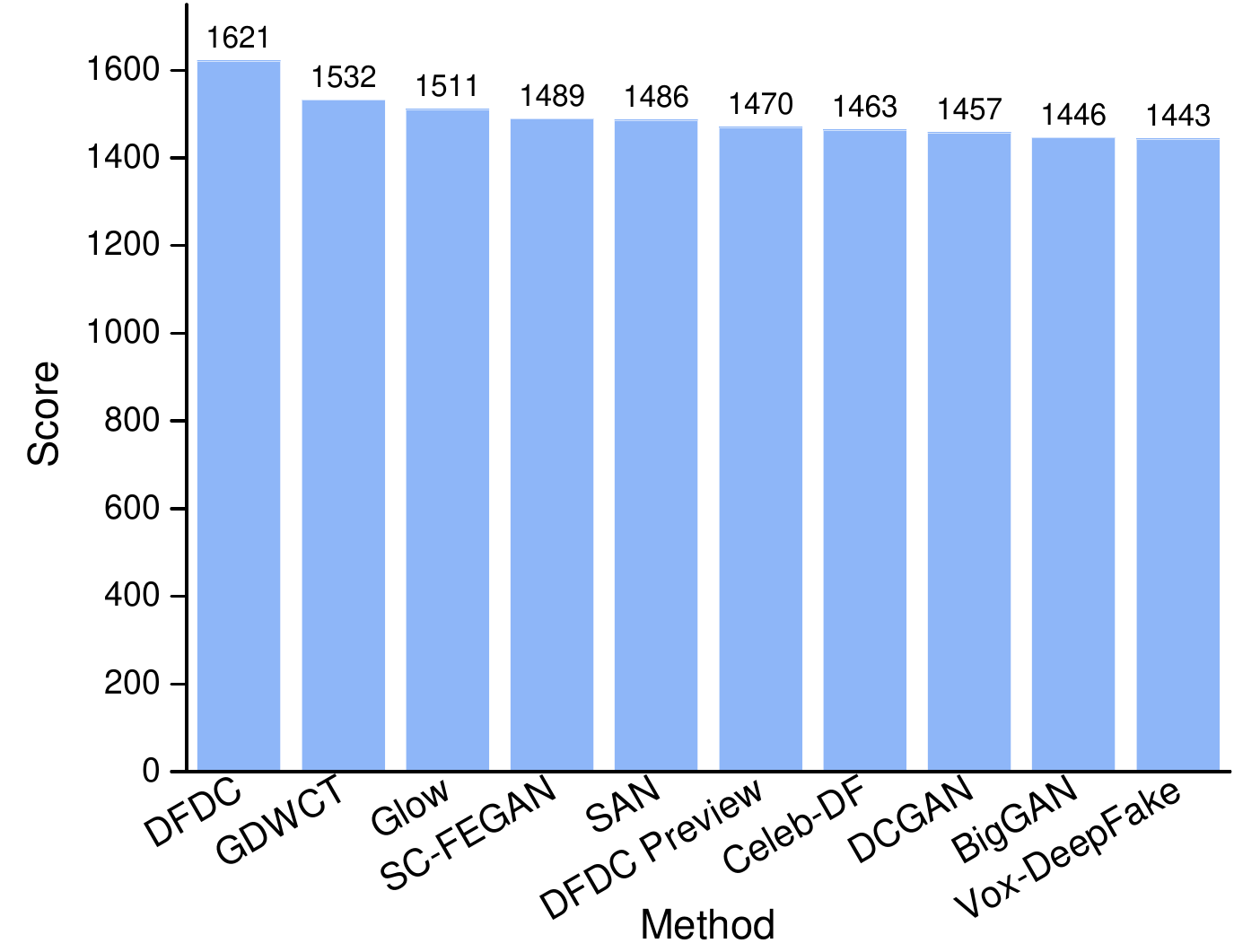}
	\caption{Top-10 DeepFake generation methods or datasets based on Elo rating \citep{Elo}. Default score is 1,400.}
	\label{fig:Elo_generation_rank}
\end{figure}

Another way is to measure the popularity of the DeepFake generation and detection methods through the citation count as well as citations normalized by the number of days since exposure. Figure~\ref{fig:citation_generation_rank} (L) shows the top-10 DeepFake generation methods or datasets based on their citations. Figure~\ref{fig:citation_generation_rank} (R) shows the top-10 DeepFake generation methods or datasets based on citations normalized by the number of days since exposure. Similarly, Figure~\ref{fig:citation_detection_rank} (L) shows the top-10 DeepFake detection methods based on citations and Figure~\ref{fig:citation_detection_rank} (R) presents the top-10 DeepFake detection methods based on citations normalized by the number of days since exposure. In addition, Figure~\ref{fig:Elo_generation_rank} shows the top-10 DeepFake generation methods or datasets based on Elo rating with a default score set to 1400.


Regarding the citation of DeepFake detection methods reported in Figure~\ref{fig:citation_detection_rank} and \ref{fig:battle}, it is actually difficult to identify some seminal milestone papers, although some papers have received more popularity than others. 
This phenomenon can be attributed to multiple factors and is a double-edged sword. 
The field of DeepFake detection is relatively new, thus it may take more time for any milestone papers to stand out. The lack of milestone papers can also be a positive indicator that the current state-of-the-art researches are multi-threaded and do not anchor on a few seminal works. Whether we are able to witness some new research hot zones emerge as the time goes by, the field is poised to progress at a fast pace.

\section{Evasion of DeepFake Detection}\label{sec:evasion}

\begin{figure}
	\centering 
	\includegraphics[width=\linewidth]{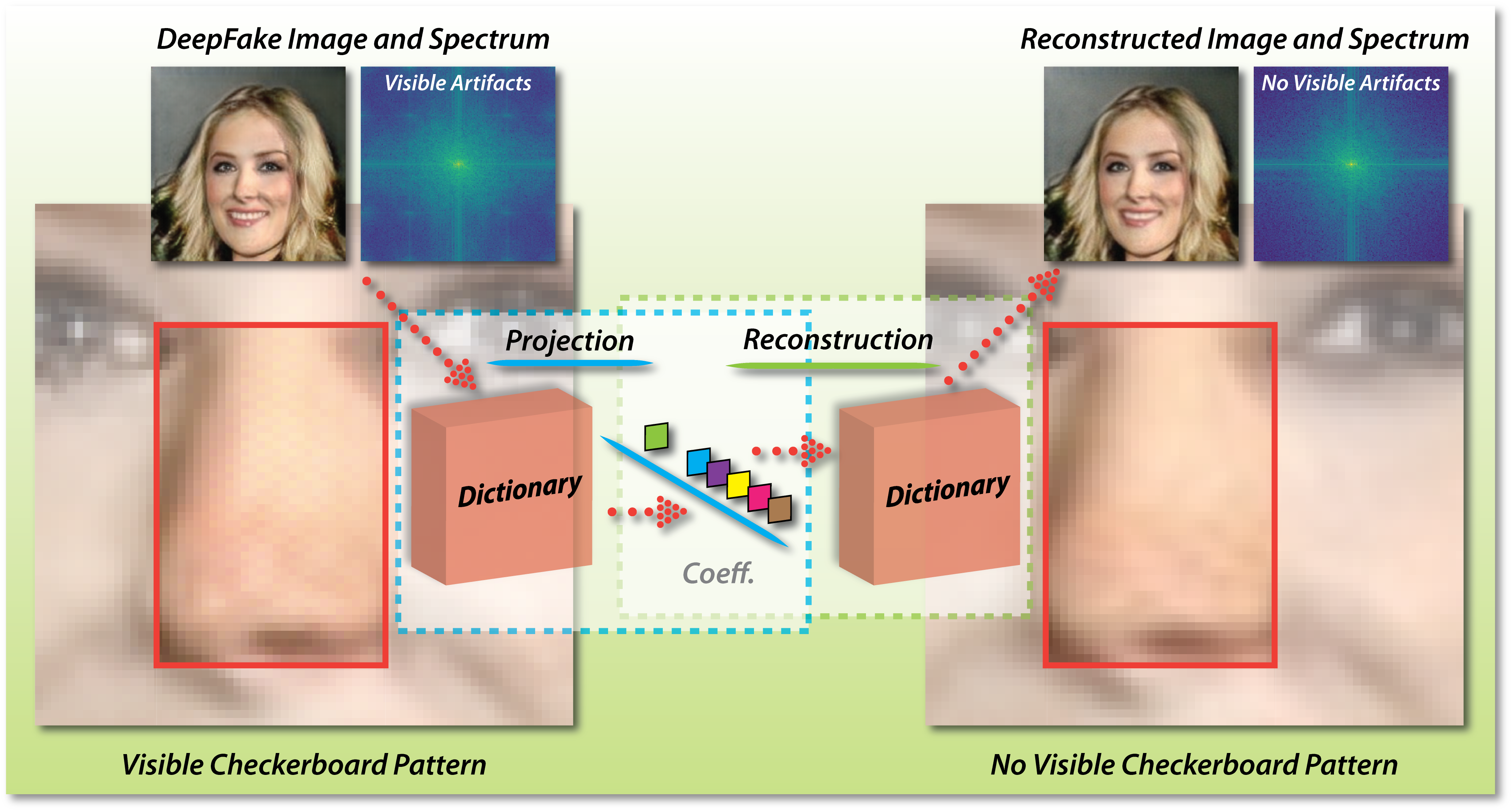}
 	\caption{Evasion of DeepFake detection via shallow reconstruction \citep{huang2020fakepolisher}.}
	\label{fig:teaser_evasion}
\end{figure}

With the rapid development of DeepFake detectors, researchers start paying attention to design methods to evade the fake faces being detected.
Specifically, given a real or fake face, evasion methods map it to a new one that cannot be correctly classified by the state-of-the-art DeepFake detectors, hiding the fake faces from being discovered. \revised{An exemplar pipeline of the evasion of DeepFake detection is shown in Figure~\ref{fig:teaser_evasion}}. We can roughly divide all methods into three types.

The \textit{first} type is based on the adversarial attack.
For example, \cite{carlini2020evading} add imperceptible adversarial perturbations to the fake/real faces and show that even the state-of-the-art DeepFake detectors are vulnerable to both white-box and black-box attacks \citep{carlini2017towards,brown2017adversarial} with significant accuracy reduction on the public datasets \citep{wang2020cnn,frank2020leveraging}.
Similarly, \cite{gandhi2020adversarial} use the fast gradient sign method \citep{goodfellow2014explaining} and C\&W attacks \citep{carlini2017towards} to fool DeepFake detectors. Then, they propose two methods with the Lipschitz regularization \citep{woods2019adversarial} and deep image prior \citep{ulyanov2018deep} to improve the adversarial robustness of DeepFake detectors. 
\cite{neekhara2021adversarial} further study the adversarial attack-based evasion methods on the more challenging DeepFake Detection Challenge (DFDC) dataset \citep{dolhansky2020deepfake} and find that the input-preprocessing steps, as well as face detection methods across DeepFake detectors, make the adversarial transferability difficult. Then, they implement a high transferability attack method based on the universal adversarial perturbations to alleviate the challenges.
In general, the adversarial attack-based methods inevitably introduce noise to the face images, leading to quality reduction.


The \textit{second} type of methods focus on removing the fake traces in the frequency domain. \revisedd{Recent works on the detection of DeepFake images have pointed out that they are actually easily distinguishable by artifacts in their frequency spectra. Thus, some generation methods attempt to repair the flaw in the generation procedure.}
\cite{durall2020watch} show that CNN-based generative deep networks with common up-sampling methods cannot reproduce spectral distributions of the real or natural training data, making the fake or generated images easily identified. To alleviate this drawback, they propose a novel spectral regularization objective term for training the GANs.
\revisedd{
\cite{jiang2020focal} also note this phenomenon and find that narrowing the frequency domain gap can improve the image synthesis quality further. To this end, they propose a frequency domain optimization target (\ie, focal frequency loss). The proposed loss enforces the model to dynamically focus on the frequency components that are hard to synthesize by down-weighting the easy frequencies. As a result, the method enhances the synthesis quality significantly.
}
%
%
\cite{jung2020spectral} identify a straightforward solution for this issue by equipping the generative models with a spectral discriminator, thus the new trained GANs can generate images with realistic frequency spectra.
These methods mainly focus on the mismatching between real and fake faces in the frequency domain while neglecting other potential factors that may make fake faces be identified easily.

The \textit{third} kind of methods regard evasion as a general image generation process and use advanced image filtering or generative models to mislead DeepFake detectors. 
\cite{huang2020fakepolisher} demonstrate that the DeepFake detectors can be easily evaded via the shallow reconstruction based on sparse coding and dictionary-based reconstruction.
In addition to the non-deep-learning solution, \cite{huang2020fakeretouch,deepnotch} propose to fool the DeepFake detectors by first adding the deliberate noise to destroy the fake trace in the frequency domain and then reconstructing the clear counterpart via a deep kernel prediction network.
Besides, \cite{neves2020ganprintr} remove the `fingerprints' in the fake faces through a pre-trained GAN model, which can spoof the DeepFake detectors while maintaining the visual quality of the fake faces.
In contrast to the above solutions of adding extra modules for evading DeepFake detection, \cite{osakabe2021cyclegan} propose to enhance the CycleGAN \citep{zhu2017unpaired} by equipping the fixed convolutional layers to remove the checkerboard artifacts.

\section{Horizon}\label{sec:horizon}

In this section, we touch upon the challenges and opportunities for future research directions surrounding DeepFake generation and DeepFake detection methods, as well as the evasion of DeepFake detection. The segmented discussions will be followed by a bird's-eye view comment of the entire DeepFake research field moving forward in the epilogue.

\subsection{Generation of DeepFakes}\label{sec:horizon:1}

We have surveyed and tabulated more than $91$ papers published through the peer-review process or posted on arXiv on the topic of DeepFake generation and the datasets tasked for the DeepFake detection. The observed findings and challenges can shed some light on the future work in creating more realistic and detection-evasive DeepFakes. A much improved DeepFake generation method will in turn push forward the development of the DeepFake detection method. 
\begin{itemize}[leftmargin=*]
    \item 
        \textbf{Lacking ultra high-resolution images.} Since PGGAN proposed a method to generate high-resolution $(1024\times1024)$ images, the new methods on synthesizing full fake images haven't been innovating towards higher resolution images. With the development of high-definition display resolution of phones or computers, $1024\times1024$ resolution may not enough in the near future.
    \item 
        \textbf{Limited properties of face manipulation methods.} The attribute manipulation methods can only change the properties given by the training set. Thus, the properties provided by these attribute manipulation methods are somewhat limited. An attribute manipulation method that is independent of the training set properties is desired.
    \item 
        \textbf{Less consideration of video continuity.} The identity swap and expression swap usually ignore the continuity of videos. They do not take physiological signals such as eye blink frequency, heart beat frequency into consideration.
    \item 
        \textbf{Lacking diversified DeepFake datasets.} The latest fake dataset are obsessed with being large scale. Most of them only expand the diversity of the content-related factors such as gender and age of the subject, the place where the face photo is taken, the illumination condition, \etc. The diversity in video quality such as various resolutions, various compression degrees, or other degradations commonly found in videos, \etc, have not been fully taken into account. Furthermore, the DeepFake generation method used by these fake dataset are somewhat limited, which may fall short when tasked to demonstrate the diversity of different generation methods. 
        The latest DeeperForensics-1.0 \citep{jiang2020deeperforensics} dataset has been a good attempt in this regard by incorporating diverse perturbations such as Gaussian blue, added noise, JPEG compression, contrast change, \etc. However, at the moment these perturbations are artificially added image-level degradations during post-processing, rather than organic video-level degradations such as bit-rate variations, choices of codec, \etc. We hope to see more organic degradations incorporated in the future generation of the dataset, \ie, DeeperForensics-2.0.
    \item 
        \textbf{Do not contain common sense fake.} Most of the fakeness lies in the image texture. However, the fake datasets usually do not contain common sense fake such as three-eye human, one-horned human, \etc. These fakes are obvious to humans but DeepFake detection methods may lack the common sense to judge properly.
    \item 
        \textbf{Lacking a platform for demonstrating the different fake datasets.} There lacks a platform for demonstrating the different fake datasets. On such a platform, we can directly see the different image style of the various fake datasets. The platform can also provide the information of the fake dataset such as (published year, image resolution, generation method, degraded or not, download link, the best detection method on each dataset, \etc).
    \item 
        \textbf{Lacking sub-categorized DeepFake detection datasets with respect to gender, age, ethnicity.} DeepFake detection benchmark datasets, like many other face recognition datasets, have data biases. For example, in many cases, the majority faces are from Caucasian males, and many of the internet-crawled datasets have the celebrity biases. With unbalanced datasets with respect to gender, age, and ethnicity to train the model, the learned DeepFake detector can become data biased as well. It is worthwhile to push for a more balanced DeepFake detection benchmark. We have seen some recent attempts to build DeepFake detection dataset based on one ethnicity group such as \citep{kwon2021kodf}. More development in this direction is needed.
    \item
        \textbf{Lacking multi-face DeepFake detection datasets.} For most of the existing DeepFake detection benchmark datasets, single face is DeepFake manipulated in the image or video, and when multiple faces are present, oftentimes, only one of the faces is DeepFake manipulated (usually the one with the largest detection bounding box size). There is a need to push for DeepFake detection benchmarks that involve multiple faces or with unknown number of DeepFake manipulated faces in the crowd. This effort will not only pose a new dimension of the challenge for DeepFake detectors, considering that the manipulation, if happens, may be hidden in the crowd and with unknown number. Also, this will foster new research into the DeepFake detection method where cues can now be drawn beyond individual faces and from the peers in the images or videos. It is good to see that one of the latest benchmarks \citep{le2021openforensics} is created towards that goal, and we hope to see more.
\end{itemize}

\subsection{Detection of DeepFakes}

We have investigated and tabulated more than $117$ papers published through the peer-review process or posted on arXiv on the topic of DeepFake detection. We have observed some interesting findings and challenges, after reviewing the papers, which could inspire future work in defending DeepFakes more effectively.
\begin{itemize}[leftmargin=*]
    \item 
        \textbf{Lacking public AI-synthesized image datasets.} Almost all the existing studies build their own image dataset with various GANs to evaluate the effectiveness of their method in defending still image DeepFakes. They do not have a consensus on which forgery image datasets need to be used in evaluation. These studies claim that they have achieved competitive results in detecting various GAN-synthesized images built on their own. However, the quality of these generated fake images is still unknown, \ie, if there are any obvious artifacts that exist in the image. A public GAN-synthesized fake image dataset needs to be developed by the community for advancing this challenging research field. 
    \item 
        \textbf{Lacking competitive baselines in comparison.} In evaluating the performance of their proposed methods, existing studies prefer to employ some simple baselines (\eg, simple DNN-based methods, naive methods by leveraging perceptible artifacts) rather than the SOTA work to demonstrate that they have beaten the prior studies. We hope that future studies could compare their work with some competitive baselines which are highlighted in Tables \ref{tab:fake_detection_1}, \ref{tab:fake_detection_2}, and \ref{tab:fake_detection_3} to demonstrate the advances of their work.  
    \item 
        \textbf{Generalization abilities of DeepFake detectors.} Tackling the unknown DeepFakes is one of the key challenges in fighting against DeepFakes. In recent years, a series of studies are working towards this goal to develop more generalized methods. Unfortunately, these works are merely evaluated on simple DeepFake video datasets, like FaceForensics++. We hope that future work can focus more on challenging datasets. 
    \item 
        \textbf{Robustness of DeepFake detectors.} In the real-world, DeepFakes can easily suffer from various degradations, such as image/video compression, added Gaussian noises, blurring, low-light \citep{btas15_pokerface}, low-resolution \citep{pr19_ssr2}, \etc. Existing studies proposed various robust methods to tackle this simple degradation. However, more than 90\% of methods leverage DNNs as their backbone to determine real and fake in the final classifier. The DNNs are vulnerable to adversarial noise attacks with imperceptible additive noises, which is demonstrated by prior works. Unfortunately, we observe that all the existing studies failed in evaluating their robustness against adversarial noise attacks. In addition, the SOTA detectors may fall short when faces are under occlusions such as facial masks \citep{zhu2022masked} where only the eye region is visible \citep{btas16_fastfood,tip15_spartans}, heavy makeups, heavy facial hairs, \etc.
    \item 
        \textbf{Capabilities of DeepFake detectors.} Improving the generalization capabilities to tackle the emerging unknown DeepFakes, enhancing the robustness against various DeepFake degradations including simple transformations and adversarial attacks, and explaining why the detector works are the three key factors in developing a practical DeepFake detector which could be deployed in the wild. In reviewing the recent papers with regard to detecting DeepFakes, we find that less than ten papers have evaluated the capabilities of their method from all three perspectives.
    \item 
        \textbf{A Comprehensive evaluation metrics.} The performance of DeepFake detectors is highly determined by the quality of DeepFakes. The low-quality DeepFakes (\eg, DeepFake-TIMIT, FaceForensic++) with observable artifacts could be easily identified by almost all the DeepFake detectors with high confidence, while, the challenging high-quality DeepFakes (\eg, Celeb-DF, DFDC) which could fool our eyes can be hardly determined by detectors. The existing studies report their experimental results by merely considering the detection accuracy and false alarms, which ignore the relation with the quality of DeepFakes, especially from the self-built DeepFake datasets. We hope that more comprehensive experimental results by considering the quality of DeepFakes should be considered in future work. Thus, new metrics for measuring the quality of DeepFakes need to be proposed by researchers.
    \item 
        \textbf{A Platform for evaluation.} In Tables \ref{tab:fake_detection_1}, \ref{tab:fake_detection_2}, and \ref{tab:fake_detection_3}, we can find that the existing DeepFake detectors can easily achieve more than $90\%$ detection accuracy in fighting the common DeepFakes. However, in a DeepFake Detection Challenge (DFDC) built by Facebook, the final competition results show that the winner can only give less than $70\%$ accuracy in detecting DeepFakes. Another DeepFake detection challenge, called DeeperForensics Challenge 2020 \citep{jiang2021deeperforensics}, is hosted on DeeperForensics-1.0 dataset which is a real-world face forgery detection dataset. However, only $25$ teams made valid submissions, and only one method adopted for generating DeepFakes in DeeperForensics-1.0. The results cannot represent the SOTA performance in DeepFake detection. Thus, the DeepFake is still a real threat to the community and academia needs to develop more practical detection methods. Obviously, the reported experimental results in academic papers can not reflect the true performance of their methods. A platform, incorporating the challenging DeepFake datasets and competitive baselines, is not ready for evaluating the true performance of existing DeepFake detectors and the future DeepFake detectors. FaceForensic++ provides a simple platform with low-quality and simple CNNs as baselines, which might fall short when tackling the ever-progressing DeepFakes.
\end{itemize}

\subsection{Evasion of DeepFake Detection}

We have discussed three kinds of methods for evading DeepFake detection in Section~\ref{sec:evasion}, which mainly aims at misleading the DeepFake detectors or removing artifacts introduced by DeepFake generations. In the near future, we hope that the evading methods would evade new DeepFake detectors by developing more advanced adversarial attacks that consider natural degradation in the real world and deeply removing the fake traces in both images and videos. More specifically, the following directions should be noted:
\begin{itemize}[leftmargin=*]

    \item
    \textbf{Misleading DeepFake detection via natural degradation.} 
    Existing adversarial attack-based evasion methods mainly rely on the additive adversarial perturbations that do not exist in the real world and might be detected by recent works on detecting adversarial examples \citep{pang2018towards,zheng2018robust}. Moreover, the state-of-the-art defense methods are also able to invalidate the adversarial attacks, thus making the evasion methods less effective.
    A possible solution for this problem is to design natural degradation-based adversarial attacks, \eg, motion blur, light variation, shadow synthetic, \etc, allowing generating realistic examples while misleading the DeepFake detection.
    For example, \cite{guo2020watch} realize an adversarial blur attack that can generate realistic-blurred images and mislead the state-of-the-art deep neural networks \citep{iccv21_advmot}. Similar works are proposed for natural degradations like weather elements \citep{arxiv22_advrain,zhai2020it,arxiv21_advhaze}, exposure \citep{cvpr22_cosal,gao2020making,cheng2020adversarial}, lighting \citep{arxiv21_ara,ijcai21_ava,tian2021bias,sun2022ala}, shadow \citep{arxiv21_sharel,fu2021auto}, defocus blur \citep{arxiv21_advbokeh}, \etc. 
    In the future, we can employ these attacks to evade the DeepFake detection with the natural adversarial examples that can be hardly defended through the state-of-the-art defense methods designed for additive adversarial perturbations. 
    
    \item
    \textbf{Faking the physiological signal in fake videos.} 
    The state-of-the-art DeepFake detector starts using physiological signal, \eg, heart rate extracted from the video, as an effective fake indicator \citep{qi2020deeprhythm}, because even the advanced GAN methods can hardly preserve the heart rate signal that usually presents as fine-grain color variation among frames. 
    To evading such new detectors, we should develop a novel evading method allowing the processed fake videos to also contain the normal heart rate single. 
    This actually requires us to learn how to add sequential color variations into the frames in a fake video, letting the rhythm detection methods obtain normal heart rate.
    
    \item
    \textbf{Joint perception and appearance fake trace removal.}
    Existing fake trace removal-based evading methods mainly focus on how to remove the known appearance artifacts, \eg, spectral distributions in the frequency domain, introduced by DeepFake generations while ignoring their influence on the perception, \eg, deep representations of fake faces, which seems to be the essential factor for effective DeepFake detection.
    Hence, in the near future, a more advanced fake trace removal could be explored by jointly removing the artifacts and perceptions of fake traces. 

\end{itemize}


\subsection{Epilogue and the Next Chapter}\label{sec:horizon:4}

Now that we have discussed the existing challenges and opportunities for future studies, it is a good segue into some final thoughts regarding DeepFakes. 

Based on the discussions throughout this survey paper, we can see that at the moment, the work on DeepFake detection heavily relies on curated datasets with the latest DeepFake generation methods incorporated that show the highest level of realisticity when the dataset is created. We would like to emphasize the significance of the continued progression of such datasets. 
Unlike the acclaimed ImageNet \citep{deng2009imagenet} classification tasks, whose image classification difficulty will remain relatively unchanged throughout the years, the DeepFake detection tasks are becoming increasingly more difficult year over year since the DeepFake generation method can produce increasingly more realistic DeepFakes. In this sense, it is imperative to hold periodic ImageNet-style contests and/or produce updated DeepFake detection datasets to keep track of the latest DeepFake generation methods and encourage competition among various research groups in order to advance the effort of countering malicious DeepFakes. A very fitting example would be the latest DeeperForensics Challenge 2020 on Real-World Face Forgery Detection \citep{jiang2021deeperforensics}.


\revisedd{Needless to say, the various DeepFake datasets, produced by the DeepFake generators, are valuable assets for developing next-generation DeepFake detectors. Over the years, we have seen that the datasets have grown tremendously in sizes, quality, diversity, and levels of challenging scenarios. How will the datasets evolve in the next five, ten years is unknown at this point, but we envision that the DeepFake datasets may evolve into a dichotomy following similar trends as other computer vision datasets. On one hand, there will be convergence of many dataset sources into a few very large-scale standardized evaluation datasets for the DeepFake community, similar to the scales of the ImageNet dataset or the COCO object detection dataset \citep{lin2014microsoft}. These large-scale datasets will be less frequently updated and will most likely be served as the go-to benchmark and tools for developing and evaluating DeepFake algorithms. Next-generation large-scale general-purpose DeepFake foundational models can be developed on these large-scale datasets. In computer vision and natural language processing, foundational models \citep{bommasani2021opportunities} are those models trained on broad data at scale (usually in multi-modality such as vision and language) and are adaptable to a wide range of downstream tasks. Examples of vision and language foundational models include Florence \citep{yuan2021florence}, CLIP \citep{radford2021learning}, ALIGN \citep{jia2021scaling}, Wu Dao \citep{wudao}, \etc. We envision that similar general-purpose DeepFake detectors that are able to deal with the majority of DeepFake types will emerge. On the other hand, proprietary datasets that are smaller in scale that are more flexible and more frequently updated are also likely to emerge. These datasets, on the contrary, are best catered towards developing and evaluating \emph{ad-hoc} DeepFake algorithms for particular types of DeepFake generators that are newly emerged, or for particular long-tail scenarios, and more importantly, for finetuning the aforementioned DeepFake foundational models with particular downstream DeepFake-related tasks. With the two types of datasets discussed above, \ie, large-scale general-purpose datasets vs. smaller-scale proprietary datasets, if one DeepFake dataset is out of date, it can still be beneficial to the community by being incorporated into the first type of the datasets because it is deemed very valuable for maintaining the large scale and diversity of the datasets. Meanwhile, newly emerged datasets from the latest DeepFake generators will carry the weight for pushing the development of next-generation DeepFake detectors, and when they become obsolete, they will be replaced by newer ones, and they will still find their way back to contribute to the first type of general-purpose datasets.}



There have not been many studies on the intersection of DeepFake generation and adversarial attack. The current research landscape is still largely fragmented \wrt the two domains. Most common adversarial attacks create image-level pixel perturbations to alter the classification output through access to the white-box model parameters. However, such perturbation is not limited to image-level representations. When we decouple the face representation into various attribute/semantic latent sub-representations such as identity, expression, gender, ethnicity, \etc, from the adversarial attack point of view, we can see clearly that the identity swapped DeepFake is merely an adversarial perturbation on the identity sub-representation, and similarly for expression swapped DeepFakes. From this viewing angle, how the DeepFake narrative can fit into the adversarial robustness studies is worth looking forward to.

Incorporating other multi-media modalities such as voice and sound will help to counter malicious DeepFakes. In this survey, although we have focused on the image and video modalities, it is quite intuitive to monitor the realism of the DNN generated fake voice and sound \citep{deepsonar} in a DeepFake video. Being able to detect multi-modal fake traces, such as traces from standalone visual cues, standalone acoustic cues, as well as the interactions between them, such as synchronization, the combined effort in detecting DeepFakes will most likely be boosted by practicing Liebig's barrel theory.

One issue surrounding the battle between DeepFake generation methods and DeepFake detection methods is that the detection method is usually lagging behind. This is generally true for adversaries and defenders in any ``battle'' scenarios such as the development of Covid-19 vaccination happening after the Covid-19 outbreak because the knowledge of the virus is required for the vaccination development. The same applies here to DeepFake problems. The currently developed DeepFake detection methods are still somewhat myopic in the sense that they can most confidently tackle DeepFakes generated by existing methods, but will be outcompeted by DeepFakes generated by future generation techniques. How to make DeepFake detection methods forward-looking and remain effective, with or without small tweaks, through iterations of DeepFake generation methods is an open challenge. \revisedd{The earlier-mentioned DeepFake foundational models can potentially provide a more friendly training paradigm for the continual evolving of the DeepFake algorithms on both sides of the battleground.} Meanwhile, there are some proactive measures that the defenders (such as social media platforms where DeepFakes are most likely disseminated) can take in order to become more effective in fighting malicious DeepFakes, for example, through the responsible disclosure of generative models using GAN fingerprinting \citep{yu2020responsible}, or embedding an invisible tag into the original clean image uploaded by the user which can remain retrievable after the DeepFake generation process so that at a later time when the DeepFake version is re-uploaded by the bad actor, the platform is able to retrieve the tag and block the dissemination \citep{wang2020deeptag}.

From an algorithmic point of view, two of the major research hot topics in the machine learning community at the moment are self-supervised learning and transformer for vision and language problems. Self-supervised learning, free of labels, enables continuous life-long learning on an infinite and smoothly changing data stream \citep{sun2020test}. What would self-supervised learning in the domain of DeepFake detection be like? Transformer language models are now being operationalized to computer vision domain with the latest ImageGPT \citep{chen2020generative}, DALL-E \citep{dalle}, and subsequent CLIP \citep{radford2021learning} all from OpenAI, as well as Google's ALIGN \citep{jia2021scaling}. They, together with other latest generative approaches such as vector quantized variational autoencoder (VQ-VAE) \citep{razavi2019generating}, VQGAN + autoregressive transformer \citep{esser2021taming}, \etc, will definitely supplement and enhance the DeepFake generation techniques. \revisedd{We have seen the latest vision transformers (ViT) \citep{dosovitskiy2020image} already equipped with the state-of-the-art image generation capabilities such as the TransGAN \citep{jiang2021transgan} and Paint Transformer \citep{liu2021paint}. Although there hasn't been a ViT dedicated for DeepFake generation or detection yet, the unprecedentedly fast growing pace and increasing ubiquity in ViTs \citep{khan2021transformers} to tackle various computer vision problems will eventually turn the landscape of DeepFake generation and detection into a ViT-based one. 
With both sides of the DeepFake battleground now equipped with the newest technological expertise, the clash between the two parties will, for sure, spark flaming research in the foreseeable future.}


The humanity may reach a stage where DeepFakes have become so genuinely looking that they are beyond human and machine's capability to distinguish from the real ones, a \emph{``DeepFake singularity''}, if you will. If this day is inevitable, be it a utopia or a dystopia, perhaps a more interesting era is upon us. Are we brave enough to embrace it?

\section{Conclusion}\label{sec:concl}

In this survey, we have provided a comprehensive overview and detailed analysis of the research work on the topic of DeepFake generation, DeepFake detection as well as evasion of DeepFake detection, with more than $318$ research papers carefully surveyed. \revised{We have presented the taxonomy of various DeepFake generation methods and the categorization of various DeepFake detection methods along with highlights of the technical evolution of the methods}, and more importantly, we have showcased the battleground between the two parties with detailed interactions between the adversaries (DeepFake generation methods) and the defenders (DeepFake detection methods). The battleground allows fresh perspective into the latest landscape of the DeepFake research and can provide valuable analysis towards the research challenges and opportunities as well as research trends and directions in the field of DeepFake generation and detection. We hope that this survey paper can help empower and fast-track researchers and practitioners in this field to identify the most pressing research topics and attract more researchers to contribute to this emerging and rapidly growing field.

\section{Acknowledgments}
This research was partly supported by the National Key Research and Development Program of China under grant No. 2021YFB3100700, the Fellowship of China National Postdoctoral Program for Innovative Talents under No. BX2021229, the Natural Science Foundation of Hubei Province under No. 2021CFB089, the Fundamental Research Funds for the Central Universities under No. 2042021kf1030, the National Natural Science Foundation of China (NSFC) under No. 61876134.
The work was also supported by the National Research Foundation, Singapore under its the AI Singapore Programme (AISG2-RP-2020-019), the National Research Foundation, Prime Ministers Office, Singapore under its National Cybersecurity R\&D Program (No. NRF2018NCR-NCR005-0001), NRF Investigatorship NRFI06-2020-0001, the National Research Foundation through its National Satellite of Excellence in Trustworthy Software Systems (NSOE-TSS) project under the National Cybersecurity R\&D (NCR) Grant (No.~NRF2018NCR-NSOE003-0001). We gratefully acknowledge the support of NVIDIA AI Tech Center (NVAITC) to our research.


{
\footnotesize
\bibliographystyle{spbasic} 
\bibliography{ref}
}